\PassOptionsToPackage{table}{xcolor}
\PassOptionsToPackage{hyphens}{url}
\documentclass[a4paper,fleqn]{cas-dc}

\usepackage{microtype}
\usepackage[numbers]{natbib}
\AtBeginDocument{\setlength{\bibsep}{0pt plus 0.2ex}}
\usepackage{booktabs}
\usepackage{graphicx}
\usepackage{multirow}
\usepackage{makecell}
\usepackage{rotating}
\usepackage{pifont}
\newcommand{\cmark}{\ding{51}}
\newcommand{\xmark}{\ding{55}}
\newcommand{\pmark}{$\circ$}

\definecolor{rowshade}{gray}{0.94}
\newcommand{\zebra}[1]{
  \ifodd#1\relax\rowcolors{#1}{rowshade}{}\else\rowcolors{#1}{}{rowshade}\fi}
\newcommand{\sem}[1]{\,{\tiny$\pm$#1}}
\newcommand{\up}{$\uparrow$}
\newcommand{\down}{$\downarrow$}
\newcommand{\readout}{\textit{readout}}
\newcommand{\snd}[1]{\underline{#1}}
\newcommand{\trd}[1]{\textit{#1}}
\newcolumntype{P}[1]{>{\raggedright\arraybackslash\hspace{0pt}}p{#1}}  
\usepackage{tabularx}
\newlength{\mcellw}\newlength{\mcellx}
\newcommand{\mcell}[2]{%
  \settowidth{\mcellw}{#1}%
  \settowidth{\mcellx}{#2}%
  \ifdim\mcellx>\mcellw \setlength{\mcellw}{\mcellx}\fi
  \parbox[t]{\mcellw}{\raggedleft\strut#1\par\strut#2}}
\newcolumntype{Y}{>{\raggedright\arraybackslash\hspace{0pt}}X}
\newcolumntype{Z}{>{\raggedleft\arraybackslash}X}

\newcommand{\auditAllPaired}{1{,}734}  
\newcommand{\auditScope}{958}
\newcommand{\auditResolvable}{455}
\newcommand{\auditCorrect}{393}
\newcommand{\auditWrong}{62}
\newcommand{\auditUnresolved}{409}
\newcommand{\auditBracket}{94}
\newcommand{\auditRate}{86.4}
\newcommand{\auditCoverage}{47.5}  
\newcommand{\auditMajority}{75.6}  
\newcommand{\taxCells}{92}  
\newcommand{\taxNegative}{92}
\newcommand{\taxWorst}{-26.3}
\newcommand{\taxMedian}{-7.0}
\newcommand{\auditPrevScope}{1{,}111}  
\newcommand{\auditPrevResolvable}{521}
\newcommand{\auditPrevRate}{78.9}
\newcommand{\auditPrevBelowRate}{87.9}
\newcommand{\auditBelowRate}{94.9}
\newcommand{\auditExclCells}{91}  
\newcommand{\auditExclResolvable}{50}
\newcommand{\auditExclCorrect}{9}
\newcommand{\auditExclWrong}{41}
\newcommand{\auditExclRate}{18.0}
\newcommand{\auditMidScope}{1{,}020}  
\newcommand{\auditMidResolvable}{471}
\newcommand{\auditMidRate}{85.4}
\newcommand{\auditMidBelowRate}{94.3}
\newcommand{\auditHundredCells}{62}  
\newcommand{\auditHundredResolvable}{16}
\newcommand{\auditHundredCorrect}{9}
\newcommand{\auditHundredWrong}{7}
\newcommand{\auditHundredRate}{56.3}
\newcommand{\auditHundredBelow}{2}  
\newcommand{\auditResidSD}{2.0}  
\newcommand{\computeRuns}{9{,}471}
\newcommand{\computeCells}{3{,}077}  
\newcommand{\computeGpuHours}{3{,}885}
\newcommand{\computeHhundred}{3{,}330}
\newcommand{\computeAmpere}{473}
\newcommand{\computeNvl}{82}
\newcommand{\computeKwh}{3{,}046}
\newcommand{\computeCarbon}{518}
\newcommand{\computePerRun}{55}
\newcommand{\vitBest}{75.3}  
\newcommand{\vitBestBase}{61.4}  
\newcommand{\vitBestDelta}{13.9}  
\newcommand{\denseCells}{216}  
\newcommand{\densePops}{6}
\newcommand{\denseVocOneDelta}{0.39}  
\newcommand{\denseVocOneRel}{6.2}
\newcommand{\denseVocPeak}{2.34}  
\newcommand{\denseVocFull}{0.46}
\newcommand{\denseVocFullRel}{0.9}
\newcommand{\densePcFull}{0.17}  
\newcommand{\densePcFullRel}{2.6}
\newcommand{\denseRelLo}{1}
\newcommand{\denseRelHi}{22}
\newcommand{\denseLawCells}{30}
\newcommand{\denseLawResolvable}{9}
\newcommand{\denseLawCorrect}{9}
\newcommand{\denseLawBracket}{2}
\newcommand{\denseLawUnres}{19}
\newcommand{\denseLawPixLo}{52}
\newcommand{\denseLawPixHi}{78}
\newcommand{\detCells}{36}  
\newcommand{\detCellCount}{12}  
\newcommand{\detFloorPcts}{1\% and 2\%}  
\newcommand{\detOneDelta}{-0.04}
\newcommand{\detOneFgAcc}{15.3}  
\newcommand{\detOneGFgAcc}{-0.16}  
\newcommand{\detTenDelta}{0.56}  
\newcommand{\detTenGFgIou}{0.0073}  
\newcommand{\detTenGFgIouSem}{0.0024}
\newcommand{\detFullAPnone}{48.9}  
\newcommand{\detFullDelta}{-0.14}
\newcommand{\detProbeOneAP}{0.22}  
\newcommand{\detProbeLiftOne}{23.8}  
\newcommand{\imprintPriorN}{10}
\newcommand{\imprintPrior}{0.40}
\newcommand{\imprintPriorG}{4.23}
\newcommand{\imprintSSLN}{5}
\newcommand{\imprintSSL}{-0.16}
\newcommand{\imprintSSLG}{7.68}  
\newcommand{\imprintRand}{-0.06}
\newcommand{\imprintTap}{-0.18}  
\newcommand{\imprintCombo}{0.49}  
\newcommand{\camBase}{0.253}
\newcommand{\camAux}{0.273}
\newcommand{\camDelta}{0.021}  
\newcommand{\camN}{2{,}000}
\newcommand{\camSigma}{8.6}
\newcommand{\numbersStamp}{ok}  

\makeatletter
\@ifundefined{camSigma}{\@latex@error{tables/numbers.tex is OUT OF DATE
  (\string\camSigma\space missing). Regenerate it with:
  python scripts/make_paper_numbers.py > paper/tables/numbers.tex}\@ehd}{}
\@ifundefined{numbersStamp}{\@latex@error{tables/numbers.tex is OUT OF DATE
  (no \string\numbersStamp). Regenerate it with:
  python scripts/make_paper_numbers.py > paper/tables/numbers.tex}\@ehd}{}
\@ifundefined{auditAllPaired}{\@latex@error{tables/numbers.tex is OUT OF DATE
  (\string\auditAllPaired\space missing). Regenerate it with:
  python scripts/make_paper_numbers.py > paper/tables/numbers.tex}\@ehd}{}
\makeatother

\usepackage{flafter}                                            
\usepackage{algorithmic}
\newcounter{algnum}
\renewcommand{\thealgnum}{\arabic{algnum}}
\newif\ifsubmissionmode
\submissionmodetrue
\ExplSyntaxOn
\RenewDocumentCommand \stmauthors { }
   {
     \group_begin:
     \stmAuthorSetup { type = authors }
     \l_stm_au_setup_tl
     \par \vskip\l_stm_augroup_before_dim
     \l_stm_augroup_align_tl
     \l_stm_augroup_size_tl
     \l_stm_augroup_shape_tl
     \l_stm_augroup_weight_tl
     \color{ \l_stm_augroup_color_tl }
     \bool_if:NTF \g_stm_augr_bool
       { \seq_use:Nn \g_stm_augr_seq { \par } }
       {
         \seq_use:cnnn { g_stm_au\int_use:N\g_stm_augr_int _seq }
         { ,\hspace{0.2em} } { ,\hspace{0.2em} }
         { \hspace{0.2em}and\hspace{0.2em} }
       }
     \par\vskip\l_stm_augroup_after_dim
     \group_end:
   }
\ExplSyntaxOff

\begin{document}
\let\WriteBookmarks\relax
\renewcommand{\topfraction}{0.92}
\renewcommand{\bottomfraction}{0.70}
\renewcommand{\textfraction}{0.06}
\renewcommand{\floatpagefraction}{0.88}
\renewcommand{\dbltopfraction}{0.92}
\renewcommand{\dblfloatpagefraction}{0.88}
\setcounter{topnumber}{4}
\setcounter{bottomnumber}{2}
\setcounter{totalnumber}{6}
\setcounter{dbltopnumber}{3}

\setlength{\textfloatsep}{10pt plus 2pt minus 3pt}
\setlength{\dbltextfloatsep}{10pt plus 2pt minus 3pt}
\setlength{\floatsep}{8pt plus 2pt minus 2pt}
\setlength{\dblfloatsep}{8pt plus 2pt minus 2pt}
\setlength{\intextsep}{8pt plus 2pt minus 2pt}

\makeatletter
\renewcommand\section{\@startsection{section}{1}{\z@}%
    {9pt \@plus 3\p@ \@minus 3\p@}%
    {3\p@}%
    {\sectionfont\raggedright\hst[13pt]}}
\renewcommand\subsection{\@startsection{subsection}{2}{\z@}%
    {6pt \@plus 3\p@ \@minus 2\p@}%
    {.1\p@}%
    {\ssectionfont\raggedright }}
\renewcommand\subsubsection{\@startsection{subsubsection}{3}{\z@}%
    {6pt \@plus 1\p@ \@minus .3\p@}%
    {.1\p@}%
    {\sssectionfont\raggedright}}
\renewcommand\paragraph{\@startsection{paragraph}{4}{\parindent}%
    {3pt \@plus0.01pt \@minus0.01pt}%
    {-6pt}%
    {\ssssparaindent%
     \ssssectionfont\itshape\raggedright}}
\g@addto@macro\normalsize{%
  \setlength{\abovedisplayskip}{4pt plus 2pt minus 2pt}%
  \setlength{\belowdisplayskip}{4pt plus 2pt minus 2pt}%
  \setlength{\abovedisplayshortskip}{2pt plus 1pt}%
  \setlength{\belowdisplayshortskip}{2pt plus 1pt}}
\makeatother

\ExplSyntaxOn
\cs_set:Npn \__reset_fig:
{
  \tl_set:Nx \l_fig_pos_tl { t }
  \tl_set:Nx \l_fig_cols_tl { 1 }
  \tl_set:Nn \l_fig_align_tl { \raggedleft }
  \skip_set:Nn \l_fig_abovecap_skip { 4pt }
  \skip_set:Nn \l_fig_belowcap_skip { 0pt }
  \skip_set:Nn \l_fig_abovefig_skip { 6pt }
  \skip_set:Nn \l_fig_belowfig_skip { 6pt }
}
\cs_set:Npn \__reset_tbl:
{
  \tl_set:Nx \l_tbl_pos_tl { t }
  \tl_set:Nx \l_tbl_cols_tl { 1 }
  \tl_set:Nn \l_tbl_align_tl { \centering }
  \skip_set:Nn \l_tbl_abovecap_skip { 0pt }
  \skip_set:Nn \l_tbl_belowcap_skip { 0pt }
  \skip_set:Nn \l_tbl_abovetbl_skip { 4pt }
  \skip_set:Nn \l_tbl_belowtbl_skip { 0pt }
}
\ExplSyntaxOff

\makeatletter
\renewcommand\section{\@startsection{section}{1}{\z@}%
    {7.5pt \@plus 2\p@ \@minus 2.5\p@}%
    {4\p@}%
    {\sectionfont\raggedright\hst[13pt]}}
\renewcommand\subsection{\@startsection{subsection}{2}{\z@}%
    {5.5pt \@plus 1.5\p@ \@minus 2\p@}%
    {.1\p@}%
    {\ssectionfont\raggedright}}
\renewcommand\subsubsection{\@startsection{subsubsection}{3}{\z@}%
    {5.5pt \@plus 1\p@ \@minus 1.5\p@}%
    {.1\p@}%
    {\sssectionfont\raggedright}}
\renewcommand\paragraph{\@startsection{paragraph}{4}{\parindent}%
    {4pt \@plus 1\p@ \@minus 1.5\p@}%
    {-6pt}%
    {\ssssparaindent\ssssectionfont\itshape\raggedright}}
\makeatother

\makeatletter
\setlength{\@fptop}{0pt}
\setlength{\@fpsep}{8pt plus 2pt minus 2pt}     
\setlength{\@fpbot}{0pt plus 1fil}
\setlength{\@dblfptop}{0pt}
\setlength{\@dblfpsep}{8pt plus 2pt minus 2pt}  
\setlength{\@dblfpbot}{0pt plus 1fil}
\makeatother

\shorttitle{When does fusing knowledge with learned representations pay?}
\shortauthors{A. AlMughrabi et~al.}

\title[mode = title]{When does fusing hand-crafted knowledge with learned
  representations pay? A cost-normalized benchmark of stacking, substitution
  and interference}

\author[1]{Ahmad AlMughrabi}[orcid=0000-0002-9336-3200]
\ead{ahmad.almughrabi@ub.edu}
\cormark[1]

\author[1]{Albert Clop}[orcid=0000-0002-0187-6288]
\ead{albert.clop@ub.edu}

\author[2]{Benjamin Busam}[orcid=0000-0002-0620-5774]
\ead{b.busam@tum.de}

\author[3]{Ricardo Marques}[orcid=0000-0001-8261-4409]
\ead{ricardo.marques@upf.edu}

\author[1]{Petia Radeva}[orcid=0000-0003-0047-5172]
\ead{petia.ivanova@ub.edu}

\affiliation[1]{organization={Universitat de Barcelona},
                city={Barcelona}, country={Spain}}
\affiliation[2]{organization={Technical University of Munich},
                city={Munich}, country={Germany}}
\affiliation[3]{organization={Universitat Pompeu Fabra},
                city={Barcelona}, country={Spain}}
\cortext[1]{Corresponding author.}
\nonumnote{Project page: \url{https://amughrabi.github.io/MomentAux}.
Code, configurations, fixed subset indices, per-run records, and all result tables: \url{https://github.com/GCVCG/MomentAux}}

\begin{abstract}
Fusing prior knowledge with data-driven learning is attractive where data is scarce, yet no controlled account says when it helps, is redundant, or harms. We benchmark one fixed hand-crafted knowledge source, a pinned bank of Gabor targets injected only during training at $\sim$2\% overhead, against data-driven alternatives (SimCLR, SimSiam, DINO, ImageNet transfer, augmentation, learned teachers) under one frozen recipe with fixed subsets: 13 datasets, 9 backbones, 150 to 1.28M images, 32--224\,px, 2.5M--86M parameters ($\computeCells$ classification configurations over $\computeRuns$ runs, plus segmentation and detection transplants). Across the training-time combinations we measure, three outcomes recur (decision-level fusion differs). Different-\emph{currency} sources can stack: the prior composes with DeiT augmentation on attention backbones and is worth $+26$ points to ViT-B/16 at $224$\,px, $+6.7$ at twice that budget. Same-currency sources substitute: against effective self-supervised pretraining, the combination never usefully exceeds the better single source. Fusing at full strength into an already-informed initialization interferes in proportion to what it carries: ImageNet transfer, $-15$ to $-17$ points, removed by a weaker auxiliary weight. Frozen-feature diagnostics measured on each source alone separate these outcomes retrospectively but do not predict them: a rule built on them calls one of nine unseen pairs. At a practitioner's own label budget, the frozen-feature gain predicts the end-to-end gain to within $0.17$ points across 30 cells and seven datasets; the underlying decomposition, $\Delta = G + \readout(\mathrm{base})$, holds in sign on $\auditRate\%$ of testable cells and is called an unseen backbone family's feature gain in advance.
\end{abstract}


\maketitle
\sloppy  

\section{Introduction}\label{sec:intro}

A bank of Gabor filters, fixed before training and free to compute, adds $+26$ accuracy points to a ViT-B/16 at $224$\,px, and $+6.7$ when its budget doubles. Injected exactly the same way into a network that has already had contrastive pre-training, it is worth nothing. Injected at the same strength into one initialized from ImageNet, it costs $15$ to $17$ points. One intervention, one frozen recipe, one code path: three outcomes spanning more than forty accuracy points. The useful question is not which number is representative, because none of them is. It is what distinguishes the three cases, and whether we can determine that \emph{before} training the combination.

This is the general situation, not a curiosity of one prior. Every practitioner who has trained a network on a small dataset faces the same menu of remedies: pre-train on ImageNet \citep{he2019rethinking}, run a self-supervised stage \citep{chen2020simclr,caron2021dino}, augment more aggressively \citep{touvron2021deit}, or inject prior knowledge into the model \citep{vonrueden2023informed}. Each fuses an external source of information (weights learned elsewhere, invariances distilled from augmented views, hand-crafted structure) with whatever the labeled data can teach on its own. What the field lacks is not remedies but a controlled account of how these sources \emph{combine}: when does adding a second source help, when is it redundant with what the first already supplies, and when does the fusion actively destroy value? These three outcomes, which we call \emph{stack}, \emph{substitute} and \emph{interfere}, are usually discovered anecdotally, one paper and one setting at a time, under training recipes that differ in a dozen
uncontrolled ways.

Our answer is that sources trade in identifiable \emph{currencies}, and that what a source is worth depends on which currency is scarce; how modern the method is matters far less. Two sources add only when their currencies differ, a necessary condition: whether the possible gain is realized also depends on the architecture and on the strength at which the second source is applied. This is not a metaphor we impose after the fact: each source's currency is measurable, on frozen features, from each source \emph{alone}, which is what separates the three outcomes above and why the same free prior can be the most valuable intervention we measure in one setting and the most damaging in another. Turning that separation into a rule that \emph{predicts} an untrained combination is a further step, and one we take and report as unsuccessful (Sec.~\ref{sec:procedure}).

This paper treats the question as a measurement problem. We build a single instrument (one frozen training recipe, fixed data subsets shared by every run, and a numerically pinned bank of Gabor and moment spectral filters \citep{luan2018gaborcnn,bruna2013scattering}) and use it to measure one deliberately austere source of prior knowledge against representative data-driven alternatives at matched or declared compute. The prior, which we call \emph{MomentAux}, could hardly cost less: a fixed bank of oriented-energy targets, injected only during training through a small auxiliary regression head whose loss weight decays to exactly zero, costing ${\sim}2\%$ extra training compute and \emph{nothing} at inference. Against it, we field SimCLR \citep{chen2020simclr}, SimSiam \citep{chen2021simsiam} and DINO \citep{caron2021dino} initializations (each ${\sim}2\times$ compute), ImageNet transfer \citep{he2019rethinking}, DeiT-strength augmentation \citep{touvron2021deit}, learned FitNets-style teachers \citep{romero2015fitnets}, and targets built from histograms of oriented gradients (HOG) \citep{dalal2005hog}. The resulting grid spans 13 datasets across six visual domains, nine backbone families from ResNet-18 \citep{he2016resnet} to ViT-B/16 \citep{dosovitskiy2021vit}, data scales from 150 to 1.28M images, resolutions from $32$ to $224$\,px, and \computeCells{} configurations trained over \computeRuns{} runs, each with a retained run record.

A grid of this size threatens to be a table dump. What rescues it is an empirical regularity, a law only in the bounded, measured-regime sense of Sec.~\ref{sec:lawform}, found, not assumed, and then tested by registering its predictions in advance and trying to break them. Writing $\Delta$ for an intervention's end-to-end accuracy gain and $G$ for the gain a linear evaluation \citep{alain2016understanding} reads from its frozen features under identical probing, we find across the grid that
\begin{equation}
  \Delta \;=\; G \;+\; \readout(\mathrm{base}),
  \label{eq:law}
\end{equation}
Run the evaluation at a practitioner's own label budget, and $G$ predicts $\Delta$ to $0.17$ points across 30 cells and seven datasets, so the intervention's effect is predominantly representation-side rather than readout-side; the sign law below is its full-label corollary, resting on far more cells (Sec.~\ref{sec:law}). Under that fixed protocol, the \readout{} term depends, to a good approximation, only on the \emph{baseline accuracy} of the cell: negative below a crossing at ${\sim}32$--$40\%$ accuracy, near zero at the crossing, and small and positive above it. The sign holds in $\auditRate\%$ of the \auditResolvable{} cells where measurement uncertainty allows it to be tested and in $\auditBelowRate\%$ of those below the crossing, and its quantitative form survived our hardest pre-registered tests: it predicted the feature gain of a backbone family it had never seen (Swin) from baseline heights alone, and at ImageNet scale it tied end-to-end gains to feature gains within $1.1$ accuracy points on five of six backbone pairs (Sec.~\ref{sec:scale}).

The law converts the grid into an explanation, and it is where currency stops being a metaphor and becomes the term $G$: heavy augmentation supplies nuisance-invariance, contrastive pre-training much the same invariance at greater expense, ImageNet weights mature photographic features, and the moment prior oriented-energy structure. Same-currency fusions substitute: the prior and SimCLR each render the other nearly worthless, and their combination sits near the better single, from about a point above it to $2.4$ points below on the convolutional off-selection cells. Different-currency fusions can stack, the prior's gain growing $1.4$--$2.4\times$ under DeiT-strength augmentation on the attention backbones, though on two convolutional populations off the selection set the same pairing stacks at one cell of six: differing currencies are necessary, not always sufficient. And fusing a shaping prior at full strength into an initialization that already carries mature features interferes in proportion to what the initialization supplied, a cost that lands almost entirely on the features themselves and that a weaker
auxiliary weight removes (Sec.~\ref{sec:fusion}).

Two headline findings illustrate what the instrument can resolve. First, small vision transformers carry a large feature deficit that the prior fills: the gain reaches $+13.0$ points for ViT-S/16 and $+26.0$ for ViT-B/16 on ImageNet-100 at $224$\,px under the standard DeiT recipe at a shared $100$-epoch budget, and persists at $+3.2$ after 1.28 million images where the ResNet baseline is exactly neutral ($+0.04$). The budget is part of that claim, so we doubled it on both models: the gains become $+4.52 \pm 0.18$ for ViT-S and $+6.71 \pm 0.90$ for ViT-B, a quarter to a third of the headline figures but \emph{growing} with model scale still, at $2.4$ standard errors (Sec.~\ref{sec:scale}). Second, the fashionable remedy is not always the right one: under plain augmentation, SimCLR beats the prior across the mid-data band on photographic datasets, but under the DeiT recipe the comparison flips at matched ${\sim}2\times$ cost, at every CIFAR-100 fraction and at all but a statistical tie at Tiny-ImageNet's largest, because the recipe supplies the invariance SimCLR was paid that compute to learn. At $5\times$ the flip narrows to the higher-data band, where the prior still wins on both populations, while the comparator takes the scarcest cells (Sec.~\ref{sec:budget}); the ordering tracks the data fraction, a statement about budget and regime, not a ranking of methods.

\paragraph{Contributions.}
\begin{enumerate}
  \item \textbf{An instrument, not just experiments} (Sec.~\ref{sec:instrument}): a frozen recipe with fixed subsets, pinned filter banks, declared cost normalization, and pre-registered predictions with falsifiers; every headline table is regenerable from the released artifacts.
  \item \textbf{The grid} (Sec.~\ref{sec:grid}): $\computeCells$ configurations, $\computeRuns$ runs, comparing a free spectral prior against SSL, transfer, augmentation, and learned teachers across 13 datasets, 9 backbones, and 150 to 1.28M images.
  \item \textbf{The law, led by its budget-matched form} (Sec.~\ref{sec:law}): at a practitioner's own label budget, the frozen-feature gain predicts the end-to-end gain to $0.17$ points over 30 cells and seven datasets, a representation-side diagnostic whose magnitude tracks the observed end-to-end gain. The full-label sign law of $\Delta = G + \readout(\mathrm{base})$ is its explanatory corollary, validated from CIFAR scale to ImageNet scale, with its boundary cases reported as findings, including the one patterned failure where features improve at sufficiency while accuracy does not follow (Sec.~\ref{sec:exceptions}) and the ImageNet-scale puzzle of three envelope shapes on identical data (Sec.~\ref{sec:scaleenvelope}).
  \item \textbf{A fusion taxonomy with mechanism} (Sec.~\ref{sec:fusion}): stack, substitute, or interfere, decided by whether the fused sources carry the same information currency, with the interference conditioned on shaping strength, and verified on the feature side, not just end-to-end.
  \item \textbf{A procedure, stated and then falsified} (Sec.~\ref{sec:procedure}): because each source's currency is measurable from that source \emph{alone}, we state the taxonomy as an algorithm over quantities available before the combination is trained, and test it on thirteen pairs whose combination did not yet exist, with every input estimated on validation data and the calls recorded first. It calls one of nine correctly, so we withdraw the predictive claim, report which predicate fails and why, and distil the measured grid into a regime-indexed guide with attached costs (Sec.~\ref{sec:guide}).
\end{enumerate}

\section{Related work}\label{sec:related}

\looseness=-1 We group prior work by the mechanism through which it supplies information to a data-limited network, the axis along which we organize our comparison: hand-crafted structure in the forward path; auxiliary and distillation targets; self-supervised and transfer initializations; augmentation recipes; small-data results specific to attention; and comparative benchmarks. Sec.~\ref{sec:fusiontheory} then places our three outcomes against the fusion literature's own account, and Sec.~\ref{sec:gap} tabulates which properties a controlled fusion study requires and which each line provides.

\paragraph{Hand-crafted structure inside learned networks.} Fixed oriented filters predate deep learning as image descriptors \citep{dalal2005hog}. Scattering networks showed that cascades of fixed wavelets rival learned features in small-data regimes \citep{bruna2013scattering}, and Gabor-structured convolutions have been used to regularize or cheapen early layers \citep{luan2018gaborcnn}. These lines place the structure \emph{in the forward path}, where it occupies input bandwidth permanently; our baselines reproduce their known mid-data penalty band, which no filter choice escapes (Sec.~\ref{sec:ablations}). Our central object differs precisely in that the structure lives in a training-only auxiliary target whose influence decays to zero, so abundant data overrides the prior instead of paying for it.

\paragraph{Auxiliary targets, deep supervision, and distillation.}
\looseness=-1 Regressing intermediate features onto targets is the mechanism of FitNets \citep{romero2015fitnets} and deep supervision \citep{lee2015deeply}; HOG targets power MaskFeat's self-supervised pre-training \citep{wei2022maskfeat}. What distinguishes our setting is the \emph{fixedness and cost} of the target: no teacher network is trained, no pre-training stage is run, and the target is a closed-form function of the input. Our controls quantify how much this matters: a learned same-data teacher moves accuracy by at most $0.73$ points anywhere on an eleven-point envelope, HOG targets recover a third to a half, and a random fixed target of identical shape is beaten by $+0.8$ to $+4.2$ points on all four populations where we ran that control (Table~\ref{tab:controls}, Sec.~\ref{abl:target}), so the gain traces to the moment structure itself, not to auxiliary regression as a mechanism.

\paragraph{Self-supervised learning (SSL) as initialization.} Contrastive and
self-distilling pre-training \citep{chen2020simclr,chen2021simsiam,caron2021dino} is the modern default source of ``free'' features. \citet{ericsson2022why} measure \emph{which} invariances such models learn and show that fusing representations with complementary invariances improves transfer, which is close in spirit to our different-currency case; they demonstrate the complementarity, where we attempt a prospective, cost-normalized decision rule for complementarity, redundancy and interference. Little is measured about the small-data, from-scratch regime we target, and each method is evaluated here
under its own published recipe for this scale. Our grid contributes three facts to that gap: negative-free SimSiam learns nothing essentially at $2.5$--$25$k images under its published CIFAR recipe, though four times that budget recovers up to $+4.3$ and we report it as a property of the budget, not of the method; SimCLR is strong on photographic statistics but loses to a free domain-agnostic prior on satellite imagery in the low-data band; and the SSL-versus-prior comparison inverts under
a modern supervised recipe at matched cost, because the augmentation stack substitutes for the invariance SSL sells (Secs.~\ref{sec:fusion} and \ref{sec:budget}).

\paragraph{Transfer learning.} ImageNet initialization remains the strongest single intervention on photo-like data, with known limitations under domain shift and with training from scratch competitive given enough data \citep{he2019rethinking}. We use transfer both as a comparator and as a test of fusion: injecting the spectral prior at full auxiliary strength \emph{on top of} ImageNet weights is the one configuration in our grid that is destructive everywhere, and linear probing shows the destruction is feature-side, early shaping erasing the mature structure the initialization supplied, in proportion to how much that structure was worth (Sec.~\ref{sec:fusion}).

\paragraph{Small-data vision transformers.} That ViTs underperform without large data or heavy regularization is well documented \citep{dosovitskiy2021vit,touvron2021deit}, and a family of remedies exists: distillation tokens \citep{touvron2021deit}; architectures that reintroduce the hierarchical, local inductive bias attention lacks, such as Swin \citep{liu2021swin}, which we include as a backbone for that reason; and auxiliary self-supervised objectives trained jointly with the classifier, of which \citet{liu2021efficientvit} is the closest antecedent to this work: a dense relative-localization task added for exactly this deficit, differing from ours in that its target is computed from the images themselves, where ours is pinned in advance. Our contribution to this literature is quantitative and comparative: we measure the deficit as a feature quantity ($G \approx 13$--$15$ points for ViT-tiny across $2.5$--$12.5$k images, roughly $2.3\times$ the ResNet-18 value at matched cells), show a fixed spectral target fills it at lower cost than contrastive pre-training and more effectively than DeiT augmentation alone, and show the deficit \emph{grows} with model scale at full data (Sec.~\ref{sec:scale}), which argues it reflects data-hunger, not
small-model capacity.

\paragraph{Large-scale comparative benchmarks.} \emph{Battle of the Backbones} \citep{goldblum2023battle} compares a broad suite of pretrained backbones across tasks over more than 1{,}500 training runs, and finds supervised convolutional pre-training still competitive; \citet{marks2025closer} show that self-supervised benchmarking conclusions depend heavily on the evaluation protocol. Both benchmark \emph{pretrained backbones}, a checkpoint someone else produced, and neither normalizes by training cost nor asks what happens when two sources are combined. Our unit is instead the \emph{intervention applied to a fixed recipe}, which makes cost normalization and the stacking question answerable.

\paragraph{Spectral structure inside transformers.} A parallel line injects spectral structure into attention architectures: FViT \citep{shi2026fvit} adds a \emph{learnable} Gabor filter to focus attention across scales and orientations, and scattering-based token mixers do the same with wavelet decompositions. These are architectural changes that persist at inference, and each is evaluated as a model proposal against published baselines. Our object is the complement: the filters are fixed, not learned; they never enter the forward path, and the deployed network is unchanged, which lets us ask the benchmark question of when adding them pays instead of the model question of whether one architecture beats another.

\subsection{Relation to fusion theory}\label{sec:fusiontheory} The outcomes we measure are not new categories, and we set out the prior claims before our own. \citet{durrantwhyte1988sensor} classified the interactions between disparate information sources as \emph{cooperative}, \emph{competitive} and \emph{complementary}; the surveyed form now standard in the field \citep{khaleghi2013multisensor} states the triple as complementary, redundant and cooperative, where complementary sources describe different aspects of the target, redundant sources describe the same aspect, and cooperative sources jointly yield what neither yields alone. Our \emph{stack} is a complementary combination, and our \emph{substitute} is a redundant combination. We adopt that vocabulary here and use our own terms only as informal labels.

\citet{dasarathy2001what} organizes the field's questions as what, where, why, when, and how to fuse, and observes that the \emph{when} has received the least attention because it cannot be answered from the architecture. This paper takes up that question: one source is hand-crafted, and the other is learned from the data. \citet{smirnov2019knowledge} surveys the patterns by which knowledge and data are combined \citep{khaleghi2013multisensor} and catalogs the same three outcomes we recover; what that survey records as patterns a designer chooses, we measure as outcomes. The field's benchmarks have generally compared \emph{algorithms} for a fixed fusion problem; ours compares \emph{sources} under a fixed algorithm, so what varies is the knowledge injected, not the method injecting it.

One difference is substantive, not terminological, and it sharpens what a representation-learning setting adds. In sensor fusion, redundancy is a \emph{virtue}: two instruments measuring the same quantity reduce variance and buy fault tolerance. In ours it is waste, because performance is bounded by the information the sources share, and supplying that information twice improves nothing. Redundancy pays for reliability and not for representation quality, and a taxonomy built for the first case does not transfer its sign to the second.

Our third outcome, in which a source destroys what its partner supplied, is likewise not new, and has at least three names: \emph{catastrophic fusion} \citep{movellan1997catastrophic}, negative transfer \citep{wang2019negative}, and the modality competition of \citet{wang2020multimodal}, who report that the best unimodal network often beats the jointly trained multimodal one despite the latter receiving strictly more information. What we add is not the phenomenon but its magnitude law: the damage scales with how much the displaced source had contributed, and vanishes exactly where it had contributed nothing.

The formal account of our informal ``currency'' is partial information decomposition \citep{williams2010nonnegative}, which splits what two sources tell us about a target into unique, redundant and synergistic atoms, and which \citet{liang2023quantifying} have already carried into deep multimodal learning with estimators that scale and with a-priori model selection as the application. We do not compute it. Our measurement is a linear-evaluation gap on frozen features: inexpensive enough for \computeCells{} cells, ordinal, not an information quantity, and blind to anything not linearly decodable. Read our result as an empirical instance of the decomposition those authors formalize, not as an estimate of it.

Two further connections position the predictor, not the taxonomy. Predicting a downstream outcome from an inexpensive measurement, without training the combined system, is the programme of transferability estimation \citep{nguyen2020leep}, which scores a single pre-trained source; ours asks the same question of a \emph{pair}. And the challenge that programme raises for us is direct: \citet{newell2020useful} report that linear evaluation does not correlate with fine-tuning performance in general. Our defence is narrow and empirical, not a rebuttal. We do not use the linear evaluation to rank absolute quality; we use the \emph{gap} between two arms probed under an identical protocol, on checkpoints that differ in one intervention, and we report how well that gap predicts the paired difference (Sec.~\ref{sec:law}) rather than assuming it does.

Finally, our knowledge source sits where \citet{vonrueden2023informed} make precise: informed learning requires two independent sources of information, data and prior knowledge, which is the reading of ``two sources'' under which this study is a fusion study. Ours is scientific knowledge about early vision, represented as a pinned filter bank, integrated into the learning algorithm; the nearest method in their taxonomy is informed pre-training, which condenses knowledge into prototypes for initialization, whereas ours enters as a decaying auxiliary target. That survey catalogs \emph{where}
knowledge can be injected; what a practitioner must still decide is \emph{when} injecting it is worth anything given what the data and the available pre-trained artifacts already supply. That is the question this benchmark answers, and we claim only that we found no controlled, compute-declared answer.

\subsection{The gap this paper closes}\label{sec:gap}

Table~\ref{tab:gap} summarizes the position. Each column is a property that a question about \emph{fusing} information sources requires, and no existing line supplies all five at once.

\begin{table}[pos=htbp]
\caption{Where the literature stands on the five properties a fusion question needs, and what this work adds. \cmark{} = present; \pmark{} = partial or incidental; \xmark{} = absent.}
\label{tab:gap}
\centering\scriptsize
\setlength{\tabcolsep}{2.6pt}
\zebra{2}
\begin{tabular}{P{0.58\columnwidth}ccccc}
\toprule
Line of work & Rec. & Cst. & Fea. & Cmb. & Prd. \\
\midrule
Fixed spectral filters in the forward path
  \citep{bruna2013scattering,luan2018gaborcnn,shi2026fvit}
  & \pmark & \xmark & \xmark & \xmark & \xmark \\
Auxiliary and distillation targets
  \citep{romero2015fitnets,lee2015deeply,wei2022maskfeat}
  & \pmark & \xmark & \pmark & \xmark & \xmark \\
Self-supervised pre-training
  \citep{chen2020simclr,chen2021simsiam,caron2021dino}
  & \xmark & \xmark & \cmark & \xmark & \xmark \\
Transfer learning studies
  \citep{he2019rethinking}
  & \pmark & \xmark & \pmark & \xmark & \xmark \\
Small-data ViT recipes
  \citep{touvron2021deit}
  & \xmark & \pmark & \xmark & \pmark & \xmark \\
Backbone and self-supervised benchmarks
  \citep{goldblum2023battle,marks2025closer}
  & \pmark & \xmark & \cmark & \xmark & \xmark \\
Classical fusion taxonomy
  \citep{durrantwhyte1988sensor,williams2010nonnegative}
  & \xmark & \xmark & \xmark & \cmark & \pmark \\
\midrule
\textbf{This work} & \cmark & \cmark & \cmark & \cmark & \cmark \\
\bottomrule
\rowcolor{white}\multicolumn{6}{@{}p{0.97\columnwidth}@{}}{\tiny Rec.: one frozen recipe. Cst.: cost-normalized, comparators reported against a declared training-compute multiple, not each method's own chosen budget. Fea.: feature-side, the effect measured on frozen representations, not only end-to-end. Cmb.: combinations, what happens when two sources are applied together. Prd.: a predictive model of the outcome; ours holds for single-source outcomes, while the pairwise fusion rule built on it failed its prospective test (Sec.~\ref{sec:procedure}).} \\
\end{tabular}
\end{table}

The first column matters because comparisons made under each method's own preferred recipe cannot attribute a difference to the method; the backbone benchmarks above inherit whatever recipe produced each checkpoint. We instead hold one recipe and one set of fixed image indices and vary only the intervention.

The second is rarely reported: self-supervised pre-training roughly doubles the training compute of the baseline it improves, yet that factor almost never appears beside the accuracy it buys, leaving ``method A beats method B'' unanswerable at a fixed budget. Every comparator here carries a declared multiple.

The third and fourth columns are what turn a table of numbers into an explanation. Linear probing is standard practice for evaluating self-supervised representations, so the feature-side view exists in that literature; what is missing is applying it to a \emph{paired} comparison, the same intervention measured end-to-end and on frozen features at once, which is what exposes redundancy between two sources. Combinations are the least studied of all:
What happens when a prior is applied on top of augmentation, of self-supervised weights, or of a pretrained initialization is precisely the fusion question, and we could find no controlled study that measures all three.

The fifth column is the one we regard as the paper's distinguishing contribution. Benchmarks describe; a law predicts. Our decomposition $\Delta = G + \readout(\mathrm{base})$ was registered in advance and used to call all three classes of unmeasured quantities listed above before those measurements existed (Secs.~\ref{sec:law} and~\ref{sec:scale}). A related decomposition of learning dynamics into representation and decoding components has been proposed for grokking \citep{liu2022grokking}; to our knowledge, no comparable predictive account exists for data-efficiency interventions.

\section{The instrument}\label{sec:instrument}

\begin{figure*}[pos=htbp]
\centering
\includegraphics[width=\linewidth]{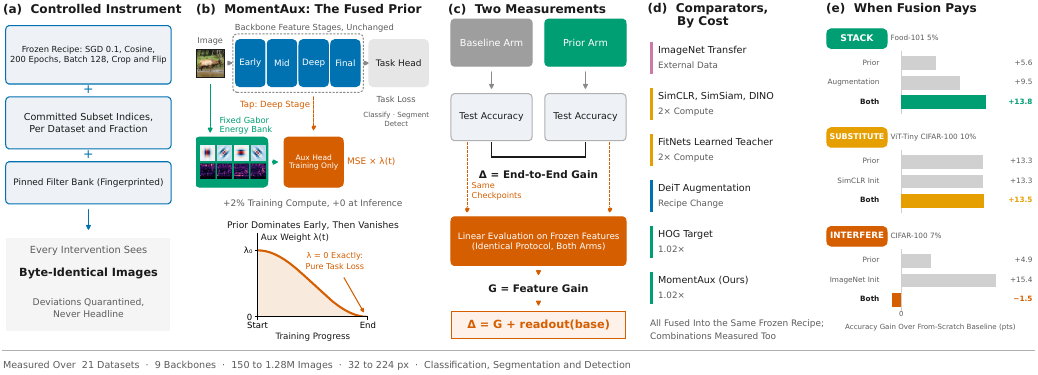}
\caption{The study in one picture. \textbf{(a)} One frozen recipe and fixed subset indices, so every intervention sees byte-identical images; deviating cells are marked diagnostic and never enter headline tables. \textbf{(b)} MomentAux: a pinned bank of oriented-energy targets regressed from a tap at the third of four feature stages, during training only, with $\lambda$ decaying to exactly zero, so the deployed network is byte-identical to the baseline. \textbf{(c)} Each cell yields the end-to-end gain $\Delta$ and the frozen-feature gain $G$, whose difference is the \readout{} term of Eq.~\ref{eq:law}. \textbf{(d)} The comparator ladder, ordered by declared training cost; all are fused into the same recipe, and pairwise combinations are measured too. \textbf{(e)} One measured cell per fusion outcome, each source alone and then both, as gain over the shared baseline: the fused bar clears both singles, sits level, or is dragged back across zero. Interference is shown at full auxiliary strength, the condition it depends on (Sec.~\ref{sec:taxsub}); the frozen features move the same way in all three cases (Sec.~\ref{sec:fusion}).}
\label{fig:method}
\end{figure*}

\looseness=-1 The benchmark's claims rest on the comparisons being controlled, so we describe the controls first; Fig.~\ref{fig:method} shows the mechanism they govern and the three fusion outcomes it produces. Three controls fix what is compared: the frozen recipe and fixed subsets that make any two cells comparable (Sec.~\ref{sec:recipe}); the knowledge source, a pinned spectral bank and the auxiliary head that injects it (Sec.~\ref{sec:momentaux}); and the competing interventions with their cost normalization (Sec.~\ref{sec:comparators}). Two fix how it is measured: the linear evaluation of frozen features that supplies $G$ (Sec.~\ref{sec:probe}) and the seed and uncertainty protocol (Sec.~\ref{sec:stats}). The last bounds what the grid can claim: how the configuration was selected and how that leaves it advantaged (Secs.~\ref{sec:selection} and~\ref{sec:parity}), and the pre-registration that makes its predictions falsifiable (Sec.~\ref{sec:prereg}).

\subsection{Frozen recipe and fixed subsets}\label{sec:recipe}
Every headline cell trains with one identical stochastic-gradient recipe, whose numeric settings are collected in Sec.~\ref{sec:compute}. What matters here is their status, not their values: the recipe is frozen in the strict sense that no cell tunes it, which makes any two cells
comparable. Configurations that must deviate- AdamW \citep{loshchilov2019adamw} for architectures the SGD recipe cannot train, reduced epochs at ImageNet scale, added augmentation stacks- are marked as diagnostic in their configuration name, which the training code enforces, and never enter a headline table. Within such cells, both arms share the deviation, so the paired difference remains valid even though the absolute numbers are not comparable to frozen-recipe cells.

Data fractions are not sampled at run time. For each dataset $\mathcal{D}$ and fraction $p$, we precompute a stratified index set $S_{\mathcal{D},p}$ by drawing, per class, a seeded permutation truncated to $p\%$, and fix it once for the whole study. Every intervention therefore consumes byte-identical images, so any accuracy difference is attributable to the intervention, not the draw. Because the recipe fixes epochs, not steps, the optimization budget scales with $|S_{\mathcal{D},p}|$; we treat data and compute as jointly varied along this axis and say so wherever it matters.

\looseness=-1 Two further controls proved necessary in practice. Dataloader worker count is part of the reproducibility contract: PyTorch seeds each worker's augmentation RNG from a base seed plus worker index, so changing the count re-draws the augmentation stream. The change is unbiased; it acts like a different augmentation seed, but it breaks byte-level reproducibility, so we pin the count per cell and record it per run. One headline pair crosses that boundary, and we name it instead of letting the contract imply otherwise: the CIFAR-10 $2\%$ cells run eight workers on the baseline arm and two on the prior arm, so their augmentation streams differ. The paired difference stands, since the redraw is unbiased, but that cell is not byte-reproducible from the recipe alone, and its two arms are not seed-matched in the way every other headline pair is. The $1\%$ pair, which Sec.~\ref{sec:ownenvelope} compares it against, is matched at eight. Second, the filter bank is \emph{pinned}: unit tests fingerprint its numerical values, so no measurement can silently drift with a library version.

\subsection{The prior: MomentAux}\label{sec:momentaux}
\begin{figure}[pos=htbp]
\centering
\includegraphics[width=1.0\linewidth]{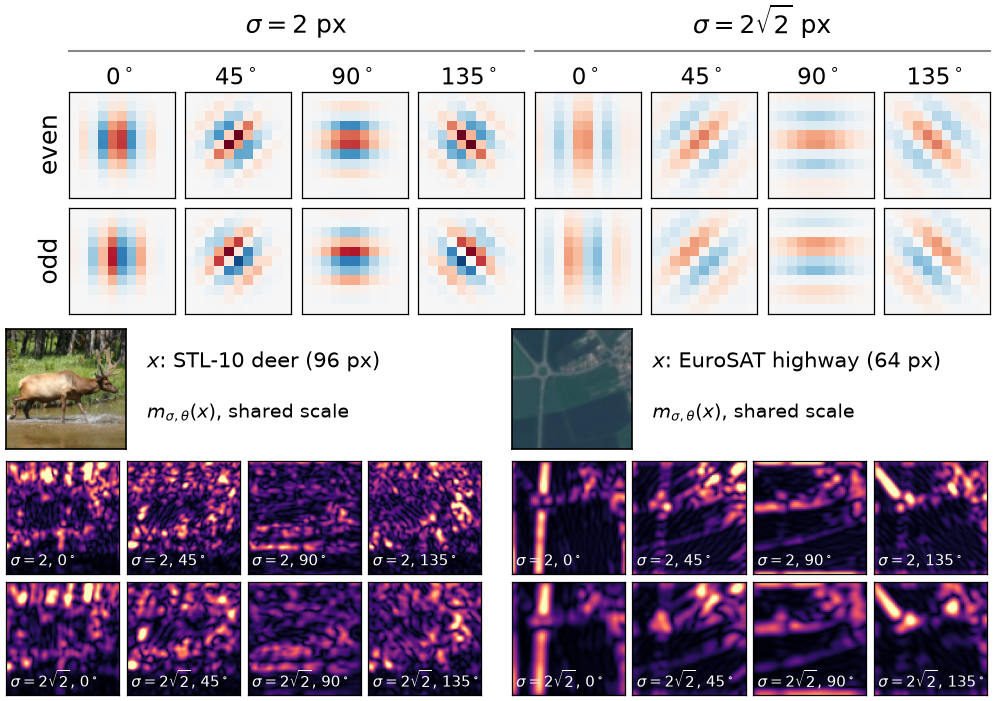}
\caption{The knowledge source, in full, and what it computes (an illustration, not evidence). Top: the eight
complex Gabor quadrature pairs of Eq.~\ref{eq:gabor}, two scales $\sigma\in\{2, 2\sqrt{2}\}$ crossed with four orientations, even and odd components. Below: a photographic (STL-10) and a satellite (EuroSAT) test image, each with its eight magnitude-response maps of Eq.~\ref{eq:energy} from the pinned bank under the study's own calibration, one color scale per example; each sample is chosen deterministically as the test image on which every orientation channel dominates somewhere. Trunks and limbs light the STL-10 orientations, roads and field boundaries the EuroSAT ones, the same fixed kernels in both domains. The bank is fixed before any training, identical everywhere, and fingerprinted by a regression test.}
\label{fig:bank}
\end{figure}

\paragraph{The knowledge source.} The bank consists of $N{=}8$ complex Gabor quadrature pairs, two spatial scales $\sigma \in \{2, 2\sqrt{2}\}$ crossed with four orientations $\theta \in \{0, \pi/4, \pi/2, 3\pi/4\}$; each of the form
\begin{equation}
  g_{\sigma,\theta}(\mathbf{u}) \;=\;
  e^{-\lVert \mathbf{u}\rVert^{2}/2\sigma^{2}}\,
  e^{\,i\,k_\sigma\, \mathbf{u}^{\!\top} (\cos\theta, \sin\theta)},
  \label{eq:gabor}
\end{equation}
with $k_\sigma$ set so each envelope carries a comparable number of cycles (Fig.~\ref{fig:bank}). The target is the \emph{phase-invariant magnitude} of the response to the greyscale input $x$,
\begin{equation}
  m_{\sigma,\theta}(x)[u] \;=\; \bigl|\, (x * g_{\sigma,\theta})[u] \,\bigr|,
  \label{eq:energy}
\end{equation}
taken pointwise at every spatial location $u$, then stacked over the $N$ pairs and average-pooled to the tapped stage's spatial resolution, giving $\mathbf{m}(x)\in\mathbb{R}^{N\times h\times w}$. Magnitude, not raw filter response, is the operative choice: the ablation in Sec.~\ref{sec:ablations} shows oriented \emph{edges} are a poor target ($-0.09$), while their phase-invariant energy is the best one \looseness=-1 measured ($+2.71$ at the same cell).

\paragraph{The fusion mechanism.} Let $f_3(x)$ be the activations of the backbone's third feature stage (\texttt{layer3} in a ResNet, \texttt{stages.2} in ConvNeXt, the matching block group in Swin; architectures whose depth does not split into four stages tap the matching depth fraction, block~8 of 12 in ViT and block~3 of 7 in MobileNetV3), and $W$ a $1{\times}1$ convolution mapping them to $N$ channels. Training minimizes
\begin{equation}
  \begin{aligned}
    \mathcal{L}(t) &\;=\; \mathcal{L}_{\mathrm{CE}}
      \;+\; \lambda(t)\,
      \bigl\lVert\, W f_3(x) - \mathbf{m}(x) \,\bigr\rVert_2^2,\\[2pt]
    \lambda(t) &\;=\; \tfrac{\lambda_0}{2}
      \Bigl(1+\cos\tfrac{\pi t}{T}\Bigr),
  \end{aligned}
  \label{eq:loss}
\end{equation}
so the prior dominates early and vanishes by the final epoch. Three properties follow directly, and each is load-bearing for the paper's claims. (i) Because $\lambda(T)=0$ \emph{exactly}, the prior contributes no gradient over the final epochs, so nothing has to be tuned to switch it off. It does \emph{not} follow that the envelope's right-hand end is a consequence of the schedule: at a matched optimizer budget the right flank does not appear at all (Secs.~\ref{sec:ownenvelope} and~\ref{sec:dataopt}). (ii) $W$ is discarded after training, so the deployed network is byte-identical to the baseline: same parameters, same floating-point operations (FLOPs), zero inference cost. (iii) The measured training overhead is ${\sim}2\%$, against $100\%$ for a self-supervised pre-training stage.

\paragraph{The scale degeneracy and its fix.} The auxiliary objective in Eq.~\ref{eq:loss} is invariant under replacing $f_3$ by $f_3/c$ and $W$ by $cW$, which lets SGD satisfy it by collapsing the tapped features and inflating the head instead of by learning structure. We traced this directly: on ResNet-50 the tapped standard deviation fell from $0.596$ to $0.051$ while $\lVert W\rVert$ grew from $1.64$ to $11.2$ and cross-entropy stalled at chance. Projecting $\lVert W \rVert$ back to its initialization norm after each step removes the degenerate direction, converting a bistable failure into a stable $+3.9$ (Sec.~\ref{sec:ablations}); we adopt it always-on. One configuration, $\lambda_0{=}1.0$, magnitude target, third-stage tap, head-norm on, transplants across every dataset and backbone without retuning, and we use it verbatim as the \emph{reference configuration} unless stated.

One qualification on that term. The CIFAR-10, CIFAR-100 and Tiny-ImageNet envelopes of Table~\ref{tab:envelope} predate the head-norm decision and run without it; Table~\ref{tab:domainenv} and the later backbone sweeps have it on. On ResNet-18, the setting is free to mildly positive: with head-norm, the CIFAR-100 cells give $+5.51$, $+3.05$, and $+0.82$ at $5$, $15$, and $25\%$ against Table~\ref{tab:envelope}'s $+5.15$, $+2.55$, and $+0.16$. We disclose this instead of re-pointing the table to the stronger cells: choosing between two measured configurations after seeing both is the practice this benchmark makes visible.

\subsection{Comparators and cost normalization}\label{sec:comparators}
We normalize by declared training cost relative to the baseline: MomentAux ${\approx}1.02\times$; SimCLR, SimSiam and DINO initializations $2\times$ (a full pre-training stage, then the frozen recipe); DeiT-strength augmentation
\citep{touvron2021deit} ${\approx}1\times$ compute but a modern recipe change; FitNets teachers $2\times$; HOG targets ${\approx}1.02\times$; ImageNet transfer amortized, but flagged throughout as the one comparator with an
external-data advantage.

Every self-supervised stage runs on the same subset of images of the cell it initializes, never on extra data, for 200 epochs (800 in the raised-budget arms of Sec.~\ref{sec:budget}) at batch 128 with that method's own published view augmentations, and is followed by the unchanged frozen recipe. Because a weak comparator would flatter our result, we test SimCLR's configuration explicitly instead of assuming it: strengthening its contrastive views to DeiT strength makes it \emph{worse} (Sec.~\ref{sec:fusion}), so its standard views are its strongest available
configuration at this scale.

\subsection{Measuring features: linear evaluation}\label{sec:probe} The law needs a feature-side quantity, and we obtain it by \emph{linear evaluation} of frozen representations, the standard linear-probe protocol \citep{alain2016understanding}. For each trained checkpoint we freeze the network, extract penultimate features $z = \mathrm{pool}(f_L(x))$ for the full training split under the deterministic (no-augmentation) transform, and fit a
multinomial logistic regression using the limited-memory BFGS optimizer (L-BFGS) and fixed hyperparameters:
\begin{equation}
  \min_{V}\; \tfrac{1}{n}\!\sum_{i} \mathrm{CE}\bigl(Vz_i, y_i\bigr)
    \;+\; 10^{-4}\lVert V\rVert_2^2.
  \label{eq:probe}
\end{equation}
Only the features differ between arms; everything else is identical. We define the feature gain as the linear-evaluation gap $G = \mathrm{acc}_{\mathrm{lin}}^{\mathrm{aux}} - \mathrm{acc}_{\mathrm{lin}}^{\mathrm{base}}$, and the \readout{} term as the residual $\Delta - G$.

Linear evaluations are diagnostics, not headline numbers; they read labels the cell never saw. Two scope rules, both learned the hard way and enforced throughout: an evaluation is interpretable only while it holds substantially more labels than the cell trained on, so at 100\% cells we report the aux-vs-baseline gap but do not split it into $G$ and \readout; and at ImageNet64 scale, where an L-BFGS fit over $1.28$M rows is impractical, evaluations use fixed per-class budgets whose $G$ values are compared only to other same-budget
evaluations.

\subsection{Statistical protocol}\label{sec:stats}
Each cell runs ${\geq}3$ seeds, ten for headline envelopes; eighteen legacy exploration cells of the \computeCells{} carry fewer. For a paired cell, we report $\Delta$ as the difference of arm means with the standard error of the mean (SEM) of that difference.

\looseness=-1 The \readout{} term needs more care than a difference of two independent measurements, because $\Delta$ and $G$ are not independent: the linear evaluation probes the very checkpoints whose end-to-end accuracy gives $\Delta$, so across seeds the two move together and $\mathrm{Cov}(\Delta, G) >
0$. Propagating the residual as $\sqrt{\mathrm{SEM}(\Delta)^2 + \mathrm{SEM}(G)^2}$ would therefore \emph{overstate} its uncertainty; on this grid it does so by a median factor of $1.7$. We instead form the \readout{} \emph{per seed} and take the standard error of that quantity directly,
\begin{equation}
  \readout_s =
  \bigl(a^{\mathrm{aux}}_s - e^{\mathrm{aux}}_s\bigr) -
  \bigl(a^{\mathrm{base}}_s - e^{\mathrm{base}}_s\bigr),
  \label{eq:sem}
\end{equation}
where $a_s$ and $e_s$ are seed $s$'s end-to-end and linear-evaluation accuracies, and only seeds present in all four arms contribute. A cell is called \emph{resolvable} when $|\readout| > 2\,\mathrm{SEM}$; only resolvable cells can test the sign law, and Sec.~\ref{sec:law} explains why counting the rest would be misleading, not conservative. Cells in which any seed collapses to chance are flagged \emph{bistable}, excluded from headline
status, and reported as findings about trainability in their own right.

Four properties of this protocol matter, because each bounds what the audit of
Sec.~\ref{sec:lawaudit} can claim.

\emph{The two-SEM rule is a declared magnitude threshold, not an exact test.} With three seeds per arm, the Student-$t$ multiplier for a 95\% interval exceeds two, so the criterion is mildly liberal at the minimum seed count. We therefore fix it in advance and report the audit across a range of thresholds, not a single defended value (Table~\ref{tab:robust}). Because the seed-paired standard error is smaller than an independent propagation would give, this rule admits about $1.6$ times as many cells to the audit as the naive formula does, and the
extra cells are the borderline ones. That makes the test harder, not easier.

\emph{Cells are not independent.} A baseline arm is shared by every variant paired against it, and cells recur across the same datasets and backbones, so the \auditResolvable{} resolvable cells come from 235 distinct (dataset,
backbone, fraction) groups. We therefore report a cluster-level audit alongside the cell-level one, and treat exact $p$-values as indicative, not literal.

\emph{The aggregate claim is not a family of per-cell tests.} We never assert significance for an individual cell, so no multiplicity correction applies to the hit rate; the resolvability filter is a selection rule applied identically to every cell, declared before the audit.

\emph{Three-seed deltas are slightly optimistic.} Deepening the three headline CIFAR-100 cells to ten seeds per arm shrank every $\Delta$ ($0.07$, $0.15$, $0.39$ points), reproducing an earlier CIFAR-10 deepening; no envelope shape changed, but headline envelopes use ten seeds for that reason.

\subsection{Configuration selection, and its cost}\label{sec:selection} The reference configuration was not given; we chose it. Target family, tap depth, $\lambda_0$, head-norm, and loss form were each swept, and those sweeps
were run on CIFAR-100 and scored on the same test split that supplies this paper's CIFAR-100 numbers. We did not hold out a validation split, and we state the consequence rather than leave it to be inferred: \emph{the CIFAR-100 cells are a selection set, and their absolute values carry a selection bias of unknown size.}

Two things limit the damage, and one argument survives intact. The sweeps selected among a small number of pre-declared alternatives instead of searching a continuous space, and the chosen setting was fixed once and then applied verbatim everywhere else. That is what makes the transplant evidence the load-bearing evidence here: the same configuration, with no retuning, was applied to ten further datasets, nine backbones and two ImageNet-scale stages that played no part in its selection, and its behavior there is not selection-contaminated. Where this paper makes a general claim, it should be read as resting on those populations, with CIFAR-100 as the selection set
that produced the candidate.

We considered re-selecting on a held-out validation split and decided against it: re-selection would invalidate every cell in the grid, since the selected configuration is the one \computeCells{} cells were trained with, and the general claims already rest on twelve populations that took no part in selection. Not splitting a validation fold from the start is a design defect, not a defensible choice (Sec.~\ref{sec:limits} carries the remedy for a successor study).

\subsection{Comparator parity}\label{sec:parity}
Cost normalization fixes the comparators' budget, and that cuts both ways: every comparator runs at its published defaults inside the frozen recipe, at the declared multiple of the baseline's training cost, on the cell's own subset, while the prior was developed here and received the sweep just described. We did not sweep the comparators.

The largest consequence is at the smallest fractions. Self-supervised pre-training at $1\%$ of CIFAR-100 means 200 epochs of contrastive learning over 500 images, which is far short of the schedules used in published small-data practice, and our check that strengthening SimCLR's views does not help (Sec.~\ref{sec:comparators}) varies the view distribution, not the budget. So a statement such as ``SimSiam learns almost nothing at this scale''
is, strictly, a statement about SimSiam \emph{at this budget}; undertraining and method weakness are not separable under a cost-normalized protocol, by construction. Readers who care about the best attainable accuracy rather than
the best accuracy per unit compute should treat our self-supervised numbers as lower bounds.

We measured that instead of leaving it as a limitation: Sec.~\ref{sec:budget} re-runs both self-supervised comparators at $4\times$ that budget across the data envelope, including the two claims of ours that do
not survive it.

\subsection{Separating data from optimization}\label{sec:dataopt}
\looseness=-1 A data-efficiency benchmark fixes epochs, so every point on its data axis is also a point on a compute axis. Five of this study's results turned on that, each with the larger effect measured where a baseline had not finished training: the dense grid reversed its conclusion between $50$ and $200$ epochs (Sec.~\ref{sec:dense}); SimCLR at $5\%$ was step-starved, not data-starved (Sec.~\ref{sec:budget}); ViT-B's $+26.01$ became $+6.71$ once its baseline trained twice as long (Sec.~\ref{sec:scale}); the model-scale trend survived only because we then compared both models at that doubled budget; and the envelope's right flank, $+0.16$ under the frozen recipe, is $+5.92$ at matched steps and $+7.40$ on the features (Sec.~\ref{sec:ownenvelope}).

\looseness=-1 The epoch budget is one instance of something more general that this grid met three times, on three unrelated constants: freezing a parameter does not make it neutral; it makes the parameter's contribution invisible and silently reassigns it to whatever the study varies. The frozen epoch count folded optimization into the data axis, so the right flank read as data sufficiency (above); the fixed evaluation budget folded a label ratio into the \readout{} term, whose negative branch read as a training phenomenon until each cell was re-evaluated at its own budget (Sec.~\ref{sec:lawaudit}); the fixed auxiliary strength folded the schedule into the fusion, whose interference read as a property of combining sources until a weaker $\lambda_0$ removed it
(Sec.~\ref{sec:taxsub}). Each time the effect was real, its attribution wrong, and the remedy the same: unfreeze the one constant and re-measure. A frozen recipe guarantees comparability \emph{across} cells, not that an effect belongs to the intervention rather than to a constant, so we report the budget with every effect and measure at matched budget where a result depends on one. The claim is about this benchmark, not the field: it caught us out five times, and re-running exposed it where reasoning had not.

\subsection{Pre-registration}\label{sec:prereg} Each experimental wave was launched with numeric prediction bands and explicit falsifiers recorded in a project ledger before any result existed, and outcomes are reported against those bands including the misses. A registration a reader cannot inspect is an assertion, so the ledger is released with the artifacts as a version-controlled file: its history carries one timestamped commit per wave,
and the commits that record bands precede the commits that record the corresponding results. Several of our predictions failed instructively, ``SSL is data-regime-bounded'' (wrong: SSL won at every conv fraction under plain
augmentation) and ``augmentation substitutes for the prior'' (wrong in the opposite direction: it amplifies), and both failures became findings in Sec.~\ref{sec:fusion}. \looseness=-1 We regard this discipline as part of
the instrument, not a presentational choice: a benchmark whose predictions cannot fail measures nothing.

\section{The grid}\label{sec:grid}

\paragraph{What a cell is.} We report the grid in \emph{cells}: one configuration, identified by its configuration name and aggregated over the seeds trained under it---a dataset, a data fraction, a backbone, one intervention with fixed hyperparameters. Every quoted number is a seed mean with the standard error of that mean. Cells are grouped by configuration name, never by parsed fields (two configurations can agree on dataset, backbone and intervention while differing in a bank or calibration setting, and grouping on fields would silently average them), and a cell is not a run: the \computeCells{} cells rest on \textbf{\computeRuns{} individual training runs}, and every $\Delta$ pairs cells that share dataset, fraction and backbone and differ in exactly one intervention. The count includes every cell we trained, diagnostics and ablations included (Sec.~\ref{sec:recipe}).

\looseness=-1 Table~\ref{tab:glance} lists the backbones and interventions; Table~\ref{tab:partitions} gives the datasets they cross, with domain, resolution, and per-fraction image counts. The rest of this section describes the 13 datasets and six domains (Sec.~\ref{sec:datasets}), the hardware, wall-clock and carbon consumed (Sec.~\ref{sec:compute}), and the three views the results are built from: the prior's own envelope (Sec.~\ref{sec:ownenvelope}), the competing interventions on that envelope at matched cost (Sec.~\ref{sec:multimethod}), and the attention backbones, where the effect is largest (Sec.~\ref{sec:attention}). Two table conventions hold throughout: alternate rows are lightly shaded, and \up{} (\down{}) marks a column in which higher (lower) is better; signed method-versus-method differences carry no marker, since neither sign is better a priori. A dash marks a cell we did not measure.

\begin{table}[pos=htbp]
\caption{The benchmark's backbones and interventions. Every cell trains the frozen recipe of Sec.~\ref{sec:recipe}; costs are declared as multiples of baseline training compute; the datasets, domains, scales, and resolutions these cross are in Table~\ref{tab:partitions}. Backbone coverage is not uniform across comparators: DINO runs only on ViT-tiny, so its cells cross datasets but not architectures.}
\label{tab:glance}
\centering\scriptsize
\setlength{\tabcolsep}{2.6pt}
\zebra{2}
\begin{tabularx}{\columnwidth}{@{}lX@{}}
\toprule
\textbf{Axis} & \textbf{Coverage} \\
\midrule
Backbones & ResNet-18, -34 and -50 \citep{he2016resnet}, MobileNetV3
  \citep{howard2019mobilenetv3}, ConvNeXt-T \citep{liu2022convnext},
  ViT-tiny, -S and -B \citep{dosovitskiy2021vit}, Swin-T
  \citep{liu2021swin} \\
Interventions & MomentAux ($1.02\times$); SimCLR, SimSiam and DINO
  ($2\times$); ImageNet transfer; DeiT augmentation; FitNets teacher
  ($2\times$); HOG target; pairwise combinations \\
Measured & $\computeCells$ cells over $\computeRuns$ retained run records
  including exploration and repeats \\
\bottomrule
\end{tabularx}
\vspace{-2em}
\end{table}

\subsection{Datasets}\label{sec:datasets}
\begin{table*}[pos=tbp]
\caption{Datasets and the partitions they were trained on. Counts are read from the fixed subset index files every arm consumes, so they carry the rounding of the per-class stratified draw. Parentheses give images per class, which sets a cell's regime. Relabeled controls and multi-source views reuse their parent's indices exactly and share its row, which is why the twenty dataset identities counted in the sign-law audit are not twenty independent populations: five are relabeled controls of their parent's pixels and four are band subsets of two parent instruments, leaving eleven independent image sources. The two blanks differ: $\times$ marks a fraction that \emph{cannot} be trained under the frozen recipe, yielding fewer than one batch of 128 images; a dash marks one that is runnable but was not run.}
\label{tab:partitions}
\centering\scriptsize
\setlength{\tabcolsep}{1.9pt}
\zebra{4}
\begin{tabular}{llrrrrrrrrrrrrr}
\toprule
 & & & & \multicolumn{11}{c}{training images at fraction (per class)} \\
\cmidrule(l){5-15}
Domain & Dataset & Cls. & Px & 1\% & 2\% & 3\% & 5\% & 7\% & 10\% & 15\% & 20\% & 25\% & 50\% & 100\% \\
\midrule
Photographic & CIFAR-10 \citep{krizhevsky2009cifar} & 10 & 32 & \mcell{500}{(50)} & \mcell{1{,}000}{(100)} & \mcell{1{,}500}{(150)} & \mcell{2{,}500}{(250)} & \mcell{3{,}500}{(350)} & \mcell{5{,}000}{(500)} & \mcell{7{,}500}{(750)} & \mcell{10{,}000}{(1000)} & \mcell{12{,}500}{(1250)} & \mcell{25{,}000}{(2500)} & \mcell{50{,}000}{(5000)} \\
 & CIFAR-100 \citep{krizhevsky2009cifar} & 100 & 32 & \mcell{500}{(5)} & \mcell{1{,}000}{(10)} & \mcell{1{,}500}{(15)} & \mcell{2{,}500}{(25)} & \mcell{3{,}500}{(35)} & \mcell{5{,}000}{(50)} & \mcell{7{,}500}{(75)} & \mcell{10{,}000}{(100)} & \mcell{12{,}500}{(125)} & \mcell{25{,}000}{(250)} & \mcell{50{,}000}{(500)} \\
 & STL-10 \citep{coates2011stl10} & 10 & 96 & $\times$ & $\times$ & \mcell{150}{(15)} & \mcell{250}{(25)} & \mcell{350}{(35)} & \mcell{500}{(50)} & \mcell{750}{(75)} & \mcell{1{,}000}{(100)} & \mcell{1{,}250}{(125)} & \mcell{2{,}500}{(250)} & \mcell{5{,}000}{(500)} \\
 & Tiny-ImageNet \citep{le2015tinyimagenet} & 200 & 64 & \mcell{1{,}000}{(5)} & \mcell{2{,}000}{(10)} & \mcell{3{,}000}{(15)} & \mcell{5{,}000}{(25)} & \mcell{7{,}000}{(35)} & \mcell{10{,}000}{(50)} & \mcell{15{,}000}{(75)} & \mcell{20{,}000}{(100)} & \mcell{25{,}000}{(125)} & \mcell{50{,}000}{(250)} & \mcell{100{,}000}{(500)} \\
Satellite & EuroSAT \citep{helber2019eurosat} & 10 & 64 & \mcell{216}{(22)} & \mcell{432}{(43)} & \mcell{648}{(65)} & \mcell{1{,}080}{(108)} & \mcell{1{,}512}{(151)} & \mcell{2{,}160}{(216)} & \mcell{3{,}240}{(324)} & \mcell{4{,}320}{(432)} & \mcell{5{,}400}{(540)} & \mcell{10{,}800}{(1080)} & \mcell{21{,}600}{(2160)} \\
Multi-sensor & So2Sat LCZ42 \citep{zhu2020so2sat} & 17 & 32 & \mcell{243}{(14)} & \mcell{481}{(28)} & -- & \mcell{1{,}206}{(71)} & -- & \mcell{2{,}412}{(142)} & -- & -- & \mcell{6{,}028}{(355)} & -- & \mcell{24{,}119}{(1419)} \\
Texture & DTD \citep{cimpoi2014dtd} & 47 & 64 & $\times$ & $\times$ & $\times$ & \mcell{188}{(4)} & \mcell{282}{(6)} & \mcell{376}{(8)} & \mcell{564}{(12)} & \mcell{752}{(16)} & \mcell{940}{(20)} & \mcell{1{,}880}{(40)} & \mcell{3{,}760}{(80)} \\
Food & Food-101 \citep{bossard2014food101} & 101 & 64 & \mcell{808}{(8)} & \mcell{1{,}515}{(15)} & \mcell{2{,}222}{(22)} & \mcell{3{,}838}{(38)} & \mcell{5{,}252}{(52)} & \mcell{7{,}575}{(75)} & \mcell{11{,}312}{(112)} & \mcell{15{,}150}{(150)} & \mcell{18{,}988}{(188)} & \mcell{37{,}875}{(375)} & \mcell{75{,}750}{(750)} \\
Histopathology & PathMNIST \citep{yang2023medmnist} & 9 & 64 & \mcell{901}{(100)} & \mcell{1{,}800}{(200)} & \mcell{2{,}700}{(300)} & \mcell{4{,}498}{(500)} & \mcell{6{,}300}{(700)} & \mcell{9{,}000}{(1000)} & \mcell{13{,}499}{(1500)} & \mcell{17{,}998}{(2000)} & \mcell{22{,}500}{(2500)} & \mcell{44{,}996}{(5000)} & \mcell{89{,}996}{(10000)} \\
Fine-grained & CUB-200 \citep{wah2011cub} & 200 & 64 & $\times$ & $\times$ & \mcell{200}{(1)} & \mcell{394}{(2)} & \mcell{400}{(2)} & \mcell{600}{(3)} & \mcell{800}{(4)} & \mcell{1{,}200}{(6)} & \mcell{1{,}594}{(8)} & \mcell{2{,}994}{(15)} & \mcell{5{,}994}{(30)} \\
ImageNet scale & ImageNet64 \citep{chrabaszcz2017downsampled} & 1000 & 64 & \mcell{12{,}820}{(13)} & \mcell{25{,}620}{(26)} & \mcell{38{,}441}{(38)} & \mcell{64{,}062}{(64)} & \mcell{89{,}685}{(90)} & \mcell{128{,}118}{(128)} & \mcell{192{,}175}{(192)} & \mcell{256{,}231}{(256)} & \mcell{320{,}284}{(320)} & -- & \mcell{1{,}281{,}167}{(1281)} \\
 & ImageNet-100 \citep{deng2009imagenet} & 100 & 224 & \mcell{1{,}270}{(13)} & \mcell{2{,}532}{(25)} & -- & \mcell{6{,}335}{(63)} & -- & \mcell{12{,}669}{(127)} & -- & -- & \mcell{31{,}671}{(317)} & -- & \mcell{126{,}689}{(1267)} \\
\bottomrule
\end{tabular}
\end{table*}

\looseness=-1 Table~\ref{tab:partitions} lists the datasets. Six visual domains are represented because the currency account predicts that which information source is scarce depends on image statistics, and a benchmark drawn only from photographic data could not test that: satellite imagery has non-photographic color statistics and meaningful rotations, texture lacks object-level structure, food is texture-rich but object-centered, histopathology is stain-dominated, and fine-grained birds require discriminating parts, not categories. PathMNIST is a public, de-identified benchmark distributed as part of MedMNIST v2 \citep{yang2023medmnist}; using it here requires no new data collection and no ethics approval. The two ImageNet stages buy the axes small datasets cannot: ImageNet64 supplies data and label scale at fixed resolution, while ImageNet-100 supplies native $224$\,px resolution and a model-scale
curve.

Two training pools differ from the sizes usually quoted for those datasets, and Table~\ref{tab:partitions} reports the pool we actually drew from, not the headline split. For DTD we follow the standard protocol and train on train${+}$val ($3{,}760$ images, 80 per class), testing on the held-out test split. For So2Sat, we train on the v4 \emph{validation} split ($24{,}119$) and test on the v4 test split: both are held out from the training cities, which makes this a harder generalization setting than the canonical split and is why the $352{,}366$-image training split plays no part here.

A second group exists purely as controls. CIFAR-100-super relabels byte-identical CIFAR-100 subsets with 20 coarse classes, so class count changes while pixels and optimization steps do not; the Tiny-ImageNet variants do the same on a second population, including a semantically ordered grouping and an arbitrary positional one whose contrast isolates whether coherence or count is doing the work. CIFAR-100 \citep{barz2020cifair} replaces the 927 CIFAR-100 test images that duplicate training images, and re-evaluating the low-data cells on it moves the prior's gain by at most $0.20$ points at any fraction, so those gains are not memorized duplicates.

\subsection{Implementation, compute and carbon}\label{sec:compute}
The frozen recipe of Sec.~\ref{sec:recipe} is, concretely: stochastic gradient descent with momentum $0.9$, learning rate $0.1$, weight decay $5{\times}10^{-4}$, a cosine schedule, 200 epochs, batch size 128, and
random-crop + horizontal-flip augmentation only. Training uses PyTorch with automatic mixed precision and \texttt{timm} model definitions, pinned per machine (PyTorch 2.4--2.8, \texttt{timm} 1.0.27, CUDA 12.x), the PyTorch version recorded in every run record; metrics come from \texttt{torchmetrics} with a scikit-learn cross-check in the test suite, not
hand-rolled implementations. The study ran on the MareNostrum5 accelerated partition at the Barcelona Supercomputing Center (up to 40 NVIDIA H100 GPUs concurrently under a shared work queue), a university cluster with H100 and H200 nodes, and two local NVIDIA RTX 3090 workstations for evaluation passes and analysis. Because these small-image cells are dataloader-bound, not compute-bound, two to three trainings share a GPU, sized so the heaviest cells still fit in memory.

\looseness=-1 Aggregating the wall-clock in all \computeRuns{} retained run records gives
\textbf{\computeGpuHours{} run-hours}: \computeHhundred{} on H100, \computeNvl{} on H100~NVL and H200~NVL, and \computeAmpere{} on RTX 3090, each figure rounded independently of the total. Run-hours, not GPU-hours,
deliberately: two to three trainings share one device, so summing per-run wall-clock overstates device occupancy by roughly that factor. Taking thermal design power as an upper bound on draw, a power usage effectiveness of $1.2$ and a Spanish grid carbon intensity of $0.17$\,kg\,CO$_2$eq per kWh, this gives at most $\computeKwh$\,kWh and \textbf{\computeCarbon\,kg CO$_2$eq}, about $\computePerRun$\,g per run: an order-of-magnitude upper bound after the Machine Learning Impact calculator \citep{lacoste2019quantifying}, not a measurement (uniform throughput assumed; idle and host-side draw, embodied carbon and storage excluded; one national grid intensity). The figure covers every retained run, which is every run whose cell survives in the released grid, since the training entry point refuses to overwrite a completed cell. Two design choices lower the total: the intervention under test adds only about $2\%$ to a run, and the expense sits in the comparators, every self-supervised arm doubling the cost of its cell.

\subsection{The prior's own envelope}\label{sec:ownenvelope}

\looseness=-1 Tables~\ref{tab:envelope} and~\ref{tab:domainenv} give the prior's gain across the data axis, Table~\ref{tab:vitenvelope} gives it on ViT-tiny. Three regularities recur. The envelope is \emph{unimodal}: gains rise from the extreme-scarcity floor, peak in a dataset-specific band, and decay toward zero as data becomes sufficient. The peak's location tracks the dataset's difficulty, not its fraction, easy ten-way CIFAR-10 plateaus across 1--2\% (the two cells differ by $0.29$, which their seed uncertainty cannot resolve) and is already negative at 15\%, while hundred-way CIFAR-100 peaks at 5\%; Tiny-ImageNet's $3$ and $5\%$ cells differ by $0.22 \pm 0.26$ ($+2.34$ against $+2.12$), unresolvable, so its peak location between them is not adjudicated. And wherever a dataset is large enough for its $100\%$ cell to reach sufficiency, the prior is neutral there within noise: CIFAR-100 $-0.01$, CIFAR-10 $-0.26$, Tiny-ImageNet $-0.42$, Food-101 $-0.47$, EuroSAT ${\sim}0$, ImageNet64 $+0.04$ and ImageNet-100 $-0.02$, which is what the decaying weight schedule is for: the fusion asks nothing once the run is long enough to pay for everything. The qualifier is not cosmetic, and we enforce it elsewhere too: DTD and CUB-200 are still positive at $100\%$ ($+3.55$, $+2.74$, Table~\ref{tab:domainenv}), but their full training sets are only $3{,}760$ and $5{,}994$ images, which by this study's own axis are low-data cells, not sufficient ones. The image count determines the class, not the label ``$100\%$''.

\begin{table*}[width=0.485\textwidth,pos=htbp]
\begin{minipage}[t]{0.485\textwidth}
\caption{Complete envelopes: gain $\Delta$ of the reference-configuration prior over the shared baseline (points), with the seed-paired standard error of each difference beside it. Ten seeds per arm at CIFAR-100 1, 5 and 10\%, CIFAR-10 1 and 2\% and Tiny-ImageNet 1 and 5\%; three elsewhere. CIFAR-100 is the selection set (Sec.~\ref{sec:selection}); the CIFAR-10 2\% pair crosses a dataloader worker-count boundary and is the one headline pair whose arms are not seed-matched (Sec.~\ref{sec:recipe}). Best in bold, second best underlined, third best in italics, per dataset column.}
\label{tab:envelope}
\centering\scriptsize
\setlength{\tabcolsep}{2.6pt}
\zebra{2}
\begin{tabular}{lrrr}
\toprule
Fraction & CIFAR-100~\up & CIFAR-10~\up & Tiny-IN~\up \\
\midrule
1\%   & $+1.42$\sem{0.07} & \snd{$+6.37$}\sem{0.15} & $+1.49$\sem{0.09} \\
2\%   & $+2.50$\sem{0.16} & $\mathbf{+6.66}$\sem{0.27} & $+1.81$\sem{0.19} \\
3\%   & $+3.68$\sem{0.29} & \trd{+5.38}\sem{0.65} & $\mathbf{+2.34}$\sem{0.23} \\
5\%   & $\mathbf{+5.15}$\sem{0.22} & $+4.41$\sem{0.15} & \snd{$+2.12$}\sem{0.12} \\
7\%   & \snd{$+4.87$}\sem{0.51} & $+2.21$\sem{0.42} & \trd{+2.01}\sem{0.31} \\
10\%  & \trd{+3.75}\sem{0.24} & $+1.09$\sem{0.57} & $+1.65$\sem{0.27} \\
15\%  & $+2.55$\sem{0.30} & $-0.66$\sem{0.22} & $+1.44$\sem{0.11} \\
25\%  & $+0.16$\sem{0.29} & $-0.83$\sem{0.20} & $+0.10$\sem{0.28} \\
100\% & $-0.01$\sem{0.13} & $-0.26$\sem{0.18} & $-0.42$\sem{0.36} \\
\bottomrule
\end{tabular}
\end{minipage}\hfill
\begin{minipage}[t]{0.485\textwidth}
\caption{Domain datasets at representative fractions (points, three seeds). Best in bold, second best underlined, third best in italics, per row; rows with only two measured entries mark the best alone. The unimodal shape recurs on satellite, food and histopathology data. DTD rises across the three fractions shown (its full envelope dips at $10\%$ and peaks at $50\%$) and CUB-200 contributes a single point, so neither exhibits a shape here; the text explains why both are low-data cells. PathMNIST above ${\sim}15\%$ is confounded by a shifted test split that penalizes \emph{all} methods, so only its low-data cells are interpretable.}
\label{tab:domainenv}
\centering\scriptsize
\setlength{\tabcolsep}{2.6pt}
\zebra{2}
\begin{tabular}{lrrrr}
\toprule
Dataset & 1\%~\up & 5\%~\up & 10--15\%~\up & $\geq$25\%~\up \\
\midrule
EuroSAT   & $\mathbf{+2.47}$ & \trd{+0.84} & \snd{$+1.22$}  & ${\sim}0$ \\
PathMNIST & $\mathbf{+5.62}$ & --      & $+1.74$  & conf. \\
Food-101  & --      & $\mathbf{+5.63}$ & \snd{$+3.18$}  & \trd{$-$0.69} \\
STL-10    & --      & --      & $\mathbf{+5.92}$  & $+4.77$ \\
DTD       & --      & \trd{+0.30} & \snd{$+2.15$}  & $\mathbf{+3.55}$\rlap{$^{\ast}$} \\
CUB-200   & --      & --      & --       & $+2.74$\rlap{$^{\ast}$} \\
\bottomrule
\rowcolor{white}\multicolumn{5}{@{}l}{\tiny $^{\ast}$value at 100\%.} \\
\end{tabular}
\end{minipage}
\end{table*}

\paragraph{The envelope is joint in data and compute, and the two flanks are separate.} The recipe fixes epochs, not steps, so more data also buys proportionally more optimizer steps, and the axis above is both at once. We separated them instead of treating them as one. Holding the total at ${\sim}600$ steps on CIFAR-100 with ResNet-18 and varying only the data gives
$+1.39$, $+2.29$, $+5.78$, $+6.30$ and $+5.92$ at 1, 2, 5, 10 and $25\%$, against $+1.42$, $+2.50$, $+5.15$, $+3.75$ and $+0.16$ under the frozen recipe. The registered falsifier, a gain flat to within $1.5$ points across fractions, did not fire: the range is $4.91$, so the data axis is real, and the rising left flank survives matched steps. The right flank does not. At a fixed step budget, the envelope rises and then plateaus with no decay at all, the two agreeing at 1--5\% and separating sharply at 10--25\%. Neutrality at sufficiency is therefore sufficiency of \emph{optimization}: more data does not by itself make the prior redundant; the extra steps that more data buys under an epoch-based recipe do. The feature side says why, on bands registered before the evaluation ran and hit at all five fractions: at a matched budget the prior's frozen-feature gain at $25\%$ is still $+7.40$, where the frozen recipe leaves $+0.44$ on the same cell. The right flank is therefore feature-side, the extra $19{,}400$ steps letting the \emph{baseline's} features catch up, not leaving the prior's gain present but unrealized; the $1\%$ cell, where the two recipes coincide and only the worker count differs, bounds that nuisance at $-0.52 \pm 0.42$, an order of magnitude below the $+6.96 \pm 0.58$ divergence at $25\%$. Every statement of the envelope here, ``unimodal'' included, is therefore joint in data and compute (Sec.~\ref{sec:dataopt}).

\subsection{The multi-method comparison}\label{sec:multimethod}
The same envelope measured for every comparator produces the head-to-head tables that Secs.~\ref{sec:fusion} and \ref{sec:guide} analyze. As a preview: on CIFAR-100 under plain augmentation, SimCLR initialization at $2\times$ compute beats the $1.02\times$ prior by $+0.85$ to $+5.0$ points between 1\% and 10\% and converges back to parity by 50\% (the full margin envelope is Appendix~\ref{app:ssl}); SimSiam gains ${\leq}+0.9$ anywhere on this envelope at that budget, a property of the budget, not of the method (Sec.~\ref{sec:parity}). DINO, a ViT-only comparator in this grid, trails the prior on ViT-tiny at every CIFAR-100 fraction below $25\%$, though not universally: on PathMNIST it is nominally above the prior at $10\%$ ($+0.40$). At $5\times$ budget, SimCLR's lead widens to $+3.1$ to $+9.2$ from 1\% to 15\%, so on convolutional backbones the prior's case is one of cost, not accuracy; on attention backbones, and on any backbone under a modern augmentation recipe, the ordering inverts (Sec.~\ref{sec:fusion}).

\looseness=-1 Masked reconstruction, the one major self-supervised family the three above omit, was added at the same $2\times$ budget and protocol on ViT-tiny at CIFAR-100: $20.5$, $29.5$ and $42.7$ at 5, 10 and $25\%$, clearly ahead of DINO and level with SimCLR and the prior (trailing the prior by $1.1$, matching, then leading by $0.5$: the same crossing with data). Its feature gains ($+10.9$, $+13.3$, $+13.8$) sit below the prior's at the two smaller fractions, so masked prediction reads here as more of the same currency, not a new one; at $1$ and $2\%$ it gains $+0.02$ and $+1.5$, below SimCLR at both, starving on a few hundred images as the other families do.

\looseness=-1 Its combination with augmentation reproduces the inversion reported above: under DeiT-strength
augmentation it reaches $21.3$, $34.5$ and $50.5$, at or below SimCLR's $21.6$, $36.1$ and $52.3$ and $6.8$ to $7.6$ below the prior's, with its $10\%$ feature gain ($+15.0$) likewise below SimCLR's $+17.3$ and the prior's $+22.2$. The three sources enter the augmented recipe from nearly indistinguishable plain-recipe feature gains ($13.3$ to $14.9$) and leave it $+1.7$, $+3.8$, and $+7.4$ higher, an ordering that reproduces the accuracy ordering: augmentation compounds least with the source whose currency most resembles the invariance it already supplies. The $5\%$ column is the noisiest (two low seeds, probe standard
error $3.5$), so the $10$ and $25\%$ cells carry the claim.

\subsection{Attention: where the prior is dramatic}\label{sec:attention}
\begin{table}[pos=htbp]
\caption{ViT-tiny: gain of the prior under the plain recipe and under DeiT-strength augmentation, with the amplification ratio, on CIFAR-100 (selection set, Sec.~\ref{sec:selection}) and, in the last three columns, Tiny-ImageNet. Augmentation does not substitute for the prior; it amplifies it and shifts its peak toward more data. Blank: not measured. Three seeds per cell, six on the DeiT $5\%$ arm. Best in bold, second best underlined, third best in italics, per $\Delta$ column; the ratio columns are derived and unstyled. Ratios, and the ranges quoted for them in the text, are computed from unrounded deltas, and should be read to about $\pm 0.3$: their denominators are augmented baselines whose seed standard deviations reach $4.9$ points on Tiny-ImageNet.}
\label{tab:vitenvelope}
\centering\scriptsize
\setlength{\tabcolsep}{1.9pt}
\zebra{5}
\begin{tabular}{lrrrrrr}
\toprule
 & \multicolumn{3}{c}{CIFAR-100} & \multicolumn{3}{c}{Tiny-IN} \\
\cmidrule(lr){2-4}\cmidrule(l){5-7}
Fraction & $\Delta_{\mathrm{plain}}$~\up & $\Delta_{\mathrm{DeiT}}$~\up & Ratio~\up & $\Delta_{\mathrm{plain}}$~\up & $\Delta_{\mathrm{DeiT}}$~\up & Ratio~\up \\
\midrule
1\%   & $+1.4$  & $+3.2$            & 2.4 & $+1.3$ & $+3.0$ & 2.4 \\
2\%   & $+3.3$  & $+6.0$            & 1.8 & $+3.3$ & $+6.6$ & 2.0 \\
3\%   & $+6.2$  & $+9.5$            & 1.5 & -- & -- & -- \\
5\%   & $+9.4$  & $+16.3$           & 1.7 & $+6.8$ & $+10.2$ & 1.5 \\
7\%   & $+11.0$ & $+18.7$           & 1.7 & -- & -- & -- \\
10\%  & \trd{+13.3} & \trd{+21.0}   & 1.6 & \trd{+9.2} & $\mathbf{+16.9}$ & 1.9 \\
15\%  & $\mathbf{+14.4}$ & \snd{$+24.6$}  & 1.7 & $\mathbf{+10.7}$ & \snd{$+16.7$} & 1.6 \\
25\%  & \snd{$+13.7$} & $\mathbf{+25.1}$  & 1.8 & \snd{$+10.5$} & \trd{+15.1} & 1.4 \\
100\% & $+9.9$  & $+13.9$           & 1.4 & $+7.2$ & $+6.6$ & 0.9 \\
\bottomrule
\end{tabular}
\end{table}

Table~\ref{tab:vitenvelope} previews the study's most striking axis. ViT-tiny trained from scratch gains $+9.9$ points from the prior even at \emph{full} CIFAR-100, a scale at which the ResNet backbones are neutral, and under the standard DeiT recipe the gains grow to $+25$ points, peaking later in the data axis. The best ViT number in the study, $\vitBest\%$ at full CIFAR-100, is DeiT augmentation \emph{plus} the prior, compared with $\vitBestBase\%$ for the recipe alone: the prior adds $+\vitBestDelta$ points on top of the recipe the literature recommends. Sections~\ref{sec:law} and \ref{sec:scale} show this is a feature-side phenomenon that persists to ImageNet scale and grows with model size.

\section{The organizing law}\label{sec:law}

\looseness=-1 A grid of $\computeCells$ cells is only as useful as the account that organizes it. The practical headline of this section is its budget-matched result: evaluated at a cell's own label budget, the frozen-feature gain predicts the end-to-end gain to $0.17$ points on average over 30 cells and seven datasets (end of Sec.~\ref{sec:lawaudit}), so the gain is carried by the representation rather than by the readout. Linear evaluation does not replace evaluating an already-trained classifier; it is a representation-side diagnostic whose magnitude tracks the end-to-end gain under matched label budgets. The full-label sign law explains that result, and the rest of the section builds on it. We state its form, in which the end-to-end gain $\Delta$ decomposes into a feature term $G$ and a \readout{} term governed by baseline accuracy, and what the law is not (Sec.~\ref{sec:lawform}); audit that form over the whole grid under a resolvability criterion that keeps uninformative cells from inflating the score (Sec.~\ref{sec:lawaudit}); give the three episodes that distinguish a law from a curve fit, calling unmeasured quantities in advance including for a backbone family it had never seen (Sec.~\ref{sec:lawpredict}); and localize what each term depends on, which makes the fusion outcomes of Sec.~\ref{sec:fusion} separable after the fact, not merely observed (Sec.~\ref{sec:lawG}).

\subsection{Form and origin}\label{sec:lawform}
For each paired cell we measure the end-to-end gain $\Delta$ and the feature gain $G$ (Sec.~\ref{sec:instrument}), and define $\readout \equiv \Delta - G$.

\textbf{What is definitional and what is not.} Eq.~\ref{eq:law} is true by construction, and we state that plainly: $\Delta = G + (\Delta - G)$ decomposes a measured quantity and cannot fail. It earns its place by naming two terms that behave and are measured differently, not by being a discovery. \emph{Every} empirical claim we make lives in the second term: that its \textbf{sign} is predictable from the accuracy of the already-trained shared baseline, before the intervention arm is evaluated with the linear probe. Where we write ``the law'' below, we mean that sign regularity, never the identity. What that sign \emph{means} depends on the evaluation's label budget, which we measure instead of assuming at the end of Sec.~\ref{sec:lawaudit}.

\begin{figure}[pos=htbp]
\centering
\includegraphics[width=\linewidth]{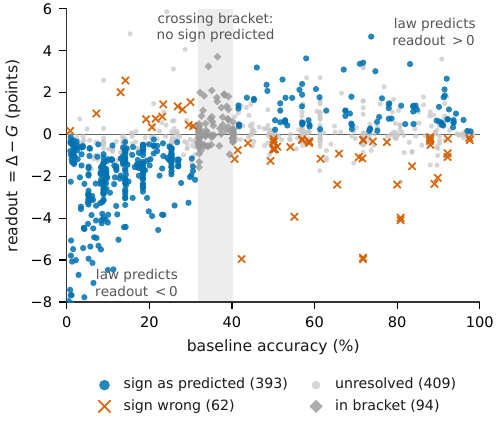}
\caption{The law, in full: \readout{} $=\Delta-G$ against baseline accuracy for all \auditScope{} law-scope cells with ${\geq}3$ seeds on both arms. Shaded band: the measured crossing bracket $[31.8,40.3]$, inside which no sign is predicted. Only the first two legend entries test the account: of the \auditResolvable{} cells whose \readout{} is resolvable against its own seed-paired uncertainty, \auditCorrect{} fall on the predicted side of zero. The other two entries keep the audit's reach visible: \auditUnresolved{} are unresolvable and \auditBracket{} sit inside the bracket, so the rate is $\auditRate\%$ at $\auditCoverage\%$ coverage against a $\auditMajority\%$ majority-sign baseline. The scatter at fixed baseline is substantial ($R^2 = 0.27$, residual SD
$\auditResidSD$ points): the account predicts the sign of this term, not its value.}
\label{fig:law}
\end{figure}

\looseness=-1 Across the grid, baseline accuracy is the largest single determinant of this term and the only one whose effect is consistently signed across datasets and backbones. We are deliberate about how strong that claim is: binned on baseline accuracy alone, the \readout{} term has $R^2 = 0.27$, dataset identity explains a further $0.13$ of the residual, data fraction $0.02$ and backbone essentially none ($0.003$), and the residual standard deviation at fixed baseline is $\auditResidSD$ points. This is a first-order trend with substantial scatter, not a curve that pins a cell's \readout{}. All five moved with the scope repairs of Sec.~\ref{sec:lawaudit}, chiefly the first, since the cells that the repair removed carry large negative $\Delta$ and $G$ and were the widest-scattered points in the fit. What is robust is the \emph{sign}, which is what this section audits. The term is strongly negative at low baselines, where features improve more than accuracy can show because a classifier with five labels per class cannot realize a better representation; it crosses zero in a narrow band we bracket at $[31.8, 40.3]$ points of baseline accuracy; and it is
positive above the crossing, largest just above it ($+1.6$ at baseline $39$) and decaying with sufficiency ($+0.4$ by baselines of $70$--$80$). Figure~\ref{fig:law} shows the full picture; we deliberately fit no parametric form and use only the measured curve for prediction.

\looseness=-1 That regularity is not a theorem but an empirical one with measured scope, registered in advance (Sec.~\ref{sec:lawpredict}) and found only after two wrong intermediate laws, a ``deficit'' law and a multiplicative realization$\times$gain form, were falsified by our own cells and retracted. Its content is asymmetric: below the crossing it predicts a large negative \readout{} and is strongly falsifiable; far above it predicts \readout{} ${\approx}0$, true but weakly discriminating, so a naive sign-count over high-data cells would produce a coin flip and wrongly suggest failure. The audit therefore counts only cells whose \readout{} is resolvable against its own \emph{seed-paired} uncertainty (Sec.~\ref{sec:stats}), tighter than independent propagation and the protocol the released script implements.

\subsection{The audit}\label{sec:lawaudit}

\begin{table}[pos=htbp]
\caption{Robustness of the audit: the rate is stable against the resolvability threshold, survives clustering on the units that share a baseline arm, and survives re-fitting the crossing bracket without the audited dataset (the out-of-sample check). It is carried by the left flank, where the account predicts strongly, not by the right, where it predicts ${\approx}0$. Intervals are Wilson 95\%.}
\label{tab:robust}
\centering\scriptsize
\setlength{\tabcolsep}{2.6pt}
\zebra{2}
\begin{tabular}{P{0.30\columnwidth}rrrP{0.20\columnwidth}}
\toprule
Audit variant & Cells & Correct & Rate & 95\% CI \\
\midrule
$|\readout| > 1$\,SEM & 629 & 509 & 80.9\% & [77.7, 83.8] \\
$>1.5$\,SEM & 531 & 446 & 84.0\% & [80.6, 86.9] \\
\textbf{$>2$\,SEM (declared)} & \textbf{\auditResolvable} & \textbf{\auditCorrect} & \textbf{\auditRate\%} & \textbf{[82.9, 89.2]} \\
$>3$\,SEM & 342 & 304 & 88.9\% & [85.1, 91.8] \\
\midrule
One vote per (dataset, backbone, fraction) & 235 & 192 & 81.7\% & [76.3, 86.1] \\
Held out: bracket fitted without the audited dataset & 391 & 347 & 88.7\% & [85.2, 91.5] \\
\midrule
Below the crossing & 314 & 298 & 94.9\% & [91.9, 96.8] \\
Above the crossing & 141 & 95 & 67.4\% & [59.3, 74.6] \\
\bottomrule
\end{tabular}
\end{table}

\looseness=-1 Table~\ref{tab:audit} summarizes the audit: of \auditResolvable{} resolvable cells spanning seven backbones and nineteen of the twenty dataset identities in scope (the caption of Table~\ref{tab:partitions} explains why identities outnumber image sources; one has law-scope cells but no resolvable one), \auditCorrect{} ($\auditRate\%$) fall on the predicted side (Wilson 95\% CI $[82.9, 89.2]$). Eleven cells entered scope late, ten of them resolvable, when the prior-with-augmentation arms of Sec.~\ref{sec:stack} and the fixed-step cells acquired their frozen-feature evaluations. All are aux-from-scratch cells with both measurements, so the law's own definition admits them; excluding them after seeing their effect would be the selection this benchmark exists to expose, and admitting them lowered the rate by three-tenths of a point on overlapping intervals.

\begin{table}[pos=htbp]
\caption{Scope-wide sign-law audit, regenerated from the per-run records by the released audit script. A cell is resolvable when $|\readout| > 2\,\mathrm{SEM}$; cells inside the crossing bracket make no sign prediction. Uncertainty is seed-paired (Sec.~\ref{sec:stats}), not propagated from independent arms. Scope is aux-from-scratch: cells carrying an ImageNet or self-supervised initialization are excluded here and scored separately in Sec.~\ref{sec:fusion}, and $100\%$ of cells report only the aux-vs-baseline gap under the scope rule of Sec.~\ref{sec:probe}, which is why the first row is much larger than the second; the rows narrow to the resolvable subset that actually tests the account.}
\label{tab:audit}
\centering\scriptsize
\setlength{\tabcolsep}{2.6pt}
\zebra{2}
\begin{tabular}{lr}
\toprule
Population & Cells \\
\midrule
Cells with paired $\Delta$ and $G$ (all interventions) & \auditAllPaired \\
\quad in scope, ${\geq}3$ seed-matched arms & \auditScope \\
Backbone variants in scope (architecture families) & 7 (5) \\
Dataset identities in scope & 20 \\
Inside crossing bracket (no prediction) & \auditBracket \\
Unresolved ($|\readout| \le 2$\,SEM) & \auditUnresolved \\
\midrule
\textbf{Resolvable (these test the account)} & \textbf{\auditResolvable} \\
\quad Sign as predicted & \textbf{\auditCorrect{} (\auditRate\%)} \\
\quad Wrong side & \auditWrong \\
\bottomrule
\end{tabular}
\end{table}

\looseness=-1 Two numbers belong beside that one. The resolvability rule decides which cells vote, and only $\auditCoverage\%$ of scope cells do, so we write the result as $\auditRate\%$ \emph{at $\auditCoverage\%$ coverage} throughout. The null matters as much: an exact binomial against a coin returns $p \approx 2\times10^{-58}$, true and useless, because nobody proposes a coin. The \readout{} term is negative in most resolvable cells, so always predicting negative already scores $\auditMajority\%$; the account beats that, but the honest effect size is the gap to $\auditMajority\%$, not to $50\%$. Below the crossing, where the account makes a strong prediction, it is $298/314 = 94.9\%$; above it, where it predicts only that the term is small, $95/141 = 67.4\%$. The \auditWrong{} exceptions group rather than scatter. One carries a single signature, $G$ clearly positive while $\Delta$ is negative at ${\geq}20\%$ data on Food-101 and PathMNIST, ``better features, worse accuracy at sufficiency''. A second is an artifact of the account's \emph{input}, not its prediction: the two largest exceptions are Swin cells whose baseline arm is seed-bistable, so their nominal baseline height averages trained and collapsed seeds and does not measure how well the task is being solved. We leave them in the count and name them instead: dropping bistable-baseline cells would raise the reported rate, so excluding them is not neutral. Table~\ref{tab:exceptions} lists the largest-magnitude exceptions individually, because where an account fails is part of what it claims.

\paragraph{Two scope repairs that moved this number, and by how much.} \looseness=-1 An earlier version of this audit reported $\auditPrevRate\%$ over $\auditPrevResolvable$ resolvable cells from a scope of $\auditPrevScope$; the figure is higher because two scope filters were repaired, not because any run, seed, or measurement changed. First, cells carrying an ImageNet initialization, the interference case of Sec.~\ref{sec:taxsub} and outside the regime the \readout{} branch was estimated in, leaked in through a string mismatch: the filter tested the exported \texttt{pretrained} field against \texttt{true} and \texttt{1} while the exporter writes \texttt{yes}. Restoring it removes \auditExclCells{} cells, \auditExclResolvable{} of them resolvable (\auditExclCorrect{} correct, \auditExclWrong{} wrong, $\auditExclRate\%$ on their own), taking the scope to $\auditMidScope$ and the rate to $\auditMidRate\%$ over \auditMidResolvable{} resolvable cells; below the crossing the $298$ correct cells are untouched while $23$ wrong ones leave ($\auditPrevBelowRate\%$ to $\auditMidBelowRate\%$), Sec.~\ref{sec:taxsub} seen from the other side. Second, at $100\%$ cells, Sec.~\ref{sec:probe} reports the aux-vs-baseline gap and refuses the $G$/\readout{} split, yet the released script carried no such filter. Removing the \auditHundredCells{} full-data cells drops \auditHundredResolvable{} resolvable cells (\auditHundredCorrect{} correct, \auditHundredWrong{} wrong, $\auditHundredRate\%$ on their own, the worst-scoring subset in the audit and exactly the cells the rule calls uninterpretable), moving the headline from $\auditMidRate\%$ to $\auditRate\%$ and the below-crossing flank to $\auditBelowRate\%$ (\auditHundredBelow{} more cells, both wrong). The price is breadth: ViT-S, ViT-B and ImageNet-100 hold law-scope cells only at $100\%$, so the span narrows from nine backbones to seven and from twenty-one dataset identities to twenty, though the five architecture families survive. Both pre-repair headlines and both excluded subsets' own rates are stated above, so the repair stays visible. Sec.~\ref{sec:scale} is untouched: Table~\ref{tab:scale} reports $\Delta$, $G$ and their gap with no signed \readout{}, which the rule permits, so the ImageNet $\Delta \approx G$ result stands while the same cells leave the sign-law count.

\begin{table}[pos=htbp]
\caption{The ten largest-magnitude resolvable exceptions, of \auditWrong{} in total, printed from the same audit as Table~\ref{tab:audit}; they cluster, not scatter. \emph{Top:} Swin cells whose baseline arm is unstable (seed SD $10$--$22$ points), so the baseline height the account takes as input is not a task-performance reading; the records flag the first two as bistable and the third narrowly not. \emph{Second:} the ``better features, worse accuracy at sufficiency'' signature. \emph{Third:} PathMNIST, whose linear evaluation is a compressed measuring stick. \emph{Bottom:} a heavily-augmented ViT cell on DTD, with \readout{} positive far below the crossing.}
\label{tab:exceptions}
\centering\scriptsize
\setlength{\tabcolsep}{2.3pt}
\zebra{2}
\begin{tabular}{lllrrrrr}
\toprule
Dataset & Backbone & Variant & Frac. & Base & $\Delta$ & $G$ & \textit{Readout} \\
\midrule
EuroSAT & Swin & aux & 20\% & 22.7 & $+71.27$ & $+55.93$ & $+15.34$ \\
EuroSAT & Swin & aux & 15\% & 18.6 & $+73.72$ & $+61.66$ & $+12.06$ \\
EuroSAT & Swin & aux & 2\%  & 42.3 & $+30.54$ & $+36.49$ & $-5.95$ \\
\midrule
Food-101 & R18 & mag3 & 50\% & 71.8 & $-1.30$ & $+4.66$ & $-5.96$ \\
Food-101 & R18 & mag6o & 50\% & 71.8 & $-0.91$ & $+4.97$ & $-5.88$ \\
Food-101 & MNet & aux & 50\% & 55.1 & $-0.23$ & $+3.70$ & $-3.93$ \\
\midrule
PathMNIST & R18 & mag6o & 1\% & 80.8 & $+5.80$ & $+9.90$ & $-4.09$ \\
PathMNIST & R18 & mag3 & 1\% & 80.8 & $+5.62$ & $+9.60$ & $-3.98$ \\
PathMNIST & MNet & aux & 3\% & 80.0 & $+2.26$ & $+4.66$ & $-2.40$ \\
\midrule
DTD & ViT & deit-aux & 50\% & 14.2 & $+17.30$ & $+14.73$ & $+2.57$ \\
\bottomrule
\end{tabular}
\end{table}

Table~\ref{tab:robust} reports what that number survives. Not the resolvability threshold: the rate moves only from $80.9\%$ over 629 cells at one SEM to $88.9\%$ over 342 at three, so a stricter or looser criterion reaches the same conclusion. Not the dependence between cells: collapsing to one vote per (dataset, backbone, fraction) leaves $192/235 = 81.7\%$. And, the check that matters most, not the estimation of the crossing bracket on the cells we then audit: re-fitting it on the other datasets and auditing each dataset out of sample gives $347/391 = 88.7\%$, indistinguishable from the in-sample figure.

That row audits 391 cells, not all \auditResolvable{}, and the shortfall is not a sampling choice. Re-fitting the bracket on the other datasets widens it beyond the pooled $[31.8, 40.3]$, since one dataset's cells no longer pin either edge (for CIFAR-100 it becomes $[29.5, 71.0]$), and cells inside a bracket make no sign prediction, so $46$ leave the held-out set that way, 25 of them CIFAR-100's; a further $18$ sit in datasets that cannot receive a fold ($391 + 46 + 18 = \auditResolvable$). The bias runs one way: the held-out row keeps the cells furthest from the crossing, where the account is most confident, so its rate is a mild upper bound rather than a like-for-like comparison, and the effect is small since $86\%$ of resolvable cells do receive an out-of-sample verdict.

What the number does \emph{not} survive is disaggregation. The evidence is overwhelmingly left-flank: below the crossing, the sign is right $94.9\%$ of the time; above it, $67.4\%$; close to a coin once the term it predicts has shrunk to nothing. One dataset behaves badly out of sample, PathMNIST at $52.6\%$: its linear evaluation is a known compressed measuring stick, its frozen-feature accuracy sitting \emph{below} its own end-to-end accuracy, so we treat its $G$ as unreliable there, not the account as refuted, and report the failure instead of excluding the dataset. We do not explain this cluster; we name it as the law's known boundary (Sec.~\ref{sec:limits}), and the released audit script enumerates every exception.

\paragraph{What the \readout{} term is, once the label budget is matched.} Every $G$ above uses one evaluation budget for all cells, the full training split, while the cell itself is trained on a fraction of it: at $1\%$, the evaluation holds a hundred times the cell's labels. That asymmetry could on its own explain the negative branch, so we measured it, re-evaluating both arms of $30$ reference-configuration cells across seven datasets, baselines $5.3$ to $93.8$, at \emph{each cell's own} per-class budget (five to $2{,}500$ labels per class).

The stronger result comes first. At matched budget, the end-to-end gain and the frozen-feature gain agree to $0.17$ points on average, median $0.11$ and worst case $0.77$, against $0.89$ under the full-label protocol. At the label budget a practitioner actually has, the frozen-feature gain therefore tracks the end-to-end gain closely, which places the intervention's effect in the representation rather than in the readout, and what had been a two-cell observation now holds over $30$ cells and seven datasets.

The second result follows from it and cuts against our own framing, which is why we registered it in advance. The \readout{} term largely disappears at matched budget: below the crossing its mean moves from $-1.66$ to $-0.19$ and above it from $+0.10$ to $-0.13$; $19$ of $30$ cells lose more than half their magnitude, and only $3$ of $30$ stay resolvable against their own uncertainty. Below the crossing, the negative sign is thus substantially a statement about the ratio of evaluation labels to training labels, for which baseline accuracy is a monotone proxy, rather than an independent property of training. The sign law remains an accurate description of the full-label protocol, the protocol under which every $G$ in this paper is measured, so Tables~\ref{tab:audit} and \ref{tab:robust} stand exactly as measured; what changes is how to read them.

\subsection{Prediction, not description}\label{sec:lawpredict}
Three pre-registered episodes distinguish the law from a post-hoc fit.

\emph{(i) A new backbone family from baselines alone.} Before any Swin-T linear evaluation existed, we read the \readout{} term off the measured curve at Swin's baseline heights and published bands for its feature gain ($G \approx 9.6$, $8.8$, $9.8$ and $7.0$ at 5, 10, 15 and 25\% of CIFAR-100; band $+7$--$+11$). The measurements landed at $10.5$, $9.3$, $10.1$, $7.6$: four of four in band, each within one point of its point call, against two live falsifiers (attention-intrinsic $G{\geq}13$; stabilization-only $G{\leq}6$), either of which would have refuted the account.

\emph{(ii) Four unseen domains.} The same procedure applied to satellite, texture, food, and histopathology cells straddling the crossing produced eight banded predictions; seven landed in band, and all eight had the predicted sign, including the two deliberately hard cases where a small $\Delta$ could have hidden a large $G$ (it did not: small $\Delta$ meant small $G$).

\emph{(iii) ImageNet scale}, the subject of Sec.~\ref{sec:scale}, where the residual $\Delta - G$ was pre-registered to stay within $1.5$ points on every backbone.

\subsection{What $G$ is a function of}\label{sec:lawG}
Controlled crossings localize the law's terms. $G$ is a property of the checkpoint and the evaluation's label space. It is invariant to which coarse partition relabels the data, but is cut roughly from $4.2$ to $2.9$ when training supervision moves from 200-way fine to 20-way coarse labels on identical pixels: fine-grained weak supervision is where the prior has the most feature work to do. The \readout{} term follows task performance, not label budget: engineered cells that lower class count without raising baseline accuracy get no \readout{} boost (a prediction our own account initially got wrong, and the cell that corrected it is in Appendix~\ref{app:controls}). And $G$ depends on the backbone as sharply as on the data: at the same Tiny-ImageNet cell, ResNet-18 carries $G = +2.66$ while MobileNetV3 carries $G = +0.01 \pm 0.23$, a measured zero on the weaker network, so ``how much a prior can add'' is not a dataset property but a (dataset, architecture) property.

\subsection{What the prior leaves behind}\label{sec:imprint}
\begin{figure}[pos=htbp]
\centering
\includegraphics[width=0.94\linewidth]{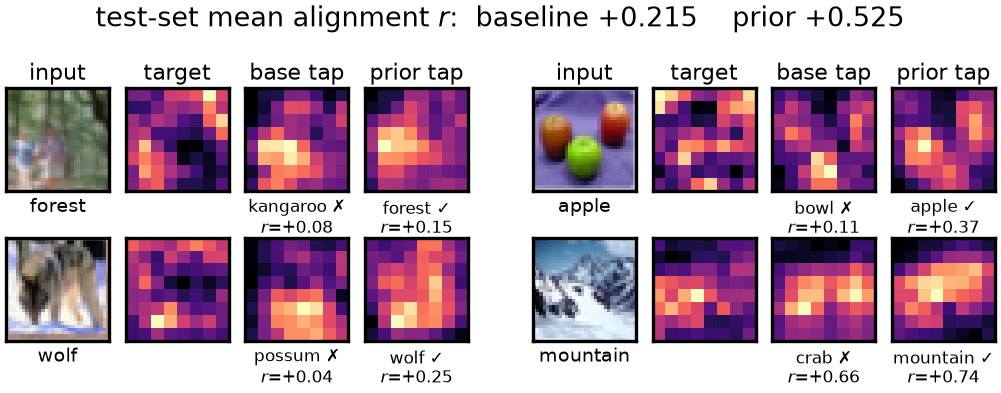}
\caption{The spectral imprint at the tap, CIFAR-100 at 5\%. Each of the four sample blocks shows the input, the fixed moment target, and the tapped-stage energy of the baseline and prior arms. Over the \emph{whole} test set, the alignment between tapped features and the target is $+0.215$ for the baseline and $+0.525$ for the prior arm. The four samples are \emph{selected} cases (prior arm correct, baseline wrong), so they illustrate the alignment difference, not sample it, and only the test-set-wide figure should be read as evidence.}
\label{fig:heatmap}
\end{figure}

$G$ is a number, and it is worth seeing what it corresponds to in the representation. The prior's target is a fixed bank (Fig.~\ref{fig:bank}), and its imprint survives to the end of training: at the CIFAR-100 5\% cell, where the feature gap is largest on this dataset ($+6.3$ points), the alignment between tapped features and the bank's oriented-energy target measured over the whole test set is $+0.525$ for the prior arm against $+0.215$ for the baseline (Fig.~\ref{fig:heatmap}). Survival is what matters, because the auxiliary weight decays to exactly zero well before training ends: the prior changes where the representation settles, not what the loss pulls on at the finish.

\begin{figure}[pos=htbp]
\centering
\includegraphics[width=0.94\linewidth]{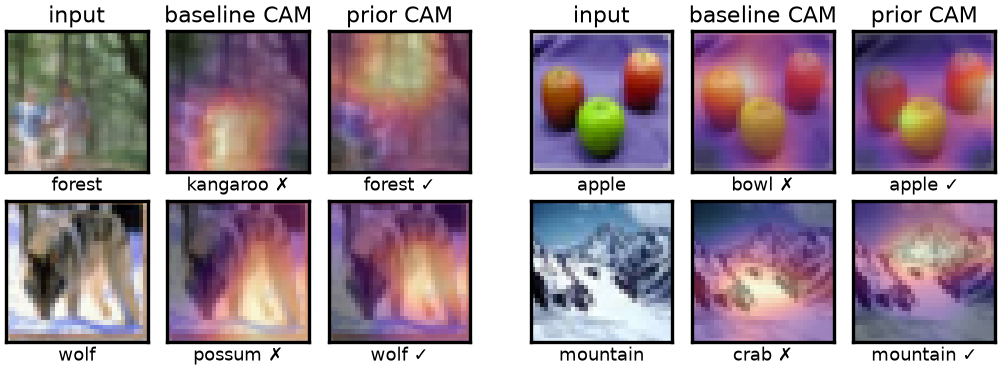}
\caption{Where the class evidence sits, CIFAR-100 at $5\%$. The four samples are \emph{selected} cases (prior arm correct, baseline wrong), so they show what the difference looks like where it is present, not how often. The evidential quantity is measured over the whole test set instead: the Gini coefficient of each class-activation map normalized to sum to one, where higher means evidence concentrated at fewer locations, gives $\camBase$ for the baseline against $\camAux$ for the prior arm over $\camN$ images ($\camSigma\,\sigma$). The prior's evidence is measurably more concentrated, by a modest amount, not the categorical difference the selected samples suggest.}
\label{fig:cam}
\end{figure}

A second, independent read asks not what the features encode but where the class evidence sits: Fig.~\ref{fig:cam} scores the concentration of each class-activation map over the whole test set, and the prior's evidence is measurably more concentrated, though only by $+\camDelta$.

\begin{figure}[pos=htbp]
\centering
\includegraphics[width=\linewidth]{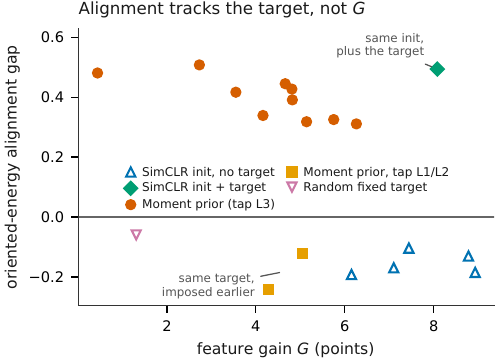}
\caption{Alignment gap against feature gain, one point per cell, CIFAR-100 / ResNet-18. Filled markers had the moment target during training. The two quantities are independent: every cell whose target was imposed at the tapped stage sits at $+0.31$ to $+0.51$ regardless of whether its $G$ is $2.7$ or $8.1$, and every cell without a target there sits below zero across the same range of $G$. The two labeled controls separate the accounts: the same target imposed two stages earlier leaves no imprint at layer~3, and the same self-supervised initialization as the negative points, with the target added, gives the largest gap measured.}
\label{fig:imprint2}
\end{figure}

\paragraph{Is that imprint specific, or is it what good features look like?} One cell cannot tell the two apart, and any intervention that improves features might show the same alignment, so we measured the same statistic across families that reach a positive $G$ by \emph{different} routes, against one pinned target for every model, since a self-supervised initialization has no auxiliary head and therefore no target of its own. Reporting the alignment gap, each arm minus its own baseline (Table~\ref{tab:imprint}, Fig.~\ref{fig:imprint2}):

\begin{table}[pos=htbp]
\caption{The imprint follows the \emph{target}, not the feature gain. The alignment gap is the arm minus its own baseline, CIFAR-100/ResNet-18, with one pinned target applied to every model. The lower block is a $2\times2$ that fell out of the design: crossing \{self-supervised initialization or not\} with \{moment target or not\}, the imprint tracks the target while $G$ tracks the initialization. Best in bold, second best underlined, third best in italics, per numeric column; the two columns' markings disagree.}
\label{tab:imprint}
\centering\scriptsize
\setlength{\tabcolsep}{4pt}
\zebra{2}
\begin{tabular}{lrrr}
\toprule
Family & $n$ & mean $G$ & alignment gap \\
\midrule
Moment prior (tap L3)        & \imprintPriorN & $+\imprintPriorG$ & \snd{$+\imprintPrior$} \\
SimCLR init, no target       & \imprintSSLN   & \snd{$+\imprintSSLG$}   & $\imprintSSL$ \\
Random fixed target          & 1              & $+1.31$           & \trd{\imprintRand} \\
\midrule
Moment prior, tapped at L1/L2 & 2 & \trd{+4.67} & $\imprintTap$ \\
SimCLR init $+$ moment target & 1 & $\mathbf{+8.08}$ & $\mathbf{+\imprintCombo}$ \\
\bottomrule
\end{tabular}
\end{table}

The self-supervised cells are the decisive comparison, and they answer it in the strongest available direction: they carry \emph{more} feature gain than the prior ($G = +\imprintSSLG$ against $+\imprintPriorG$) and \emph{less} oriented-energy structure than their own baselines ($\imprintSSL$). We had registered a band of $-0.05$ to $+0.10$ for them and a falsifier at $\geq +0.15$; the measurement came back at the opposite sign, so oriented-energy alignment is not simply what better features look like on this data.

Two controls fell out of the design unplanned, and each isolates one factor more sharply than the comparison we set out to make. Moment-prior cells \emph{tapped two stages earlier} use the identical target and show $\imprintTap$ when read at layer~3: the imprint is localized to where the target is imposed, which a mechanism predicts, and an incidental correlation does not. And a cell with the same SimCLR initialization as the $\imprintSSL$ rows, plus the moment target, shows $+\imprintCombo$, the largest gap we measure. Crossed, these give a $2\times2$ in which the imprint follows the target and $G$ follows the initialization.

What this does \emph{not} establish: alignment is a correlation between channel-\emph{mean} energy maps, so it measures agreement in spatial layout, not reproduction of the individual oriented channels, and a null on it would not rule out a finer imprint. The two control rows are single cells.

\begin{figure}[pos=htbp]
\centering
\includegraphics[width=0.92\linewidth]{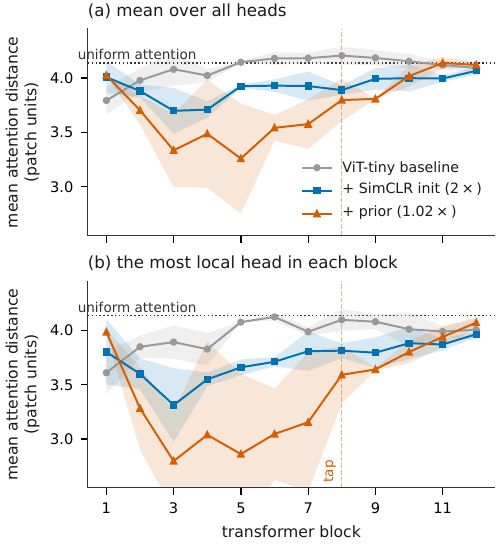}
\caption{What the prior gives a small transformer: locality. Mean
attention distance per block, ViT-tiny on CIFAR-100 at 10\%, three seeds
(band: $\pm$1 SD). The dotted line is the distance a \emph{uniform} attention
map would give on this $8\times8$ token grid ($4.14$ patch units), computed, not assumed. The baseline sits on that line at every depth, never
learning to attend locally from 5{,}000 images; the prior pulls the middle blocks well below it, most sharply \emph{upstream} of the tap. A SimCLR initialization at $2\times$ the compute does the same, more weakly.}
\label{fig:attention}
\end{figure}

The same question applies to the backbone, where the prior matters most, and the answer is unusually legible. A vision transformer has no built-in notion that nearby patches belong together; the standard diagnostic for whether it has learned one is mean attention distance. Figure~\ref{fig:attention} measures it for ViT-tiny at CIFAR-100 10\%: the baseline's heads straddle the uniform-attention distance at every depth, never discovering locality on 5{,}000 images, while the prior pulls the middle blocks well below it. The prior does not merely improve the features; it supplies the spatial inductive bias the architecture lacks, which a convolutional backbone gets for free and is why the same intervention is worth several times more here.

We had predicted this effect in the \emph{early} blocks, decaying with depth; instead it peaks at blocks~3--7 and the curves rejoin by block~11. Interpretable after the fact (the target is regressed from block~8 and the locality appears in the blocks that feed it), but not predicted, and recorded as a missed call. A SimCLR initialization produces the same signature at about half the depth-integrated magnitude, consistent with the two being substitutes on this backbone.

We are deliberate about how far the qualitative evidence goes. The individual-image maps of Figs.~\ref{fig:heatmap} and \ref{fig:cam} are selected on cases the prior gets right and the baseline wrong, so they illustrate rather than demonstrate; a t-SNE of the frozen features moves the all-class silhouette only from $-0.072$ to $-0.064$, since a $+6.3$-point feature gap need not be legible in two projected dimensions, and we do not print that panel as evidence. The alignment statistic and the linear evaluation are the measurements; the pictures illustrate them.

\section{Fusion outcomes: stack, substitute, interfere}\label{sec:fusion}

The grid's central comparative finding is that fusion outcomes are organized by the \emph{currency} each information source supplies: measured from each source alone, it separates the three outcomes after the fact, though comparing those measurements did not predict unseen combinations when we tested exactly that (Sec.~\ref{sec:procedure}).

\textbf{What we mean by currency, operationally.} The word is a metaphor used once as one; what it stands for is a measurement from single-source quantities only, so nothing in it requires training the combination. A source's currency is characterized from that source \emph{alone}: the frozen-feature gain $G$ it produces under the declared protocol, together with \emph{what} it improves, that is, which test images its arm fixes and how similar its learned representation is to the other source's, both measured on the single-source arms (Sec.~\ref{sec:samecurrency}). Two sources trade in the same currency when, at a matched cell, their $G$ values agree \emph{and} their single-source arms agree, on images fixed and in representation, about as closely as two seeds of one source do; both halves are necessary because two representations can raise linear accuracy equally while encoding different things. The combination's $G$ appears nowhere in the definition; it is what the definition \emph{predicts}, same currency implying that the combination's $G$ will not exceed the better single source, and Secs.~\ref{sec:substitute} and~\ref{sec:samecurrency} test that prediction. $G$ is a cheap ordinal proxy for source overlap, not an estimate of the unique and redundant atoms that partial information decomposition formalizes (Sec.~\ref{sec:fusiontheory}); what it buys is measurability on each source alone, at the cost of one linear fit, which is what makes a prospective rule possible at this grid's scale. Retrospective analyses characterize source overlap on the held-out test set; any prospective use must estimate the same diagnostics on validation data only, as Sec.~\ref{sec:procedure} does.

This section presents the three outcomes in order of how favorable they are: different currencies can compound, differing currency being necessary while architecture and intervention strength decide how much is realized (Sec.~\ref{sec:stack}); the same currency adds nothing, with the refinement that rescues the rule from its apparent counterexamples (Sec.~\ref{sec:substitute}); and one fusion, at full shaping strength, is uniformly destructive, with the feature-level measurement that explains why (Sec.~\ref{sec:taxsub}). Section~\ref{sec:taxonomy} collects the three into a single table with the evidence for each row.

\subsection{Different currencies can stack}\label{sec:stack}
DeiT-strength augmentation supplies invariance to nuisance transforms; the prior supplies oriented-energy structure. Fusing both (Table~\ref{tab:vitenvelope}) does not trade off: the prior's ViT gain grows under augmentation on both populations, by $1.5$--$2.4\times$ on CIFAR-100 and $1.4$--$2.4\times$ on Tiny-ImageNet across $1$--$25\%$. We report the two ranges separately because we once reported one: a registered ``$1.8$--$2.4\times$ throughout'' missed low on Tiny-ImageNet at 5, 15, and $25\%$, so amplification is population-dependent, and the CIFAR-100 range was overgeneralized. On Tiny-ImageNet it also vanishes at full data ($0.92\times$), where both interventions are decaying anyway. The feature side confirms the mechanism: the prior's $G$ \emph{rises} under augmentation (from $14.8$ to $22.2$ at 10\%), because once nuisance variation is handled, the network can commit more capacity to the structure the target teaches. Our own pre-registered bet here was wrong in the pessimistic direction: we predicted partial substitution and measured strong complementarity.

\paragraph{How far that reaches, on a convolutional backbone.} Every cell above is a ViT, on CIFAR-100 (the selection set) or Tiny-ImageNet (no part in selection, Sec.~\ref{sec:selection}). What the result does not span is a convolutional backbone off the selection set, so we ran the three missing fusion arms there, augmentation alone, prior with augmentation, and prior with a SimCLR initialization, on EuroSAT and Food-101 with ResNet-18 at 5, 10 and $25\%$. The substitution result transplants; the stacking result does not.

\looseness=-1 Prior with SimCLR falls below the better single source at all six cells ($-0.14$ to $-2.41$), so on these convolutional populations it does not merely fail to add; it costs, and the registered falsifier for the substitution rule, a combination beating the better single by ${\geq}1.5$, fired nowhere. Prior with augmentation stacks at exactly one cell of six, Food-101 at $5\%$ ($+4.32$, eleven standard errors), and is negative at the other five ($-0.37$ to $-3.07$). That tripwire required failure on \emph{both} populations, so it did not fire on the letter; we report it as a narrowing rather than a pass. Prior-with-augmentation stacking is therefore a result about attention backbones, holding on both the selection population and an off-selection one, plus a single convolutional cell; it is not a general property of the two currencies.

\looseness=-1 What holds across these cells and the schedule control of Sec.~\ref{sec:taxsub} concerns strength. Adding the prior at full weight costs whenever the partner is already strong: where augmentation is harmful, the prior leads (EuroSAT at $5\%$, augmentation alone $-7.67$ against the prior's $+0.84$), and where augmentation is strong, the prior subtracts (Food-101 at $10\%$, augmentation alone $+9.26$, adding the prior $-1.81$). Frozen-feature evaluation of the same arms agrees: measured against the same partner alone, the prior moves $G$ by $+4.68$, $-1.22$ and $-1.51$ on Food-101 over augmentation at 5, 10 and $25\%$, and by $-1.38$, $-2.29$ and $-1.27$ over the SimCLR initialization, so the cost is feature-side here as for the transfer tax. One more thing the attention-only evidence had hidden: DeiT-strength augmentation is not universally good, worth $+9.45$, $+9.26$ and $+3.05$ alone on Food-101 but $-7.67$, $-0.14$ and $+0.11$ on EuroSAT, so a claim about augmentation as a fusion source has to name its domain.

\subsection{Same currency substitutes}\label{sec:substitute}

\looseness=-1 SimCLR's objective \emph{is} invariance to an augmentation family. Three measurements pin the consequence. First, prior and SimCLR never stack: on convolutional backbones the combination costs $-1.5$ points against SimCLR alone, and on ViT it is exactly neutral against the better single arm. Their feature gains are the same gain ($G = 14.8$ against $13.5$ at the same cell), so the second source has nothing left to add. Second, the head-to-head flips with the recipe (Table~\ref{tab:flip}): under plain augmentation SSL overtakes the prior once data is plentiful, but under the DeiT recipe, the recipe anyone training a small ViT would actually use, the prior wins at every CIFAR-100 fraction, by ${\sim}5$ points across the mid-band, and the feature side shows why: augmentation lifts the prior's $G$ by $+7.4$ but SimCLR's by only $+3.8$ at 10\%. The same flip reproduces on Tiny-ImageNet in the low-data band. Third, SimCLR was not under-equipped: giving its contrastive views DeiT strength makes it \emph{worse} at every fraction (by $4$--$12$ points e2e, halving its $G$), so its standard configuration was its strongest available one.

\begin{table}[pos=htbp]
\caption{Prior versus SimCLR initialization ($2\times$ compute) on ViT-tiny, CIFAR-100 (the selection set, Sec.~\ref{sec:selection}), and the effect of recipe. Under plain augmentation, SSL catches the prior at ${\sim}5$k images; under the DeiT recipe, the prior wins at every fraction because the recipe supplies the invariance SSL was selling. Three seeds per cell, six on the DeiT-recipe $5\%$ arm. A positive entry is a fraction where the prior wins; no better-direction marker is used, since neither sign of a method-versus-method difference is better a priori. Per fraction the recipe under which the prior stands better is bold; with two rows the second-best and third-best marks do not apply.}
\label{tab:flip}
\centering\scriptsize
\setlength{\tabcolsep}{2.6pt}
\zebra{4}
\begin{tabular}{lrrrrr}
\toprule
 & \multicolumn{5}{c}{prior $-$ SimCLR (points)} \\
\cmidrule(l){2-6}
Recipe & 1\% & 5\% & 10\% & 25\% & 100\% \\
\midrule
Plain crop+flip & $+0.7$ & $+1.3$ & $0.0$ & $-1.4$ & $-3.9$ \\
DeiT augmentation & $\mathbf{+2.5}$ & $\mathbf{+7.2}$ & $\mathbf{+5.2}$ & $\mathbf{+5.0}$ & $\mathbf{+2.0}$ \\
\bottomrule
\end{tabular}
\end{table}

The substitution rule refines rather than universalizes: fusing the prior onto \emph{ineffective} SSL stacks trivially. SimSiam at its published budget learns almost nothing at study scale ($\Delta \leq +0.9$ at every CIFAR-100 and CIFAR-10 fraction, ${\leq}+1.8$ on any population), and prior-on-SimSiam recovers the prior's full gain ($+6.2$ and $+4.3$ over SimSiam alone at CIFAR-100 5 and 10\%, and $+2.0$ at Tiny-ImageNet 5\%). The operative rule is therefore \emph{the prior is redundant exactly insofar as the other source already bought the same features}, judged by the other source's own measured gain, not by its family name.

\looseness=-1 The grid contains the intermediate case, and it contradicts the binary phrasing of Table~\ref{tab:taxonomy}'s second row while supporting the rule just stated. On ViT-tiny at CIFAR-100, the prior with a DINO initialization beats the better single arm by $+0.92$ at $10\%$ ($30.66$ against the prior's $29.74$, about five standard errors) and $+1.02$ at $5\%$ ($22.62$ against $21.60$), and a combination exceeding both singles is what that row says cannot happen. The refined rule predicts it: DINO is a partial source here, its feature gain $+7.90$ roughly half the prior's $+14.83$, so it has not bought the same features and work is left for the prior to do. The feature side confirms rather than merely permits that reading, since the combination's $G$ is $+14.77$, level with the prior alone, not above it: the two still substitute in feature space even as the combination edges ahead end to end. We therefore state the row as \emph{combination $\leq$ best single when the partner's $G$ matches the prior's, and partial when it does not}, and we count these cells as the case that forced the qualification.

That qualification is not sufficient either. Three further pairs, trained for the prospective test of Sec.~\ref{sec:procedure} and so never used to shape any statement here, exceed their better single arm by $+1.04$ to $+1.43$ points at $2.3$ to $3.8$ standard errors, two of them with partner feature gains that \emph{do} match the prior's. What survives is the magnitude, not the inequality: across these combinations the result sits within a point or two of the better single arm either way, against single-arm gains of $6$ to $15$, so same-currency fusion buys or costs a rounding error where different-currency fusion buys multiples. We retire any exact-inequality phrasing of the row.

\subsection{Fusing into mature features interferes}\label{sec:taxsub}

The one uniformly destructive fusion in the grid is instructive (Table~\ref{tab:tax}). The prior's early shaping at full strength, so valuable from scratch, \emph{erases} mature ImageNet features. We call the measured degradation an \emph{interference tax}, and the tax runs $-15.2$ to $-16.8$ on the two datasets where transfer was worth $+10$ to $+18$ on the cells of Table~\ref{tab:tax}, decays to zero at full data (where the initialization no longer carries the performance), and, the decisive measurement, lands almost entirely on $G$. Where the initialization had little to offer (histopathology), there is nothing to destroy and the tax is a measured zero. Fusion with an information-rich partner must respect what the partner already encodes; a shaping prior does not.

\begin{table}[pos=htbp]
\caption{The interference tax of the ImageNet-transfer fusion, and its
feature-side origin. Injecting the prior on top of ImageNet weights \emph{at full auxiliary strength} is destructive in proportion to what the initialization supplied; linear evaluation shows the loss is carried almost entirely by $G$ (features), not by the \readout. PathMNIST, where ImageNet features barely transfer, is the null control. Ten seeds on the two CIFAR pairs end-to-end, three on PathMNIST, and three for every linear evaluation, so the \readout{} column differs by a ten-seed $\Delta$ against a three-seed $G$. The text below reports the same three cells at reduced $\lambda_0$, where the tax disappears.}
\label{tab:tax}
\centering\scriptsize
\setlength{\tabcolsep}{2.6pt}
\zebra{2}
\begin{tabular}{lrrrrr}
\toprule
Dataset & Frac. & Transfer adv. & $\Delta$ (tax)~\up & $G$~\up & \textit{Readout} \\
\midrule
CIFAR-10 & 5\% & $+9..{+}14$ & $-15.2$ & $-15.8 \pm 1.7$ & $+0.6$ \\
CIFAR-100 & 7\% & $+10..{+}18$ & $-16.8$ & $-16.3 \pm 3.1$ & $-0.6$ \\
PathMNIST & 10\% & $+1.8..{+}2.9$ & $-0.4$ & $+0.3 \pm 0.6$ & $-0.7$ \\
\bottomrule
\end{tabular}
\end{table}

Table~\ref{tab:tax} shows the three cells we could also probe, and three cells cannot carry a claim about a regime, so we state the population they come from: of the \taxCells{} transfer cells that carry the prior \emph{at full auxiliary strength from the first step} ($\lambda_0 = 1.0$, no delayed onset) at three or more seeds, \taxNegative{} have a negative $\Delta$, spanning nine datasets, median $\taxMedian$, worst $\taxWorst$; not one is positive. Full strength \emph{defines} that population rather than describing it, so the released records hold more aux-on-pretrained cells than the figure counts: the reduced-strength and delayed arms below are the \emph{control} for this claim, and counting them inside the population they exist to explain would remove the ``not one is positive'' by construction. What the \taxCells{} cells establish is that \emph{this} shaping schedule interferes, in proportion to what the initialization supplied, not that a structural prior interferes with mature representations however it is applied.

\paragraph{The tax is a property of the schedule, not of the fusion.} We ran that control on the three cells of Table~\ref{tab:tax} at $\lambda_0 = 0.3$, $0.1$ and $0.05$, plus a delayed arm holding the prior at zero for the first half of training, ten seeds per arm against ten-seed baselines on the two CIFAR cells and three on PathMNIST. The registered falsifier was that any arm reaching $\Delta \geq +0.5$ would show the tax to be schedule-dependent, and two of the four CIFAR-100 arms clear it.

\looseness=-1 The tax does not scale smoothly with $\lambda_0$; it essentially vanishes below full strength. On CIFAR-10 at $5\%$, $-15.2$ becomes $-0.18$, $+0.20$ and $+0.05$ at $\lambda_0 = 0.3$, $0.1$ and $0.05$; on CIFAR-100 at $7\%$, $-16.8$ becomes $+1.50$, $+0.53$ and $+0.21$. The two halves carry different weight: the disappearance is overwhelming, a $15$- to $18$-point move on both datasets, while the CIFAR-100 arms coming out nominally \emph{positive} is not established, pooling to $+0.70 \pm 0.61$, $1.1$ standard errors from zero, so we do not claim the prior helps a pretrained initialization. Deepening from three seeds to ten shrank six of the eight arms toward zero, every CIFAR-100 arm among them, the regression to the mean Sec.~\ref{sec:stats} warns about, on our own result. PathMNIST, where the initialization had nothing to supply, is unchanged across every arm ($-0.52$ to $+0.07$), as the currency account requires. The practical rule is therefore not ``never'' but ``not at full strength'': at $\lambda_0 \leq 0.3$ the fusion is neutral on the cells we tested.

One prediction in this control failed, and it bounds the mechanism. We expected the delayed arm between $\lambda_0 = 0.3$ and $0.1$, because the damage is done in the early high-rate phase. It is instead the worst arm below full strength on both datasets and negative on both, $-1.29$ on CIFAR-10 at $6.3$ standard errors (the one clearly resolved cell here) and $-0.31$ on CIFAR-100: withholding the prior and then applying it at full weight is worse than applying a weak one throughout, so timing is not the whole account and strength matters independently of it.

\subsection{The taxonomy}\label{sec:taxonomy}
\begin{table}[pos=htbp]
\caption{The retrospective fusion taxonomy the grid supports, with the measurement that establishes each row. The rule's call uses only single-source measurements, the two sources' frozen-feature gains and what each improves, and the measured outcome matches it in all five rows. Every call here is retrospective; the one prospective test of the rule, on pairs outside this table, is called one of nine (Sec.~\ref{sec:procedure}), so this is a taxonomy the rule organizes, not a prospective scoreboard. Each row is scoped to the evidence beside it, and the first row is the narrowest: on two convolutional populations outside the selection set, the same pairing stacks at one cell of six and mildly interferes at the rest (Sec.~\ref{sec:stack}).}
\label{tab:taxonomy}
\centering\scriptsize
\setlength{\tabcolsep}{2.6pt}
\zebra{2}
\begin{tabular}{P{0.195\columnwidth}P{0.205\columnwidth}llP{0.225\columnwidth}}
\toprule
Fusion & Currencies & Call & Measured & Key evidence \\
\midrule
Prior + augmentation & structure + invariance & stack & \textbf{stack} & $\Delta$ amplified $1.4$--$2.4\times$ (1--25\%); $G\uparrow$ (ViTs; 1 of 6 conv cells off-selection) \\
Prior + effective SSL & structure $\approx$ invariance$^{*}$ & substitute & \textbf{substitute} & combo near best single: $+1.4$ above (ViT) to $-2.4$ below (conv off-selection; Sec.~\ref{sec:substitute}) \\
Prior + ineffective SSL & structure + ${\sim}$nothing & stack & \textbf{stack} & full gain recovered on SimSiam \\
Prior + ImageNet init & structure vs.\ mature features & interfere & \textbf{interfere} & $-15$ to $-17$ e2e at $\lambda_0{=}1.0$, carried by $G$; null on shift; gone at $\lambda_0{\leq}0.3$ \\
Augmentation + SSL & invariance + invariance & substitute & \textbf{substitute} & DeiT recipe collapses SimCLR's margin \\
\bottomrule
\rowcolor{white}\multicolumn{5}{@{}p{0.95\linewidth}@{}}{\tiny $^{*}$Same currency in effect: at matched cells the two sources' feature gains coincide, which is why neither adds to the other.} \\
\end{tabular}
\end{table}

Table~\ref{tab:taxonomy} states the taxonomy compactly, and we are precise about its status. Across the measured combinations, pairs whose feature-level contributions substantially overlapped substituted rather than stacked, and different-currency pairs could stack when the strength and the architecture cooperated (Sec.~\ref{sec:stack}). Those single-source quantities separate the outcomes after the fact; comparing them did \emph{not} predict unseen combinations when we tested exactly that (Sec.~\ref{sec:procedure}).

\subsection{Same currency, same images}\label{sec:samecurrency}
Everything above infers currency from the \emph{magnitude} of a source's feature gain. That is indirect: two sources could each add fourteen points while improving entirely different images, and calling them the same currency would then be an artifact of comparing scalars. We tested it directly, with the prediction recorded first. Take three interventions on one backbone, dataset, and baseline (ViT-tiny, CIFAR-100 at 10\%): the prior, a SimCLR initialization, and DeiT-strength augmentation. The prior and SimCLR are the candidate same-currency pair because their single-source feature gains are close, whereas the prior and augmentation differ more in their feature-side effects; we therefore predicted, before computing the overlap diagnostics, that the first pair would agree more closely in the images it fixes and in representation geometry. Their neutral and stacking combinations are downstream outcomes, not inputs to this definition. Because easy images are fixed by everything, we normalize each cross-intervention agreement by that intervention's own agreement across seeds, the most same-currency comparison available.

\begin{figure}[pos=htbp]
\centering
\includegraphics[width=0.92\linewidth]{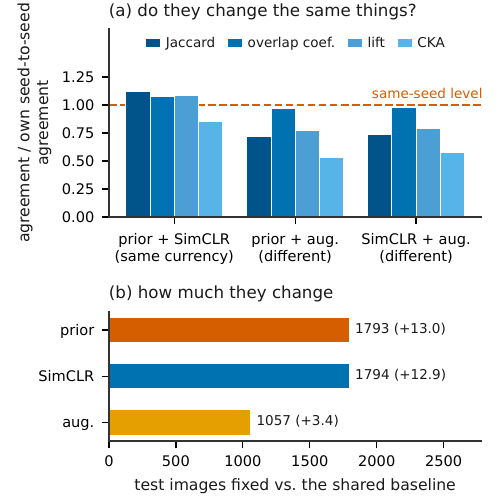}
\caption{Currency tested directly, not inferred from magnitudes. All arms are ViT-tiny on CIFAR-100 at 10\% with one shared baseline, so the only thing that varies is the fused source. \textbf{(a)} agreement between two interventions in \emph{which} test images they fix and \emph{which} representations they learn, each normalized by that intervention's own seed-to-seed agreement, so $1.0$ means the two sources differ by no more than two seeds of one of them. Four measures are shown because they do not agree on the effect size. \textbf{(b)} the number of images each intervention fixes: the two same-currency sources fix all but one the same number.}
\label{fig:currency}
\end{figure}

The prediction held on all four measures (Fig.~\ref{fig:currency}). The prior and SimCLR fix $1793$ and $1794$ test images, respectively, and their normalized agreement on \emph{which} images is $1.13$ against $0.72$ for the prior with augmentation; centered kernel alignment between their representations gives $0.85$ against $0.53$. A value above one has a direct reading: for the question of which images get fixed, it matters less whether you used a fixed spectral target or contrastive pre-training than which seed you used. That is what a substitute is.

Two qualifications. Augmentation fixes far fewer images ($1057$), and the Jaccard measure penalizes a small set for being small; on the size-robust overlap coefficient the separation shrinks to $1.08$ against $0.97$, so part of the headline gap is a magnitude effect and only the representation-level measure separates the pairs decisively. And this is one cell on one backbone; it corroborates the taxonomy where the taxonomy already had its strongest end-to-end evidence, instead of testing it somewhere new.

\emph{Takeaway.} What a source is worth depends on which currency is scarce, not on how modern the source is: the same free prior is the most valuable intervention we measure on an augmented ViT and the most damaging one on an ImageNet-initialized ResNet.

\section{Validation at scale}\label{sec:scale}

Everything above could, in principle, have been a small-image regularity. We therefore pre-registered a falsifier block and bought the two scale axes separately: ImageNet64 (1.28M images, 1000 classes, 64\,px, 13$\times$ the images and 5$\times$ the label space of anything else in the grid) and ImageNet-100 at native $224$\,px with a model-scale curve (ViT-S/16, ViT-B/16, ResNet-50, the ViTs under the standard DeiT recipe). Deviations (reduced epochs; native-resolution transforms) apply equally to both arms of every pair.

\begin{table}[pos=htbp]
\caption{The pre-registered scale block. Every falsifier was declared before the first run and none fired, with three qualifications kept in view: ViT-S's $G$ landed $0.05$ below its registered band; ViT-B's $|\Delta-G|$ of $3.12$ exceeds the registered $1.5$, on a cell whose evaluation-ceiling limitation was also registered in advance and whose residual is unresolvable; and ResNet-50's $G$, registered at $-0.7$ to $-0.2$, measured $+0.19$, a $0.39$ miss across zero on a quantity that is a measured zero either way ($|\Delta - G| = 0.21$). $G$ at ImageNet64 uses the fixed-budget evaluation (250 per class shown); ImageNet-100 the standard full-train evaluation. Three seeds per cell; the ImageNet-100 cells run $100$ epochs, and the text repeats both ViT pairs at $200$, where the gains are $+4.52 \pm 0.18$ and $+6.71 \pm 0.90$.}
\label{tab:scale}
\centering\scriptsize
\setlength{\tabcolsep}{1.8pt}
\zebra{2}
\begin{tabular}{P{0.160\columnwidth}P{0.165\columnwidth}rrrP{0.145\columnwidth}}
\toprule
Cell & Registered band & $\Delta$~\up & $G$~\up & $|\Delta{-}G|$~\down & Verdict \\
\midrule
\multicolumn{6}{@{}l}{\emph{ImageNet64: 1.28M images, 1000 classes}} \\
ResNet-18 & $\Delta$: $0.0{\pm}0.5$ & $+0.04$\sem{0.08} & $+0.05$ & $0.01$ & neutral, exactly \\
MobileNetV3 & (none: in bracket) & $+1.95$\sem{0.40} & $+1.81$ & $0.14$ & consistent \\
ViT-tiny & $\Delta \geq +3$ & $+3.24$\sem{0.84} & $+2.84$ & $0.40$ & deficit persists \\
Swin-T &--& \multicolumn{3}{l}{baseline collapses at chance} & stabilized by prior \\
\midrule
\multicolumn{6}{@{}l}{\emph{ImageNet-100 @ 224\,px, DeiT recipe on ViTs}} \\
ResNet-50 & $\Delta$: $0.0{\pm}1.0$ & $-0.02$\sem{0.26} & $+0.19$ & $0.21$ & neutral at 224\,px \\
ViT-S/16 & $\Delta$: $+2..{+}8$; $G$: $12.0..12.6$ & $+13.00$\sem{0.24} & $+11.95$ & $1.05$ & above band \\
ViT-B/16 & $\Delta \geq \Delta_{\mathrm{ViT\mbox{-}S}}$ & $+26.01$\sem{2.01} & $+22.89^{*}$ & $3.12^{*}$ & grows at full data \\
\bottomrule
\rowcolor{white}\multicolumn{6}{@{}p{0.95\linewidth}@{}}{\tiny $^{*}$The ViT-B aux arm evaluates at its own ceiling ($69.8$ vs.\ e2e $69.3$), a pre-registered limitation that compresses $G$; the residual is not resolvable against its uncertainty ($\pm 2.4$).} \\
\end{tabular}
\end{table}

Table~\ref{tab:scale} reports the block. Three conclusions.

\emph{Convolutional redundancy at sufficiency is exact and scale-stable.} ResNet-18 at 1.28M images gains $+0.04 \pm 0.08$ end-to-end and $+0.05 \pm 0.10$ feature-side; the neutral regime is measured to a hundredth of a point at the largest data scale in the study. ResNet-50 at native $224$\,px repeats it: the seed-paired difference on three matched seeds is $-0.02 \pm 0.26$, inside the pre-registered band ($0.0 \pm 1.0$), and the falsifier that would have made convolutional redundancy a low-resolution artifact ($\Delta \geq +1.5$) is ruled out.

\emph{The attention deficit is real, persistent under the measured budgets, and grows with model scale at full data.} ViT-tiny still gains $+3.2$ points at 1.28M images. At $224$\,px under the standard recipe and a shared $100$-epoch budget, ViT-S/16 gains $+13.0$ at $21.7$M parameters on a properly configured transformer, and ViT-B/16 gains $+26.0$, twice ViT-S (both repeated below at a doubled budget, where the ordering holds), with the baseline's seed variance three times the prior arm's. That asymmetry invites the objection that the gain is the optimization stabilization Sec.~\ref{abl:stab} treats as a separate property, so we settle it on the seeds: the ViT-B baseline runs $46.5$, $39.9$ and $43.6$, none near the $1\%$ chance level, so the cell does not trip the collapse rule of Sec.~\ref{sec:stats}, and the \emph{worst} prior seed still beats the \emph{best} baseline seed by $22$ points; nothing here is a rescued run. The pre-registered fork read: if the deficit is data-hunger, the bigger model should show the larger gain. At full data it does, by 13 points. Notably, the \emph{smaller} ViT with the prior still outperforms the bigger one ($78.4$ vs.\ $69.3$ at $100$ epochs, $83.99$ against $82.02$ at $200$): right-sizing plus structure beats scale.

\looseness=-1 That ordering was registered as a prediction across the whole envelope, with a falsifier if it inverted at any fraction at or above $5\%$. It inverts at $5$, $10$ and $25\%$ (Table~\ref{tab:inenv}), so that the falsifier fired and we report it (it also inverts nominally, unresolved, at $1$ and $2\%$). We do not read those cells as refuting the full-data result, for a property of the arms themselves: below $100\%$ the ImageNet-100 ViT-B baseline never trains, flat at $4.1$--$7.1\%$ on a 100-way task across a $12.5\times$ data range, so its gain there cannot carry a claim about model scale, and the ViT-S comparator carries baseline seed standard deviations up to $8.2$ points at exactly the fractions ($10$ and $25\%$) driving the inversion. The defensible claim is the narrower one: at full data the deficit grows with model scale, and across the low-data fractions of this population the comparison is not determined.

\paragraph{Is the ViT-B gain an under-trained baseline?} The $+26.01$ is the largest number in this paper and it is measured at $100$ epochs, where an 85.9M-parameter transformer on $126$k images may not have finished training. We doubled the schedule on both arms, six seeds each. The baseline gains $+31.98$ from the schedule alone, reaching $75.31 \pm 0.48$; the prior arm reaches $82.02 \pm 0.76$; the gain is $+6.71 \pm 0.90$, $95\%$ confidence interval $[4.94, 8.47]$, $7.4$ standard errors from zero. About a quarter of the headline survives a doubled schedule, so $+26.01$ is a $100$-epoch figure and should be read with $+6.71$ beside it.

\looseness=-1 We had registered a falsifier at $\Delta \leq +6$, which would have withdrawn the model-scale conclusion outright. The interval contains $6.0$, which sits $0.78$ standard errors below the estimate, so the falsifier is neither fired nor excluded, and we report it as exactly that. Two facts belong beside it. The estimate moved with seeds: $+7.79$ at three arms against two, $+7.79$ at three against three, and $+6.71$ at six against six, so a three-seed report would have published a clearance the data do not support. And one prior-arm seed, $78.58$, sits $4.6$ standard deviations below the other five ($82.71 \pm 0.90$); it is not a collapse under the rule of Sec.~\ref{sec:stats} and there is no principled reason to drop it, so it stays, and it is why this arm's uncertainty exceeds the baseline's. Excluding it would give $+7.39 \pm 0.63$ and put the threshold outside the interval, which is exactly why we keep it.

\paragraph{The trend at a budget where both models are trained.} The comparison motivating that trend, $+13.00$ against $+26.01$, is made at $100$ epochs, where the control above shows ViT-B still $+31.98$ short of where the schedule alone takes it, so we repeated ViT-S at $200$ epochs too, six seeds per arm. It reaches $79.47 \pm 0.16$ against $83.99 \pm 0.08$, a gain of $+4.52 \pm 0.18$, against ViT-B's $+6.71 \pm 0.90$. The ordering holds by $-2.19 \pm 0.92$, and the pre-registered falsifier, which would have withdrawn the model-scale claim everywhere had ViT-S matched or exceeded ViT-B, did not fire. The uncertainty is almost entirely ViT-B's (standard errors $0.48$ and $0.76$ against ViT-S's $0.16$ and $0.08$), so the ordering is clear at $2.4$ standard errors and not tighter. We tested it expecting it to invert.

\emph{The law's predictive form survives.} Five of six pairs tie $\Delta$ to $G$ within $1.1$ points; the sixth (ViT-B) carries a pre-registered evaluation-ceiling limitation and an unresolvable residual. That ceiling caveat belongs to every ImageNet-100 row, not to ViT-B alone: at these $100\%$ cells the evaluation's labels are the cell's own, so the $|\Delta - G|$ column is a gap read under the scope rule of Sec.~\ref{sec:probe}, not a signed \readout{}, which is why the cells stand here and not in the audit of Sec.~\ref{sec:lawaudit}; the ImageNet64 rows, evaluated at fixed per-class budgets, carry no ceiling. The ViT-S band, derived from the law before the evaluation ran, was hit on its edge ($11.95$ vs.\ $12.0$--$12.6$). As an additional robustness check, we independently re-ran the ViT-S and ResNet-50 evaluations on a second machine with a from-scratch environment; their means agree with the primary measurement to $0.01$--$0.05$ points.

\subsection{The envelope at ImageNet scale}\label{sec:scaleenvelope}

The block above measures the right-hand end of the envelope. Because that would leave the shape untested at scale, we also ran both ImageNet stages across fractions, on the same fixed subsets and with the same reduced-epoch budget as their $100\%$ cells (Table~\ref{tab:inenv}). This is the largest data-efficiency envelope in the study, and it reports a registered miss before it reports anything else.

\emph{The miss.} We registered an interior peak at $5$--$20\%$ for the convolutional envelope. ResNet-18 instead falls from the smallest runnable fraction, $+1.90$ at $1\%$ to zero by $10$--$15\%$, staying within $0.2$ of zero thereafter, with no interior peak: the falsifier fired. The cause is that we predicted in \emph{fraction} space when the operative axis is absolute data. One percent of ImageNet64 is $12{,}820$ images, which, in every smaller population in the grid, is already past the peak. The envelope is not absent at scale; its peak sits below the smallest fraction this dataset can express.

\begin{figure}[pos=htbp]
\centering
\includegraphics[width=0.92\linewidth]{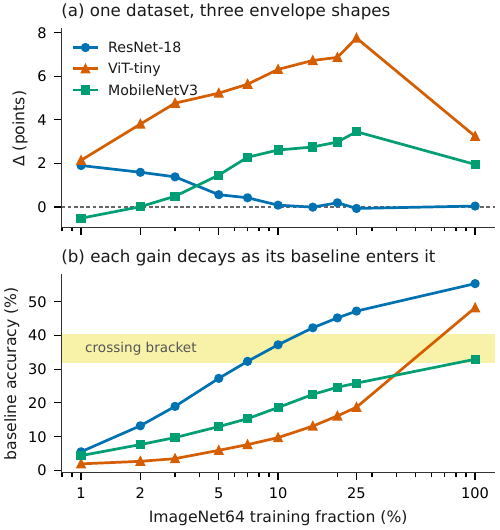}
\caption{The envelope at ImageNet scale, and why its shape is a property of the backbone. Dataset, recipe, fixed subsets and label space are held constant; only the backbone varies. \textbf{(a)} Three backbones, three envelope shapes: ResNet-18 falls to zero and stays there, while ViT-tiny and MobileNetV3 climb across the whole range. \textbf{(b)} The same baselines against the crossing bracket. Each gain decays as its own baseline enters the band, and the three arrive at very different fractions: ResNet-18 by $7$--$15\%$, MobileNetV3 only at $100\%$, ViT-tiny passing above it at $100\%$, exactly where its gain falls. Three seeds per cell.}
\label{fig:inenv}
\end{figure}

\emph{What the same cells then show, which we did not anticipate.} Holding dataset, recipe, subsets, and label space fixed and varying only the backbone gives three different envelope shapes (Fig.~\ref{fig:inenv}). ResNet-18 falls to zero and stays there; ViT-tiny and MobileNetV3 rise monotonically across the entire measured range, ViT-tiny from $+2.14$ to $+7.75$ at $25\%$. Under the account this is not three phenomena but one: the peak sits where the baseline crosses the \readout{} bracket, and only ResNet-18's baseline reaches it in range ($5.5 \to 47.2$, crossing $[31.8,40.3]$ between $7$ and $15\%$), which is exactly where its gain reaches zero ($+0.42$ at base $32.3$, $-0.01$ at base $42.3$). Across $1$--$25\%$ neither of the others comes near the band, and both are still climbing; extend to full data, and they arrive too, MobileNetV3 into it ($32.9$) and ViT-tiny above it ($48.2$), and both gains duly fall ($+3.45 \to +1.95$ and $+7.75 \to +3.24$). One rule, three apparent shapes: the envelope's peak is a property of where a \emph{backbone} sits on its own \readout{} curve, not of the dataset, and this is the cleanest demonstration of that in the paper, because the classification grid could only vary it by changing datasets too.

\begin{table}[pos=htbp]
\caption{The envelope at ImageNet scale: $\Delta$ by fraction, three seeds per cell, both arms sharing the reduced-epoch budget. Baseline accuracy is given beneath each gain. ImageNet64's three backbones see identical images, recipe, and label space, so the difference in envelope \emph{shape} between them is attributable to the backbone alone. The ImageNet-100 ViT columns below $100\%$ are reported with a caveat, not as headline cells: their baselines sit close to chance and carry seed standard deviations up to $8.2$ points, so the gains there are large but poorly determined. Every ImageNet-100 cell here uses the $100$-epoch budget; the ViT-B pair at $200$ epochs gives $+6.71$ (Sec.~\ref{sec:scale}). Best in bold, second best underlined, per fraction within each stage (the remaining entry is third).}
\label{tab:inenv}
\centering\scriptsize
\setlength{\tabcolsep}{2.2pt}
\zebra{5}
\begin{tabular}{lrrrrrr}
\toprule
& \multicolumn{3}{c}{ImageNet64, 64\,px} & \multicolumn{3}{c}{ImageNet-100, 224\,px} \\
\cmidrule(r){2-4}\cmidrule(l){5-7}
Frac. & R18 & ViT-Ti & MNetV3 & R50 & ViT-S & ViT-B \\
\midrule
1\%   & \snd{$+1.90$} & $\mathbf{+2.14}$ & $-0.52$ & $+0.54$ & $\mathbf{+2.31}$  & \snd{$+0.69$} \\
      & \tiny 5.5 & \tiny 1.9 & \tiny 4.3 & \tiny 11.5 & \tiny 7.3 & \tiny 5.3 \\
2\%   & \snd{$+1.59$} & $\mathbf{+3.80}$ & $+0.01$ & $+1.21$ & $\mathbf{+7.59}$  & \snd{$+5.11$} \\
      & \tiny 13.2 & \tiny 2.6 & \tiny 7.6 & \tiny 18.2 & \tiny 6.4 & \tiny 4.1 \\
3\%   & \snd{$+1.38$} & $\mathbf{+4.76}$ & $+0.49$ &--&--&-- \\
      & \tiny 18.9 & \tiny 3.4 & \tiny 9.7 &&& \\
5\%   & $+0.56$ & $\mathbf{+5.22}$ & \snd{$+1.46$} & $+1.84$ & $\mathbf{+14.15}$ & \snd{$+4.47$} \\
      & \tiny 27.3 & \tiny 5.9 & \tiny 12.9 & \tiny 36.5 & \tiny 7.3 & \tiny 6.1 \\
7\%   & $+0.42$ & $\mathbf{+5.63}$ & \snd{$+2.28$} &--&--&-- \\
      & \tiny 32.3 & \tiny 7.6 & \tiny 15.3 &&& \\
10\%  & $+0.08$ & $\mathbf{+6.31}$ & \snd{$+2.61$} & $+1.74$ & $\mathbf{+14.23}$ & \snd{$+2.30$} \\
      & \tiny 37.2 & \tiny 9.7 & \tiny 18.6 & \tiny 55.7 & \tiny 11.9 & \tiny 6.6 \\
15\%  & $-0.01$ & $\mathbf{+6.72}$ & \snd{$+2.75$} &--&--&-- \\
      & \tiny 42.3 & \tiny 13.1 & \tiny 22.5 &&& \\
20\%  & $+0.19$ & $\mathbf{+6.86}$ & \snd{$+2.98$} &--&--&-- \\
      & \tiny 45.2 & \tiny 16.1 & \tiny 24.6 &&& \\
25\%  & $-0.07$ & $\mathbf{+7.75}$ & \snd{$+3.45$} & $+2.30$ & $\mathbf{+30.39}$ & \snd{$+7.36$} \\
      & \tiny 47.2 & \tiny 18.7 & \tiny 25.8 & \tiny 72.5 & \tiny 23.2 & \tiny 7.1 \\
100\% & $+0.04$ & $\mathbf{+3.24}$ & \snd{$+1.95$} & $-0.02$ & \snd{$+13.00$} & $\mathbf{+26.01}$ \\
      & \tiny 55.4 & \tiny 48.2 & \tiny 32.9 & \tiny 85.9 & \tiny 65.4 & \tiny 43.3 \\
\bottomrule
\end{tabular}
\end{table}

\looseness=-1 Two further readings. The convolutional column reaches sufficiency inside the measured range and stays there, so the neutrality result of Table~\ref{tab:scale} is not a single point but a plateau from $10\%$ of ImageNet64 onward. And ResNet-50 at $224$\,px is the one convolutional column that is \emph{positive} in the low-data band ($+0.5$ to $+2.3$ from 1 to $25\%$) before returning to neutrality at full data. We had registered $+1$ to $+6$ across 1--5\% with a decay through $25\%$, and both halves missed: the $1\%$ cell lands at $+0.54$, below the band, and the column rises instead of decaying ($+1.74$ at $10\%$, $+2.30$ at $25\%$). What held is the direction and the falsifier, which would have fired had $\Delta$ been non-positive everywhere: convolutional gains do survive native resolution and are a low-data, not a low-resolution, phenomenon, but we did not call their size or their shape.

\emph{Takeaway.} Scale does not retire the prior: convolutional redundancy at sufficiency survives 1.28M images, while attention's deficit grows with model size, not shrinking. Filling the envelope at scale cost us a registered prediction and returned a sharper claim in its place: which fraction the peak sits at is a property of the backbone's position on the \readout{} curve, not of the dataset.

\section{Buying the comparator more compute}\label{sec:budget}

\looseness=-1 Cost normalization makes the comparisons in Sec.~\ref{sec:law} interpretable, and Sec.~\ref{sec:parity} states its price: a comparator held to its published budget inside our recipe may be undertrained, not weak, and the two are not separable by construction. We measured that reservation rather than leave it as a limitation, re-running both self-supervised comparators at four times their pre-training budget and at every fraction from 1 to 25\%, since every other effect here moves with the data regime. Table~\ref{tab:budget} reports it, and Secs.~\ref{sec:budget:simsiam}--\ref{sec:budget:features} work through it. The objection is confirmed outright: it costs us an accuracy claim on convolutional backbones, attaches a budget and a data regime to the claim that survives on attention, and replaces one account of our two populations' difference with another.

\begin{table}[pos=htbp]
\caption{Buying the comparators more compute. CIFAR-100 (the selection set, Sec.~\ref{sec:selection}), ResNet-18, gain over the baseline by data fraction; three seeds per pre-trained cell, ten for the baseline and the prior at 1, 5 and 10\%. SSm is SimSiam and SC is SimCLR, at 200 pre-training epochs ($2\times$ baseline compute) or 800 ($5\times$); the prior costs $1.02\times$. The last two columns give the prior minus each $5\times$ comparator, where positive means the free prior still wins.}
\label{tab:budget}
\centering\scriptsize
\setlength{\tabcolsep}{2.6pt}
\zebra{6}
\begin{tabular}{lrrrrrrr}
\toprule
 & \textbf{Prior} & \multicolumn{2}{c}{$2\times$ SSL} & \multicolumn{2}{c}{$5\times$ SSL} & \multicolumn{2}{c}{prior $-$ $5\times$} \\
\cmidrule(lr){3-4}\cmidrule(lr){5-6}\cmidrule(l){7-8}
Frac. & $\mathbf{1.02\times}$~\up & SSm~\up & SC~\up & SSm~\up & SC~\up & SSm & SC \\
\midrule
1\%  & $+1.42$ & $-0.02$ & $+2.27$ & $-0.09$ & $\mathbf{+4.54}$ & $+1.51$ & $-3.12$ \\
2\%  & $+2.50$ & $-0.11$ & $+4.88$ & $+0.19$ & $\mathbf{+8.33}$ & $+2.31$ & $-5.83$ \\
3\%  & $+3.68$ & $+0.11$ & $+5.99$ & $+0.54$ & $\mathbf{+10.82}$ & $+3.14$ & $-7.14$ \\
5\%  & $+5.15$ & $+0.13$ & $+9.05$ & $+1.90$ & $\mathbf{+14.38}$ & $+3.25$ & $-9.23$ \\
7\%  & $+4.87$ & $+0.87$ & $+9.86$ & $+3.39$ & $\mathbf{+13.88}$ & $+1.48$ & $-9.01$ \\
10\% & $+3.75$ & $+0.61$ & $+8.76$ & $\mathbf{+4.33}$ & $\mathbf{+10.92}$ & $-0.58$ & $-7.17$ \\
15\% & $+2.55$ & $+0.17$ & $+6.64$ & $+3.65$ & $\mathbf{+8.36}$ & $-1.10$ & $-5.81$ \\
25\% & $+0.16$ & $-0.03$ & $+2.81$ & $+0.79$ & $\mathbf{+2.77}$ & $-0.63$ & $-2.61$ \\
\bottomrule
\end{tabular}
\end{table}

\subsection{The near-null was a budget artifact}\label{sec:budget:simsiam}
SimSiam at its published 200-epoch budget gains nothing anywhere on the envelope, $-0.11$ to $+0.87$. At 800 epochs it gains up to $+4.33$. We therefore withdraw the statement that negative-free self-supervision learns almost nothing at this scale: it is budget-starved at the cost we normalized to, and its published default is a poor guide to what it can do on a few thousand images. Its budget response is data-dependent: \looseness=-1 buys most in the mid band ($+3.7$ at 10\%) and nothing at 1\%.

That data-dependence is why the envelope was necessary. The full envelope shows a crossover: the prior wins from 1 to 7\% by up to $3.25$ ($2.6$ to $12.2$ standard errors) and \emph{loses} from 10 to 25\%, most clearly at 15\% ($-1.10$, $5.2$). A read at 5 and 10\% alone, where the objection is sharpest, would have missed a resolved reversal, the same sparse-grid error we identify in our own earlier readings.

\subsection{The convolutional case becomes a cost case}\label{sec:budget:conv} Against the stronger comparator, the prior loses at every fraction, including the scarcest. SimCLR at $5\times$ beats it by $3.12$ to $9.23$ points, at 12 to 37 standard errors, from 500 images to 7{,}500 (1--15\%), and by a narrowing $2.61$ at 25\% as both interventions decay toward sufficiency. Our earlier positioning noted near-parity with self-supervision at 1\%; that was a statement at $2\times$, and it does not survive $5\times$. On convolutional backbones, the prior's case is therefore a cost case and not an accuracy case, and we say so without qualification.

\subsection{Attention, given the same test}\label{sec:budget:vit}
Leaving the attention comparison at $2\times$ while raising the convolutional one to $5\times$ would reproduce exactly the asymmetry this section exists to remove, so we ran it there too. Against ViT-tiny under the DeiT recipe, $5\times$ SimCLR gives $30.45$, $40.57$ and $55.85$ at 5, 10 and 25\%, against the prior's $28.87$, $41.29$ and $57.32$. The comparator \emph{wins} at 5\% by $1.57$ ($3.9$ standard errors, six seeds), the two are level at 10\% ($+0.71$, $1.2$), and the prior wins at 25\% by $1.47$ ($7.2$). The prior is not overturned on attention as it is on convolutions, but the claim must name its budget: at $2\times$ the prior leads at every CIFAR-100 fraction, and at $5\times$ only in the higher-data band. Section~\ref{sec:budget:tin} shows the same boundary on a second population.

That shape contradicted a prediction recorded before the runs: we expected the prior to hold at the scarce fractions and 25\% to be the cell most likely to flip, and the reverse happened. What the extra budget buys shows why: $+8.80$, $+4.51$ and $+3.53$ at 5, 10 and 25\%, decreasing with data, because 200 epochs over 2{,}500 images is fewer than four thousand steps: the comparator was step-starved at 5\%, not data-starved (Sec.~\ref{sec:dataopt}). On convolutional backbones, the same increment peaks in the mid band, so the budget response is itself backbone-dependent.

\subsection{A second population, and a failed account}\label{sec:budget:tin} Repeating the comparison on Tiny-ImageNet gives the same ordering at every matched fraction. Writing the prior minus a $5\times$-budget SimCLR initialization, both under DeiT augmentation on ViT-tiny, CIFAR-100 gives $-1.57$, $+0.71$ and $+1.47$ at 5, 10 and 25\%, and Tiny-ImageNet gives $-1.35$, $-0.31$ and $+1.71$. The comparator wins at 5\% on both ($3.9$ and $3.7$ standard errors), the two are level at 10\% on both, and the prior wins at 25\% on both ($7.2$ and $5.1$). So at five times the comparator's pre-training budget the prior leads in the higher-data band and loses in the scarcer one, and that pattern is not specific to a dataset. The ordering tracks the data fraction, not the absolute image count: $5{,}000$ images is 10\% of CIFAR-100, but 5\% of Tiny-ImageNet, and the two disagree there while agreeing at every matched fraction.

Getting there cost us a prediction and an account. With only the 5 and 10\% cells measured, where the comparator was ahead and level, we wrote that the prior led nowhere on this population, predicted the 25\% cell at $40$--$44$ explicitly against our own method, and expected the two populations to line up by image count. The cell landed at $36.53$, below the band; the recorded falsifier fired; and the matched-fraction table shows why the account was wrong: matching on image count compares different points on two different learning curves, and what determines which source wins is where a cell sits on its own dataset's curve. The ``different stories'' reading was an artifact of a two-point grid, the same error this section has already documented once.

One measurement in this pass we report without explaining. At Tiny-ImageNet 25\%, four times the pre-training made the comparator \emph{worse}: $38.48$ at $2\times$ against $36.53$ at $5\times$, a drop of $1.95 \pm 0.42$. It is the only cell in the study where extra pre-training budget costs accuracy, and it echoes our earlier finding that DeiT-strength contrastive views also hurt. We note it and decline to fit an explanation to one cell.

\subsection{What the budget actually buys}\label{sec:budget:features}
The comparison so far is end-to-end, while the taxonomy is a claim about \emph{what} a source supplies, not how much accuracy it yields, so we probed these cells under the frozen-feature protocol used throughout. On CIFAR-100 the feature gain of $5\times$ SimCLR is $+22.24$, $+21.27$ and $+20.77$ at 5, 10 and 25\%, against the prior's $+21.37$, $+22.19$ and $+23.05$. The two orderings agree cell by cell: the comparator's feature gain is larger where it wins end-to-end, level where the two are level, and smaller where the prior wins. The prior's feature gain \emph{rises} with data while the comparator's \emph{falls}, and they cross between 5 and 10\%, which is where the accuracy ordering crosses. Extra pre-training budget therefore buys more of the same currency, not a different one, and the flip is a property of the representations, not of the classifier reading them. This is the substitution outcome measured on a budget axis, the one axis the taxonomy had not been tested on.

\looseness=-1 Including the second population strengthens the agreement and supplies the mechanism the image-count account lacked. Across both datasets and all six measured fractions, whichever intervention has the larger frozen-feature gain wins end-to-end: the sign of $G_{\mathrm{SSL}} - G_{\mathrm{prior}}$ matches the sign of the accuracy difference six times out of six. The clearest case is the cell that overturned our own prediction. At Tiny-ImageNet 25\%, where the prior wins by $1.71$, its feature gain is $+14.27$ against the comparator's $+12.43$; that difference of $1.84$ agrees with the accuracy difference to within $0.13$, so the outcome is feature-side almost entirely. The population term is equally unmysterious once measured. At a matched 5{,}000 images, the two datasets diverge because the \emph{prior's} contribution diverges: its feature gain is $+22.19$ on CIFAR-100 against $+15.67$ on Tiny-ImageNet, while the comparator's is $+21.27$ and $+17.72$. Moving to the 200-way $64$\,px task costs contrastive pre-training about $3.5$ points of feature gain and the fixed spectral target about $6.5$: the hand-crafted prior supplies proportionally less of what a finer label space at higher resolution requires, and that, not the number of images, decides which source wins.\footnote{The Tiny-ImageNet comparator probes were computed on a different machine from the cells they pair with. Cross-machine agreement under this protocol was verified elsewhere in the study to within $0.05$ points, but we note the mixed provenance instead of leaving it implicit.}

The second comparator extends the same agreement. Probing SimSiam at $5\times$ across all eight fractions gives a feature gain that sits below the prior's from 1 to 7\% ($-5.3$ to $-1.9$ points of difference), crosses at 10\%, and rises above it at 15\%, matching the end-to-end ordering reported in Sec.~\ref{sec:budget:simsiam} at seven of eight cells; the one mismatch is the crossing cell itself, where the feature difference is a near-zero $-0.08$. Across two self-supervised methods and two backbone families, the source with the larger frozen-feature gain wins.

\looseness=-1 One result in this pass we report without an account. Probing the convolutional budget envelope shows the prior's own residual tracking the fitted branch closely ($-2.74$, $-2.65$, $-2.07$, $-1.11$ below the crossing, $+0.06$ and $+0.21$ inside the bracket, and $-0.18$, $-0.28$ above it), while $5\times$ SimCLR departs from it, running positive from 3 to 15\% ($+1.32$, $+3.11$, $+3.58$, $+2.23$, $+1.81$) at baselines where the branch is negative or near zero. Two things narrow what the anomaly can be. It is not a property of self-supervised initialization as such: $5\times$ SimSiam, initialized the same way and evaluated identically, stays within $[-0.42, +0.86]$ at every fraction. And across all sixteen self-supervised convolutional cells, the residual correlates with the size of the intervention itself ($r = +0.52$ between $\Delta$ and the residual), so what departs from the branch is large gains rather than a particular method. We recorded in advance that these initializations lie outside the scope in which the branch was estimated, so we do not count the departure against it, and we prefer to leave it as a measured regularity rather than to fit an explanation to sixteen points.

\emph{Takeaway.} A comparator's budget is part of the claim. Two of our own accuracy claims did not survive this section and are now cost claims; the finding that did survive is that heavy augmentation, not the prior, is what collapses contrastive pre-training's advantage.

\section{Fusing genuinely different sources}\label{sec:multisource}

Everything to this point fuses one knowledge source into one image stream. This section asks whether the same rule governs fusion in the sense the term usually carries: two \emph{sensors} (Sec.~\ref{sec:multisensor}) and two \emph{classifiers} (Sec.~\ref{sec:decisionfusion}). Both are tests of transfer, not restatements: the rule was derived from training-time fusion of a prior into a network, and neither of these is that.

\subsection{Two sensors, and when a second one pays}\label{sec:multisensor}
We build two multi-source populations that differ strongly in both source asymmetry and modality; their contrast is informative, but the design does not identify which factor is causal.

\textbf{EuroSAT-MS} splits \emph{one} instrument: Sentinel-2's 13 bands become \textbf{visible} (3), \textbf{non-visible} (the other 10, spanning red-edge, near-infrared, short-wave infrared and cirrus) and \textbf{sensor-fused} (all 13). \textbf{So2Sat LCZ42} pairs \emph{two satellites}: Sentinel-1 synthetic-aperture radar (8 channels) with Sentinel-2 optical (10 channels), over 17 local-climate-zone classes. The second is the harder and more informative test, since radar backscatter and optical reflectance share no detector and no physics. In both, tiles and fixed subset indices are identical across sources, and the configuration is the reference configuration verbatim, with only the source changed, so only the input varies.

\emph{One instrument's bands are complementary; two satellites are not.} Fusing EuroSAT's band sets beats the better single source at every fraction, by $+0.45$ to $+2.96$. Fusing So2Sat's two satellites never does: the 18-channel arm lands between $-0.73$ and $\pm0.00$, at or below optical alone throughout. Radar is not uninformative in isolation, reaching $35$--$53\%$ against a $5.9\%$ chance rate, but adding it to optical buys nothing here. Had we run only the multispectral population, we would have concluded that a second sensor stacks, and reported a result that the first genuinely cross-modality test contradicts.

\emph{What separates them, and what we cannot separate.} The accuracy asymmetry between the two sources is $0.11$--$6.68$ points on EuroSAT-MS and $25.98$--$33.77$ on So2Sat, and the population with the matched arms is the one where fusion pays, which is consistent with Eq.~\ref{eq:ensgain}, our decision-level fit on CIFAR-100 ensembles, dominated not by diversity but by exactly this asymmetry term.

\looseness=-1 A correlation of $r = -0.80$ between asymmetry and fusion gain across the ten cells of Table~\ref{tab:sensorfusion} does not survive inspection, and we do not report it as one. The arithmetic is right, but the statistic is not: within EuroSAT-MS the correlation is $+0.08$, and within So2Sat it is $-0.05$, so it is carried entirely by the gap between two clusters that do not overlap on either axis, and its effective sample size is two populations, not ten points. Those two populations also differ in \emph{two} ways at once, since the low-asymmetry population splits one instrument's bands while the high-asymmetry one pairs two satellites with no shared detector or physics. Nothing in this design separates ``the sources are too unequal'' from ``the sources are cross-modal''. What the experiment supports is therefore a contrast, not a predictor. Distinguishing the two accounts would need a third population that breaks the confound, such as two well-matched cross-modal sources or two badly matched same-instrument ones, and we have not run it.

\begin{table}[pos=htbp]
\caption{Does a second source pay? Fusion gain is the fused arm minus the better single source, at matched data. Sorted by data fraction. Three seeds per cell, ResNet-18; the reference configuration is used verbatim throughout. Splitting one instrument's bands pays at every fraction; pairing two satellites does not pay at any. Best in bold, second best underlined, third best in italics, in the EuroSAT-MS fusion column; the SO2Sat fusion column is left unmarked because there is no gain (its best value is zero, which is the finding), and the asymmetry columns are unmarked inputs.}
\label{tab:sensorfusion}
\centering\scriptsize
\setlength{\tabcolsep}{3.2pt}
\zebra{5}
\begin{tabular}{rrrrrr}
\toprule
& \multicolumn{2}{c}{EuroSAT-MS (one instrument)}
& \multicolumn{2}{c}{So2Sat (two satellites)} \\
\cmidrule(lr){2-3}\cmidrule(lr){4-5}
Data & $|\Delta_{\mathrm{acc}}|$~\down & Fusion~\up
     & $|\Delta_{\mathrm{acc}}|$~\down & Fusion~\up \\
\midrule
1\%  &  6.68 & \snd{$+1.40$} & 26.31 & $-0.57$ \\
2\%  &  1.07 & $\mathbf{+2.96}$ & 33.77 & $-0.14$ \\
5\%  &  0.58 & \trd{+1.34} & 30.61 & $-0.73$ \\
10\% &  0.66 & $+1.15$ & 28.51 & $-0.11$ \\
25\% &  0.11 & $+0.45$ & 25.98 & $\pm0.00$ \\
\bottomrule
\end{tabular}
\end{table}

\emph{The prior itself transplants to both sensors, and the radar column needs two caveats stated before its numbers.} \looseness=-1 On optical it produces the study's usual envelope, $+1.13$, $+1.91$, $+1.19$, $+0.68$ and $-0.56$ at 1, 2, 5, 10 and 25\%: positive while data is scarce, decaying to neutral at sufficiency. On SAR it gives $+1.24$, $-3.09$, $+2.59$, $-0.46$ and $-0.24$ at the same five fractions.

The first caveat is that our channel reduction is not defined for radar. The target is built from a luminance weighting, generalized for inputs with other than three channels to a uniform mean over them: defensible for optical bands, but for synthetic-aperture radar it averages backscatter channels and carries no photometric meaning. Any weak or null result on SAR is therefore ambiguous between ``a fixed oriented-energy target does not suit radar statistics'' and ``this channel reduction is wrong for radar'', and we cannot separate the two. A radar-specific reduction is a new design decision, not a defect repair, and we have not run it.

\looseness=-1 The second concerns the status of the SAR expectation, stated against our own interest. Elsewhere the bands and falsifiers were fixed before the cells ran; here they were not: when the SAR expectation was written down, the 10\% baseline cell and one seed of its prior arm were already on disk, so that band is a post-hoc reading of one seed and we do not claim the accompanying $\Delta \le -1.0$ tripwire as a survived test. As measurement, not adjudication: the 2\% cell sits well below that line at $-3.09$, with both arms unstable there (seed spreads of $3.5$ and $5.5$ points against $0.1$--$1.1$ elsewhere in the column). On the multispectral population, all-optical, the picture at $5\%$ is the usual one: the prior adds $+1.63$, $+2.60$, and $+3.30$ on the visible, non-visible, and fused sources.

\emph{Whether the prior amplifies a second sensor is backbone-dependent, and we claim only what both backbones support.} On EuroSAT at $5\%$ the ResNet gain rises with sensor count, while on ViT-tiny at $10\%$ it inverts, $+5.14$, $+2.69$, $+1.75$, tracking baseline weakness ($84.80$, $90.19$, $91.52$) as the rest of this study does, so the defensible statement is only that a second source does not absorb the prior on either backbone. On the arms carrying the prior, EuroSAT's fusion gain is $+1.92$, $+1.46$, $+2.04$, $+1.01$ and $+0.46$ against the baseline arms' $+1.40$, $+2.96$, $+1.34$, $+1.15$ and $+0.45$: prior and second sensor coexist rather than substituting.

\emph{The feature side carries part of it.} Under identical frozen-feature evaluation the prior's $G$ on So2Sat at 5\% is $+1.31$, $+1.54$ and $+0.88$ on radar, optical and fused, against end-to-end gains of $+2.59$, $+1.19$ and $+1.35$; on EuroSAT at 5\% $G$ is $+0.98$, $+1.29$ and $+1.16$ against $+1.63$, $+2.60$ and $+3.30$. The gain is therefore partly feature-level, with the \readout{} term supplying the rest when data is scarce. Every baseline here sits between $43.5$ and $98.3$, far above the crossing, where Sec.~\ref{sec:lawaudit} reports the \readout{} term decayed to nothing; the residuals run from $-0.96$ to $+2.14$, and we do not read signs from them.

\subsection{Two classifiers: decision-level fusion}\label{sec:decisionfusion}
Multi-classifier fusion is a different level again, and we did not derive the rule there. From the CIFAR-100 5\% checkpoints, with no further training, we fuse every pair of arms $(i,j)$ by averaging class posteriors, $p_{ij}(y \mid x) = \tfrac{1}{2}\bigl(p_i(y \mid x) + p_j(y \mid x)\bigr)$, and characterize each pair by three scalars on the test set $\mathcal{T}$. The \emph{gain} over the better single arm,
\begin{equation}
  g_{ij} \;=\; \mathrm{acc}(p_{ij}) \;-\; \max\bigl(\mathrm{acc}(p_i),\,
  \mathrm{acc}(p_j)\bigr),
  \label{eq:ensgaindef}
\end{equation}
the \emph{disagreement} rate, the classical diversity measure
\citep{brown2005diversity},
\begin{equation}
  d_{ij} \;=\; \frac{1}{|\mathcal{T}|} \sum_{x \in \mathcal{T}}
  \mathbf{1}\bigl[\, \hat{y}_i(x) \neq \hat{y}_j(x) \,\bigr],
  \label{eq:disagree}
\end{equation}
where $\hat{y}_i(x) = \arg\max_y p_i(y \mid x)$, and the \emph{accuracy asymmetry} $a_{ij} = \lvert \mathrm{acc}(p_i) - \mathrm{acc}(p_j) \rvert$. Here $g_{ij}$ and $a_{ij}$ are accuracy points and $d_{ij}$ is a percentage of test images, so $\beta_d$ converts one into the other. The self-supervised arm here is a 50-epoch pre-training checkpoint, not the 200- or 800-epoch ones of Sec.~\ref{sec:budget}, and the reason belongs in the open: the analysis needs \emph{matched} arms, and $50$ epochs is the budget at which SimCLR lands within a point of the prior on this cell ($30.81$ against $30.51$) where $200$ and $800$ put it $3.9$ and $9.2$ ahead. Pairing the strongest arms would confound diversity with the budget effect Sec.~\ref{sec:budget} isolates; the cost is that we have \emph{not} measured how a longer-pre-trained arm ensemble performs.

\looseness=-1 The first thing the data say is a caution, not a confirmation. Least squares over the $n = 28$ pairs gives
\begin{equation}
  \hat{g}_{ij} \;=\; \beta_0 \;+\; \beta_d\, d_{ij} \;+\;
  \beta_a\, a_{ij},
  \label{eq:ensgain}
\end{equation}
with $\beta_0 = -2.74 \pm 0.48$, $\beta_d = +0.089 \pm 0.011$ and $\beta_a = -0.341 \pm 0.047$ ($R^2 = 0.74$, residual standard deviation $0.42$ points, every coefficient resolved at $\lvert t \rvert > 5$). Those are ordinary least-squares errors, and the $28$ pairs are not independent: all are pairs of the same eight arms, so each arm enters seven. A leave-one-arm-out jackknife, which clusters on the shared unit, leaves the coefficients unchanged and widens the errors to $0.48$, $0.012$ and $0.060$, the asymmetry term by about a quarter, with every coefficient still resolved at $\lvert t \rvert > 5$: the dependence costs precision without changing the reading. Both terms matter, but the dominant one is not diversity, since a one-point accuracy gap costs $\lvert \beta_a \rvert = 0.34$ points of gain, which requires four points of extra disagreement to recover. Restricting to pairs within one point of each other, with currency as the free variable, leaves the three pairs in Table~\ref{tab:ensemble}.

\begin{table}[pos=htbp]
\caption{Decision-level fusion on CIFAR-100 at 5\%, restricted to pairs whose two arms are within one accuracy point of each other so that the information source, rather than an accuracy gap, is what differs. Gain is over the better single arm; disagreement is the fraction of test images on which the two arms predict different classes. Averaging posteriors requires no further training. Best in bold, second best underlined, third best in italics, in the gain column.}
\label{tab:ensemble}
\centering\scriptsize
\setlength{\tabcolsep}{2.6pt}
\zebra{2}
\begin{tabular}{P{0.42\columnwidth}rr}
\toprule
Pair & Gain~\up & Disagreement \\
\midrule
Prior + SimCLR (50 ep)            & $\mathbf{+1.20}$ & 47.2\% \\
SimCLR (50 ep) + prior, wide bank & \snd{$+1.13$} & 47.2\% \\
Prior + prior, wide bank          & \trd{$-$0.04} & 28.1\% \\
\bottomrule
\end{tabular}
\end{table}

Two models built on the same knowledge source disagree on 28\% of test images and ensembling them buys nothing; pair either with a self-supervised model and disagreement rises to 47\%, and ensembling buys ${\sim}1.2$ points. That is the currency rule with its mechanism visible, and it is what the classical account would predict.

We report the qualification with equal weight, because it bounds the taxonomy rather than extending it. The prior and SimCLR do \emph{not} stack at training time (Sec.~\ref{sec:substitute}), yet they ensemble usefully at decision time ($+1.20$). Redundancy in the sense of ``jointly training on both adds nothing'' therefore does not imply redundancy in the sense of ``their errors are correlated''. The substitution result is a statement about training-time fusion, and we do not claim it transfers to decision fusion. With three matched pairs this is suggestive, not established, and we say so.

\subsection{A procedure for deciding whether to fuse}\label{sec:procedure}
If the taxonomy had practical content beyond description, it would be that the outcome of a combination can be estimated \emph{before} the combined system is built, from measurements of each source alone. We state this as a hypothesized procedure, not a finding, since it is what a practitioner would actually run; we then run it prospectively, it fails, and this subsection reports the failure as a result.

\begin{table}[pos=htbp]
\refstepcounter{algnum}
\label{alg:procedure}
\centering
\begin{minipage}{\columnwidth}
\rule{\linewidth}{0.8pt}\\[1pt]
{\footnotesize\textbf{Algorithm \thealgnum}\hspace{0.6em}A hypothesized rule for predicting the outcome of fusing two sources before the combination is trained, evaluated prospectively below, where it calls one pair of nine. Every quantity is measured on the sources \emph{separately}; the only cost beyond training each source once is two frozen-feature evaluations, which run in minutes. Line~\ref{alg:asym} is the weakest-supported step: it dominates our decision-level fit (Eq.~\ref{eq:ensgain}), but our two sensor populations differ in modality as well as in asymmetry (Sec.~\ref{sec:multisensor}), so it should rank candidates, not reject one. ``Differ in kind'' at line~\ref{alg:stack} means the two sources fix different test images and shape different directions in feature space, which Sec.~\ref{sec:samecurrency} measures directly.\par}
\vspace{1pt}\rule{\linewidth}{0.4pt}\par
\footnotesize
\vspace*{-\topsep}\vspace*{4pt}
\renewcommand{\algorithmicand}{\textnormal{and}}
\renewcommand{\algorithmicor}{\textnormal{or}}
\begin{algorithmic}[1]
\REQUIRE sources $A$, $B$; a shared baseline arm; a labeled split
\ENSURE one of \textsc{stack}, \textsc{substitute}, \textsc{interfere}, \textsc{n/a}
\STATE train the baseline, $A$ alone and $B$ alone; record gains
  $\Delta_A, \Delta_B$ and declared compute multiples $c_A, c_B$
\STATE $G_A, G_B \gets$ linear evaluation of the three frozen
  representations under one identical protocol
\IF{$A$ or $B$ is a mature (pre-trained) initialization}
  \STATE \textbf{return} \textsc{interfere}: at full auxiliary strength the cost is
    proportional to that initialization's advantage over training from
    scratch, which is already measured; at $\lambda_0 \le 0.3$ the cost is
    neutral on the cells we tested
\ENDIF
\IF{$|G_A| \le 1$ \OR $|G_B| \le 1$ (points; the gate the prospective test used)}
  \STATE \textbf{return} \textsc{n/a}: a source supplying nothing cannot
    overlap; the other's gain stands in full
\ENDIF
\IF{$|\Delta_A - \Delta_B|$ is large} \label{alg:asym}
  \STATE deprioritize: matched arms are where fusion paid
\ENDIF
\IF{$G_A \approx G_B$ \AND both arms improve the same cells}
  \STATE \textbf{return} \textsc{substitute}: the combination will land
    at the better single arm; take the cheaper source,
    $\arg\min(c_A, c_B)$
\ELSE
  \STATE \textbf{return} \textsc{stack}: a combined gain larger than either
    alone is possible; architecture and intervention strength moderate how
    much of it is realized \label{alg:stack}
\ENDIF
\end{algorithmic}
\vspace*{-\topsep}\vspace*{4pt}
\rule{\linewidth}{0.8pt}
\end{minipage}
\end{table}

Algorithm~\ref{alg:procedure} states it. Two properties matter beyond its cost, as the caption indicates. It is \emph{stated as a prediction, not a retrospection}, which places it alongside transferability estimation \citep{nguyen2020leep}, which scores a single pre-trained source, whereas this scores a \emph{pair}. And it is \emph{falsifiable}: the \textsc{substitute} branch asserts that equal feature gains imply no benefit from combining, which Secs.~\ref{sec:substitute} and \ref{sec:decisionfusion} both test and which the training-time-versus-decision-time discrepancy above already qualifies.

\emph{Tested prospectively, it fails.} \looseness=-1 The illustration below is retrospective: the combination it predicts was already measured. So we ran the procedure as intended. Thirteen pairs whose combination had never been trained were assembled; every input was estimated on a validation split carved per class from the unused portion of each cell's training pool, so no quantity touched test; the calls were recorded; and only then were the thirty-nine combination runs submitted. Four pairs were declined at the $|G|\le1$ gate, three of which then moved from the strong single arm by more than two standard errors. Of the nine reaching a \textsc{stack} or \textsc{substitute} call, \textbf{one matched} what training showed, scoring the outcome on the validation split the calls were made on, and one again on test; crediting every statistically indistinguishable pair to whichever call it received still gives at most four of nine.

\looseness=-1 The antecedent breaks, not the taxonomy. The procedure reads same-currency off \emph{agreement in the magnitude} of two feature gains, and that proxy confuses same-currency-in-different-amounts with different-in-kind: on CIFAR-100 at $10\%$ the prior's $G$ is $+4.1$ and SimCLR's $+8.3$, a $51\%$ gap the rule sends to \textsc{stack}, on the very pair whose substitution Sec.~\ref{sec:substitute} documents, and the combination lands $1.7$ points \emph{below} the better arm at $7.4$ standard errors. Four of the five resolved misses have that shape. We therefore withdraw the claim that these outcomes can be anticipated before the combination is trained and read Algorithm~\ref{alg:procedure} as a statement of the taxonomy, not a validated predictor. Nothing measured is affected: every outcome here is still one of the three and still separated on frozen features, and the matched-budget result of Sec.~\ref{sec:law} concerns a single intervention, not an untrained pair. The predicate that would replace a magnitude comparison is the open problem the next paragraph names.

\emph{How the rule reads a cell.} \looseness=-1 On the moment prior and a SimCLR initialization on ViT-tiny at CIFAR-100 $10\%$, step~1 records $\Delta = +13.26$ at $1.02\times$ compute against $+13.30$ at $2\times$ and step~2 gives $G = +14.83$ against $+13.46$. Neither arm is a pre-trained initialization, and neither $G$ is near zero, so \textsc{interfere} and \textsc{n/a} do not fire; the arms are level, so nothing is deprioritized; the feature gains agree to $1.4$ points where the effects are $13$--$15$, leaving \textsc{substitute}. That is what training shows: $\Delta = +13.46$ with a feature gain of $+14.14$ sitting \emph{between} the two singles. On the prior and DeiT-strength augmentation, the gains differ in kind; the branch is \textsc{stack}, the combination exceeds both, and $G$ rises from $14.83$ to $22.20$; Sec.~\ref{sec:stack} bounds how far that reaches. Both readings are retrospective, and the prospective test above is what they could not substitute for.

The antecedent is the limit. We compare feature gains by magnitude and by which cells improve, not by a formal decomposition of the information each source carries, and the magnitude half is exactly what the prospective test broke. Partial information decomposition \citep{williams2010nonnegative,liang2023quantifying} is the principled version of that comparison, and an estimator that scaled to these representations would replace our proxy with a quantity that means something on its own.

\emph{Takeaway.} The currency account extends past training-time fusion, but the two experiments carry different weight, and we attribute the claim only to the one that supports it: in the decision-level fit, how asymmetric two sources are predicts whether combining them pays, and the sensor experiments are consistent with that but cannot establish it, their two populations differing in asymmetry and modality at once.

\section{Does any of this hold off classification?}\label{sec:dense}

Every number to this point is top-1 accuracy. That is a real limit on what the benchmark can claim, since the auxiliary target is itself a \emph{dense} map, oriented energy at every location, so a reader is entitled to ask whether the fusion outcomes and the law $\Delta = G + \readout$ are properties of the method or of the metric. This section answers that on semantic segmentation, then on object detection, where the answer changes.

\textbf{State the bias before the result.} A dense spatial target on a dense spatial task is the most favorable venue this prior could be given, so a \emph{positive} result here would be weak evidence of generality; the headline finding is negative, and negatives are not flattered by a favorable venue.

\begin{figure}[pos=htbp]
\centering
\includegraphics[width=1.0\linewidth]{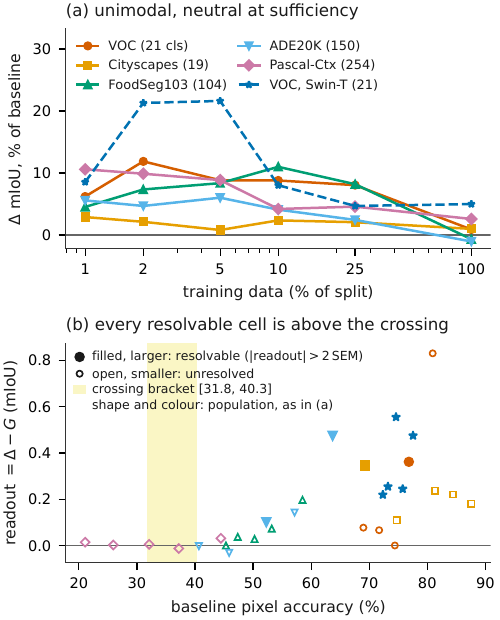}
\caption{Semantic segmentation, six populations, 200 epochs, three seeds per cell. \textbf{(a)} $\Delta$ as a percentage of each cell's own baseline, not absolute mIoU, since the baselines differ eightfold. Read this way, four of the six envelopes are unimodal, Cityscapes is flat within noise, and Pascal-Context is highest at $1\%$ and lower everywhere after; all six occupy the same \denseRelLo--\denseRelHi\% band classification does across $1$--$25\%$, and three go neutral or negative at full data. \textbf{(b)} The \readout{} term against the head's own classification scale (pixel accuracy), not mIoU: the crossing bracket is an accuracy bracket, and scoring the law on mIoU puts every dense cell on the wrong flank. Filled markers are the resolvable cells; all sit above the bracket, so segmentation confirms the law's positive branch and cannot test its negative one.}
\label{fig:dense}
\end{figure}

\paragraph{Setup.} Six populations spanning label-space width and domain: \textbf{PASCAL VOC-2012} (21 classes, augmented split), \textbf{Cityscapes} (19, driving, half resolution), \textbf{FoodSeg103} (104, texture-dominated), \textbf{ADE20K} (150, scene-centric, no background class), \textbf{Pascal-Context} (254), and a \textbf{Swin-Tiny} arm on VOC. The encoder is the study's ResNet-18 at output stride 8 with an FCN head, so the prior's bank, tap, $\lambda$ schedule, and head normalization transfer without redesign or new hyperparameters. Table~\ref{tab:dense} gives the full envelope and Fig.~\ref{fig:dense} summarises it. \densePops{} populations $\times$ six fractions $\times$ two arms $\times$ three seeds $=$ \denseCells{} runs (72 cells in the sense of Sec.~\ref{sec:stats}), all completed.

Pascal-Context is the control that carries most of the weight: it re-annotates the \emph{identical} VOC images at 254 classes instead of 21, so it separates a label-space effect from a pixel effect exactly as CIFAR-100-super does on the classification side.

\paragraph{Two measurement decisions, both of which we got wrong first.}
Absolute mIoU is not comparable across these populations: their baselines differ eightfold ($51.5$ on VOC at full data, $6.4$ on Pascal-Context). Comparing raw $\Delta$ across them makes the label-space effect look like a collapse, more than $10\times$ in the mid band in absolute terms, when in relative terms it is at most $2.3\times$, and runs the other way at $1\%$ and $100\%$. We report both, and read comparisons off the relative column.

Second, and more consequential: \emph{readout must be evaluated on the head's own classification scale, not on mIoU.} The crossing bracket $[31.8, 40.3]$ was measured in accuracy. mIoU is a far harsher statistic: it averages per-class IoU with equal weight, so classes the model never learns dominate it, and a cell at $6.2$ mIoU may be classifying $69\%$ of its pixels correctly. Scoring the law on mIoU places every dense cell deep on the left flank and predicts a negative \readout{} everywhere; scoring it on pixel accuracy places them on the decayed positive branch, which is where they are. We registered this limitation in advance and then violated it in our own prediction, which is how we found it.

\begin{table}[pos=htbp]
\caption{The dense envelope. $\Delta$ in mIoU points, with $\Delta$ as a percentage of that cell's own baseline beneath it. Three seeds per cell, ResNet-18 with an FCN head at output stride 8, 200 epochs at every fraction, the same budget the classification recipe uses. Pascal-Context is VOC's own pixels relabelled to 254 classes. The last block ($\dagger$) is a Swin-Tiny encoder and is a \emph{diagnostic} row, not a headline one: it trains under AdamW, not the frozen recipe, for the reason given in Sec.~\ref{sec:denseattn}, and carries the bistability caveat that attaches to every Swin cell.}
\label{tab:dense}
\centering\scriptsize
\setlength{\tabcolsep}{3.0pt}
\zebra{2}
\begin{tabular}{lrrrrrrr}
\toprule
Population & Cls. & 1\% & 2\% & 5\% & 10\% & 25\% & 100\% \\
\midrule
VOC           & 21  & $+0.39$ & $+0.97$ & $+1.15$ & $+1.61$ & $+2.34$ & $+0.46$ \\
\quad rel.    &     & 6\%     & 12\%    & 9\%     & 9\%     & 8\%     & 1\%     \\
Cityscapes    & 19  & $+0.49$ & $+0.41$ & $+0.20$ & $+0.68$ & $+0.73$ & $+0.48$ \\
FoodSeg103    & 104 & $+0.12$ & $+0.20$ & $+0.26$ & $+0.43$ & $+0.56$ & $-0.12$ \\
ADE20K        & 150 & $+0.18$ & $+0.20$ & $+0.38$ & $+0.38$ & $+0.37$ & $-0.29$ \\
Pascal-Ctx    & 254 & $+0.11$ & $+0.13$ & $+0.16$ & $+0.10$ & $+0.16$ & $+0.17$ \\
\quad rel.    &     & 11\%    & 10\%    & 9\%     & 4\%     & 5\%     & 3\%     \\
\midrule
Swin-Tiny, VOC$^{\dagger}$ & 21 & $+0.42$ & $+1.29$ & $+1.76$ & $+0.90$ & $+0.75$ & $+1.31$ \\
\quad rel.    &     & 9\%     & 21\%    & 22\%    & 8\%     & 5\%     & 5\%     \\
\bottomrule
\end{tabular}
\end{table}

\subsection{Target--task alignment is not the source}\label{sec:densealign} This is the section's main result, and it is a negative one. VOC at $1\%$ is 107 images over 20 classes, about five per class, the same supervision density at which classification shows a universal ${\approx}+1.5$ floor (CIFAR-100 $+1.42$, Tiny-ImageNet $+1.49$). Dense prediction at that density returns \textbf{$+\denseVocOneDelta$ mIoU} (\denseVocOneRel\%), and the frozen-feature probe agrees ($G = +0.31$), so it is not a \readout{} artifact.

So the venue built to flatter the prior, a dense target supervising a dense task, pays \emph{less}, not more. Whatever the prior supplies, it is not "structure that happens to match the output space." That makes the mechanism claim in Sec.~\ref{sec:fusion} more precise, not weaker: the prior injects oriented-energy structure into the representation, and what the head does with it afterward is a separate question.

\subsection{Regimes transfer; magnitudes are metric-bound}
In absolute mIoU, the dense gains look tiny beside the classification ones. In relative terms, they run \denseRelLo--\denseRelHi\% of baseline across the $1$--$25\%$ band (three populations go negative at $100\%$, to $-1.1\%$), which is the range classification occupies (CIFAR-100 at $5\%$ is $+5.15$ on $25.36$, i.e.\ $20\%$). The shape transfers on four of the six populations, and we give the other two instead of averaging them away: reading the relative column throughout, VOC peaks at $2\%$ and FoodSeg103 at $10\%$ (in absolute mIoU both instead peak at $25\%$, VOC at $+\denseVocPeak$) and ADE20K at $5\%$, while Cityscapes is flat within noise, dipping at $5\%$ and rising again, and Pascal-Context falls from $1\%$ with one uptick between $10$ and $25\%$. ADE20K's absolute column would instead put its peak at $10\%$, but the two candidate cells differ by $0.006$ mIoU against standard errors of $0.02$ and $0.15$, so that location is unresolvable and we do not assert it.

At full data, the gain decays to neutrality or below on three of the six (VOC $+\denseVocFull$, i.e.\ \denseVocFullRel\%; FoodSeg $-0.12$; ADE20K $-0.29$), and stays positive on Cityscapes ($+0.48$, though indistinguishable from VOC's $+\denseVocFull$), Pascal-Context ($+0.17$, resolved at ten standard errors) and Swin-VOC ($+1.31$). Structural neutrality at sufficiency therefore holds on the three populations large enough to reach sufficiency, and the exceptions are the ones this study's own image-count axis predicts: Cityscapes trains on $2{,}975$ images and Pascal-Context on $4{,}998$, which are low-data cells here whatever their label says, exactly as DTD and CUB-200 are in the classification grid.

\paragraph{A validity note that changed a conclusion.} We first ran this grid at 50 epochs and reported an envelope that rose monotonically on every population, with no right flank. That was an artifact: at 200 epochs the VOC $10\%$ baseline rises from $7.23$ to $18.28$ mIoU ($+153\%$), and the 50-epoch curves were still climbing at ${\sim}0.7$ mIoU/epoch when the cosine schedule annealed the step size away, so their apparent convergence was the schedule. Re-pinning to 200 epochs, matching the classification budget instead of choosing one to suit the task, moved $\Delta(\mathrm{VOC},100\%)$ from $+2.57$ to $+0.46$, from a nominal falsification of neutrality at sufficiency to a confirmation of it: an under-trained instrument produced the opposite answer twice (Sec.~\ref{sec:dataopt}).

\subsection{The law on a different task and metric}
Of the \denseLawCells{} cells carrying both $\Delta$ and $G$ (full-data cells are excluded: there the probe's labels \emph{are} the cell's labels, so no split is interpretable), \denseLawBracket{} sit inside the crossing bracket where the law makes no sign call and \denseLawUnres{} have $|\readout| \leq 2$\,SEM, which on the right flank is expected, not disappointing: $\Delta$ and $G$ are both near zero there and the sign of their difference is noise. That leaves \denseLawResolvable{} cells that actually test the law, and \denseLawCorrect{} of \denseLawResolvable{} fall on the predicted side.

\looseness=-1 The qualification is that all \denseLawResolvable{} sit \emph{above} the crossing (pixel accuracy \denseLawPixLo--\denseLawPixHi), so segmentation confirms the law's positive branch on a new task and a new metric and does not test its negative branch. Pascal-Context was built to supply exactly that test, and cannot: resolving a \readout{} requires not only a baseline below the bracket but a $\Delta$ large enough that $\Delta - G$ exceeds its own uncertainty, and Pascal-Context's $\Delta$ runs $+0.10$ to $+0.17$ mIoU, so its \readout{} is a difference between two small numbers whatever its sign. It delivers a null by construction: trained pixel accuracy $21.1$ and $25.9$ at $1$--$2\%$, clearly below the bracket where a negative \readout{} is required, and a \readout{} of $+0.02$ and $+0.00$, zero, not negative. The pre-registered falsifier needed a clearly positive \readout{} and did not fire; the prediction was not confirmed either; we record the cells as \emph{unresolved} and claim nothing from them. A population that could settle it needs abundant labels, low pixel accuracy \emph{and} a material $\Delta$ at once, and none of our six has all three.

\subsection{Attention, narrowed}\label{sec:denseattn}
On classification, the prior is worth roughly twice as much to a hierarchical-attention backbone as to a convolutional one ($G_{\mathrm{swin}} = 7.6$--$10.5$ against $G_{\mathrm{r18}} = 3.55$--$6.26$ on CIFAR-100). On dense prediction, that advantage survives only in a narrow band. Relative feature gain, Swin against ResNet-18 on identical images: $4.6$, $20.7$, $17.5$, $7.0$, $2.0\%$ at $1$--$25\%$, against $6.3$, $13.4$, $10.8$, $8.1$, $5.8\%$. Swin leads at $2\%$ and $5\%$ by ${\sim}1.6\times$ and trails at $1\%$, $10\%$ and $25\%$; by $25\%$ its $G$ has collapsed to $+0.27$ against convolution's $+1.51$. At full data it still gains ($+1.31$, $5\%$ relative), while the convolutional arm is near-neutral ($+0.46$, $1\%$), which mirrors the classification result.

We state this narrowly on purpose. An earlier, under-trained version of this experiment supported the opposite conclusion, that the attention deficit does not transfer to dense prediction, and a four-fraction reading of the corrected data supported a broader one. Both were over-readings of the fractions then available. Swin is not ViT, this is one dataset, and these are AdamW diagnostic cells; what they establish is that the attention advantage is task- and regime-dependent, not absent.

\subsection{The label-space effect largely dissolves}
VOC and Pascal-Context are byte-identical pixels at 21 and 254 classes. In absolute mIoU, the gap is large at every fraction, and at full data VOC's $+\denseVocFull$ against Pascal-Context's $+\densePcFull$ invites a conclusion about label granularity. Normalized, it mostly disappears: \denseVocFullRel\% against \densePcFullRel\% at full data, $9\%$ against $9\%$ at $5\%$, and Pascal-Context is \emph{higher} at $1\%$ ($11\%$ against $6\%$). Relative feature gain behaves the same way. The defensible statement is a mid-band tendency for the coarser label space, not the collapse the absolute numbers suggest, and it is visible as such only because the control holds the pixels fixed.

\subsection{Detection: the prior does not transfer}\label{sec:detection} Segmentation still asks a network to \emph{label}, one decision per pixel. Detection is the only task here whose head \emph{regresses coordinates}, and oriented energy is about edges and extents, so if the prior were going to help anything localize, this is where it would show. We run it on the same VOC images and the same committed subset indices the segmentation cells use, with the dense recipe verbatim (200 epochs, identical crops and augmentation) and a single-level anchor-free FCOS head at stride 8 on the same tapped backbone. Detection therefore differs from segmentation only in the head and the loss. \detCells{} runs (\detCellCount{} cells in the sense of Sec.~\ref{sec:stats}): six fractions, both arms, three seeds.

\paragraph{The result, and one number that was not real.} Every measure is flat (Table~\ref{tab:det}). The single exception is $+\detTenDelta$ AP50 at 10\%, and we do not report it as a gain: its own components are all consistent with zero ($\Delta$fg\_acc $-0.32$, $\Delta$fg\_iou $+0.0032$), and at AP25 it falls to $+0.35$, inside noise. A gain that is neither classification nor localization and does not survive a threshold change is ranking noise.

\begin{table}[pos=htbp]
\caption{Detection is a null. $\Delta$ is aux minus baseline on PASCAL VOC. AP50 is the task metric; fg\_acc and fg\_iou are the head's 20-way accuracy and its mean box IoU \emph{at ground-truth foreground locations}, so neither can collapse the way a ranked-precision integral does. $G$ is the same quantity under a frozen trunk with a fresh $1\times1$ head fitted on the full split. Cells marked $\dagger$ (\detFloorPcts) fall under the pre-declared floor rule: both arms below 1.0 AP50, where a difference is not a measurement.}
\label{tab:det}
\centering\scriptsize
\setlength{\tabcolsep}{3.6pt}
\zebra{2}
\begin{tabular}{lrrrrr}
\toprule
Data & $\Delta$AP50 & $\Delta$fg\_acc & $\Delta$fg\_iou & $G$(fg\_acc) & $G$(fg\_iou) \\
\midrule
1\%$^\dagger$  & $\detOneDelta$ & $-0.23$ & $-0.0003$ & $-0.16$ & $+0.0017$ \\
2\%$^\dagger$  & $-0.00$ & $-0.23$ & $-0.0001$ & $-0.37$ & $-0.0006$ \\
5\%            & $-0.08$ & $+0.11$ & $-0.0048$ & $+0.34$ & $+0.0130$ \\
10\%           & $+0.56$ & $-0.32$ & $+0.0032$ & $+0.08$ & $+0.0073$ \\
25\%           & $-0.41$ & $+0.01$ & $+0.0017$ & $-0.47$ & $+0.0011$ \\
100\%          & $\detFullDelta$ & $+0.64$ & $+0.0031$ & --      & --        \\
\bottomrule
\end{tabular}
\end{table}

We also had to withdraw a larger number. At 5\% the raw arithmetic gives $+0.81$ AP50, and it is entirely one collapsed baseline seed: its fg\_iou is \emph{exactly} $0.0000$ while its fg\_acc matches its siblings, and its regression loss pins at exactly $1.0000$ from epoch 29 to 199. The predicted boxes went to zero area, the GIoU term saturated, its gradient vanished, and that branch could never recover while the classification branch kept improving. This is a \emph{partial} collapse of one branch of a two-headed model, and the seed-level check used elsewhere in this study would not have caught it, because the cell still trains. Excluding it, $\Delta$ at 5\% is $-0.08$.

\paragraph{The null is the prior, not the head.} A flat end-to-end result on a task with a weak head admits a second reading, and it is the one this study documents everywhere else: at five images per class the frozen-feature probe on CIFAR-100 sees $+4.16$ of feature gain where the trained classifier realizes $+1.42$ (the reference-configuration $1\%$ pair of Table~\ref{tab:envelope}; Appendix~\ref{app:law}). So we probed the detection trunks under the same protocol, and the answer is unambiguous: $G(\mathrm{fg\_acc}) = \detOneGFgAcc$ at 1\% and $-0.37$ at 2\%, against a pre-registered falsifier at $+1.5$. A fresh head with a hundred times the cell's labels extracts nothing more from the prior's trunk than from the baseline's.

What licenses that reading is that the probe demonstrably works here: at 1\% it lifts fg\_acc from the trained cell's \detOneFgAcc{} to \detProbeLiftOne. The trained detection head really is label-limited at 107 images, which is exactly the condition under which the classification left flank hides feature gain. The condition is met, and there is nothing to find. One quantity does move, $G(\mathrm{fg\_iou}) = +\detTenGFgIou \pm \detTenGFgIouSem$ at 10\%, resolvable on the localization side, not the classification side, but $0.007$ of IoU is not a result we will build on.

\paragraph{Two things this cannot be blamed on.} The instrument is adequate: at full data the detector reaches \detFullAPnone{} AP50 from scratch, so the null is not a broken head, and the recorded multi-level-FPN conditional does not reopen. And the recipe is the segmentation recipe unchanged, on the same images, so this is not the undertrained-instrument error of the 50-epoch dense grid.

\paragraph{What detection cannot do, stated as a limitation.} We had hoped it would supply the negative branch of the sign law, which segmentation could not: baseline fg\_acc is $\detOneFgAcc$--$28.9$ at 1--5\%, below the crossing bracket, and $\Delta \approx 0$ there, so $\readout = -G$. It does not, and the obstruction is structural. The law needs $G$ on the \emph{same} metric as $\Delta$, and the probe's own AP50 floors at $1$--$2\%$ ($\detProbeOneAP$ at 1\%) and sits at the floor's edge at $5\%$ ($0.97$ and $1.15$ on the two arms). Above the crossing $\Delta$ and $G$ are both near zero, so their difference is unresolvable. Detection contributes no resolvable law cell, the same outcome as Pascal-Context and for the same reason. The negative branch remains untested by either non-classification task, and we prefer to say so rather than construct a fifth instrument to chase it.

\paragraph{The task axis.} Taken with Sec.~\ref{sec:densealign}, the three tasks order cleanly and the ordering is the sharpest limit we can put on the method: what the prior supplies is realized \emph{fully} by a whole-image classifier, \emph{partly} by a per-pixel classifier, and \emph{not at all} by a coordinate regressor. In the last case, we now know why: the features are not there, not the head being unable to reach them.

\section{Ablations and controls}\label{sec:ablations}

\looseness=-1 The benchmark's defensibility rests on knowing \emph{which} ingredient carries the gain, so we ablate every one, moving outward from the target itself to the design choices around it and ending with a second effect the accuracy tables alone would conflate with the first: the prior's stabilization of backbones whose baselines collapse. Section~\ref{abl:notablated} states what we did \emph{not} vary.

\subsection{The target is the ingredient}\label{abl:target}

\looseness=-1 Table~\ref{tab:controls}, on CIFAR-100: a random fixed target of identical shape moves accuracy by at most $0.12$ points in either direction, so on that dataset the mechanism is not ``regression as regularization''. A FitNets teacher trained on the same data at twice the cost does nothing anywhere we measured it (eleven fractions from $1$ to $100\%$, $-0.49$ to $+0.73$), making the free hand-crafted target strictly better than an expensive learned one in this regime. Among hand-crafted families, phase-invariant \emph{magnitude} wins at both fractions by a factor of $1.6$ to $2$ over the next-best target, and discarding its magnitude scale (cosine loss) forfeits most of the gain. The ordering \emph{below} magnitude is not stable, and we would rather say so than rank six targets off one column: the structure tensor is fourth at $5\%$ and second at $10\%$, and only the two ends hold at both fractions, magnitude first and heavily processed invariants last and actively harmful.

\begin{table}[pos=htbp]
\caption{Target and mechanism controls on CIFAR-100 with ResNet-18 (each row is the reference configuration with one ingredient varied; $\Delta$ vs.\ the shared baseline, three seeds in the varied arm against the released ten-seed baseline). These cells hold $\lambda{=}0.3$, not the reference configuration's decaying schedule, so they are comparable to each other, as an ablation requires, but not to Table~\ref{tab:envelope}. On this dataset, the gain needs the moment structure specifically; the text quantifies each control. CIFAR-100 is the set on which the configuration was selected, and the text below reports the same ablation transplanted to three populations that are not.}
\label{tab:controls}
\centering\scriptsize
\setlength{\tabcolsep}{2.6pt}
\zebra{4}
\begin{tabular}{P{0.33\columnwidth}rrP{0.40\columnwidth}}
\toprule
 & \multicolumn{2}{c}{$\Delta$ at fraction~\up} & \\
\cmidrule(lr){2-3}
Aux target & 5\% & 10\% & Note \\
\midrule
Moment magnitude (ours) & $+3.18$ & $+2.71$ & phase-invariant energy \\
Structure tensor & $+1.07$ & $+1.68$ & mild nonlinearity \\
Steerable harmonics & $+1.55$ & $+0.91$ & principled rot.-inv. \\
Rotation-invariant energy & $+1.43$ & $+0.74$ & \\
Oriented edges (Gabor) & $+0.56$ & $-0.09$ & raw edges do not help \\
2nd-order invariants & $-0.43$ & $-3.07$ & over-processed \\
\midrule
HOG target & $+1.08$ & $+1.27$ & descriptor helps; moments ${\sim}2.5\times$ better \\
Learned teacher (FitNets) & $-0.49$ & $+0.06$ & $2\times$ cost; $-0.49$ to $+0.73$ over eleven fractions \\
Random fixed maps & $-0.12$ & $+0.04$ & rules out ``any aux signal'' \\
\midrule
Cosine loss (not MSE) & $+0.62$ & $+0.36$ & magnitude scale matters \\
\bottomrule
\end{tabular}
\end{table}

\paragraph{The same ablation off the selection set, where three of four registered falsifiers fired.} CIFAR-100 is the set the configuration was chosen on (Sec.~\ref{sec:selection}), so we repeated the comparison where nothing was: all eight targets on Tiny-ImageNet at $5$ and $10\%$, and the decisive random control on Food-101 ($64$\,px) and CIFAR-10 ($32$\,px).

\looseness=-1 The random control does not translate as a null, which was our sharpest tripwire. On Tiny-ImageNet it reaches $+0.67 \pm 0.13$ at $5\%$ and $+0.37$ at $10\%$, five standard errors from zero and $38\%$ of magnitude's $+1.78$; on Food-101 it gives $+0.06$ and $+0.38$; on CIFAR-10 it is \emph{negative}, $-0.70$ and $-1.46$. Every $32$\,px cell sits at or below zero and every $64$\,px cell at or above it, so ``a random fixed target does nothing'' is at best a $32$\,px statement, and at $32$\,px an uninformative fixed target is an active cost, as the forward-path band (Sec.~\ref{abl:forward}) and the transfer tax (Sec.~\ref{sec:taxsub}) also report for shaping toward structure the data does not want. What survives on all four populations and both fractions is the \emph{margin}: the moment target beats the random one by $+0.82$ to $+4.19$, and that replaces the null as our claim. One of the eight moment cells has two seeds after a run failure, but both endpoints of that range come from three-seed cells.

Two further tripwires fired on the letter. Magnitude leads on Tiny-ImageNet at $5\%$ but is third at $10\%$ ($+1.19$ against structure's $+1.33$ and HOG's $+1.28$), both gaps inside their own uncertainty ($\pm 0.38$, $\pm 0.34$), so the honest reading is first at $5\%$, tied for first at $10\%$, never worse than third, and HOG is not ${\sim}2.5\times$ behind off the selection set but $1.8\times$ at $5\%$ and tied at $10\%$. Heavily processed invariants, last and actively harmful on CIFAR-100 ($-3.07$), are \emph{positive} on Tiny-ImageNet at both fractions ($+1.28$, $+0.67$).

The real finding is about the instrument. On Tiny-ImageNet, all eight targets help ($+0.23$ to $+1.78$) and best-to-worst at $10\%$ spans $1.1$ points against seed uncertainties of $0.2$ to $0.4$, so most pairwise differences there are not resolvable: the ablation's discriminating power is itself a CIFAR-100 property, where the whole envelope is two to three times larger. Magnitude stays the best single choice we measure, and the margin over an uninformative target holds everywhere, but \emph{which} hand-crafted family is used matters less off the selection set than Table~\ref{tab:controls} alone suggests.

\subsection{Placement, schedule, and head}\label{abl:placement}
Tap depth is flat across the first three stages (within seed noise at both 1\% and 10\%), with a cliff only at the final stage: the one design knob in the study that is \emph{not} a data-regime knob. The schedule start $\lambda_0$ \emph{is} a regime knob (best $2.0$ at 1--2\%, $1.0$ at 3--10\%, $0.3$ at 15--25\%), but its reach is bounded: three attempts to push $\Delta$ past the measured feature gain by retuning $\lambda_0$ all failed. Consistent with the law, the knob cannot manufacture $G$; it can only realize it. Head-norm projection, derived from a traced scale degeneracy (the aux loss is invariant under feature-shrink with head-inflate, which collapses bottleneck ResNets), converts ResNet-50 from a bistable failure ($\{41.2, 36.4, 42.4\}$ across seeds) into a stable $+3.9$ ($\{44.5, 45.0, 44.4\}$) and is free elsewhere; we adopt it always-on. Three plausible fixes tried \emph{before} tracing the mechanism (gradient balancing, disabling automatic mixed precision, BatchNorm on the tap) all failed or backfired, the last making collapse $25\times$ faster: a small case study in mechanism-first debugging.

\subsection{One configuration for every architecture}\label{abl:universal}
A single reference setting transplants without retuning: at CIFAR-100 with $10\%$ of the data, ResNet-18, -34 and -50 land $+4.3$, $+4.0$ and $+3.9$ against their own baselines (these are the head-norm cells of the backbone sweep, not the frozen-recipe envelope of Table~\ref{tab:envelope}); the new depth sweep confirms one $\lambda_0$ across ResNet-34 and -50 on six further domains ($+0.4$ to $+9.0$ at 5--10\%), with two exceptions we name instead of absorbing: CUB's deep-scarcity floor, and DTD at $5\%$ at $-0.25$. The right flank decays on eight of the twelve cells; the four that rise at $25\%$ are DTD and CUB on both depths, and their $25\%$ cells hold a few hundred images, which on this study's own image-count axis makes them low-data cells, not right-flank ones. Head \emph{form} is irrelevant where it plausibly might not have been: linear, cosine, and nearest-class-mean readouts measure the same aux-vs-baseline gap to within $0.3$ points. The left-flank bottleneck is label information, not classifier expressivity.

\subsection{The forward-path alternative}\label{abl:forward}
Placing the same bank \emph{in the forward path} as concatenated fixed channels at the stem, the classical hybrid design, yields the study's largest structural contrast. Forward-path stems are the low-data accuracy leaders (energy-magnitude reaches $+2.5$ and $+3.5$ at CIFAR-100 1--2\%, beating the auxiliary prior there), but every fixed forward variant we test (linear Gabor at two kernel scales, and five nonlinear energy families) pays a penalty band at 10--25\% data that no filter choice escapes. Its depth is what the choice governs: the mild case is the $k5$ Gabor stem, which reaches only $-0.19$ at $10\%$, and the severe cases are the nonlinear energy families, which run to $-7.4$ to $-11.9$ at $25\%$. Even at $100\%$ they do not wash out, ending between $-1.0$ and $-2.1$: a hard constraint permanently occupies input bandwidth that abundant data would rather use. The auxiliary placement of the \emph{same} knowledge is what removes the band, which is the design argument for fusing priors through training-only targets instead of through architecture.

\subsection{Bank width}\label{abl:width}
Widening the pinned 8-pair bank to 12 channels (via a third octave or two extra orientations, indistinguishable at $10{\times}10$ seeds) buys a real but local $+0.3$--$0.5$ at Tiny-ImageNet's 64\,px and nothing at 32\,px; across five further 64\,px domains the pooled excess fails our pre-registered re-pin criterion. The pinned 8-pair bank therefore stays, with cross-domain evidence, not convenience, behind it.

\subsection{Stabilization as a second, distinct effect}\label{abl:stab}
Across ConvNeXt (under the frozen SGD recipe), ResNet-50 (without head-norm), Swin-T (on five datasets), and Swin at ImageNet64, baselines are bistable: some seeds train, some sit at chance, while the prior arm trains tightly every time. Swin's feature-level variance shows the instability lives in the \emph{features} (baseline evaluation $\sigma$ up to $4.3$ vs.\ $0.2$ with the prior). We report stabilization as a real, separate property; we do not claim routine variance reduction (a $10{\times}10$-seed test of that claim came back negative, and we report that too).

\subsection{What we did not ablate}\label{abl:notablated}
Completeness cuts both ways, so we list the design constants this paper does \emph{not} vary. Four are applied identically to both arms of every pair, so they bound absolute values without being able to explain a paired difference. Two concern training: the recipe, which is the instrument and is never tuned (Sec.~\ref{sec:recipe}), and average pooling of the target to the tapped stage's resolution, with max and learned pooling untested. Two concern what reads it: the auxiliary head, a single $1{\times}1$ convolution we never widen or deepen, although we ablate the \emph{classifier} head thoroughly (Sec.~\ref{abl:placement}), and the linear evaluation's ridge constant of $10^{-4}$, which shifts every arm together, so $G$'s sign and ordering are unaffected while its numeric value is conditional on it.

\looseness=-1 Three matter more, because they define the knowledge source, not the measurement. The bank's envelope width is tied to its frequency by a fixed relation, so every filter carries the same number of cycles; we vary how many scales and orientations the bank holds (Sec.~\ref{abl:width}) but never that constant. The target is computed on BT.601 luminance, which makes the prior colorblind by construction, and a color-opponent target is untested. And while the schedule's endpoint is ablated and load-bearing ($\lambda(T)=0$ against $\lambda(T)=0.1$), its cosine \emph{shape} is not. None of these is a confound here, since a paired $\Delta$ and a same-protocol $G$ gap are differences taken under identical settings, but they bound what one bank, one head form, and one schedule shape can establish.

\vspace{-0.5em}
\section{A decision guide}\label{sec:guide}
\vspace{-0.5em}
Table~\ref{tab:guide} presents the practical distillation of the benchmark. Its most effective row is the modern-ViT, where the prior integrates with the standard recipe rather than competing with it.

\begin{table}[pos=htbp]
\caption{What to use when: the grid distilled to recommendations. Costs are training-compute multiples of the baseline; gains are typical measured ranges, not promises. ``Prior'' is the reference MomentAux configuration verbatim.}
\label{tab:guide}
\centering\scriptsize
\setlength{\tabcolsep}{2.6pt}
\zebra{2}
\begin{tabularx}{\columnwidth}{@{}P{0.255\columnwidth}P{0.205\columnwidth}lX@{}}
\toprule
Regime & Recommendation & Cost & Measured basis \\
\midrule
Conv backbone, photo-like data, mid band (2--15\%), $2\times$ affordable & SimCLR init & $2\times$ & $+2..{+}9$ over baseline; beats prior by $+2..{+}5$ \\
Conv, photo-like, $2\times$ \emph{not} affordable & prior & $1.02\times$ & $+1.4..{+}6.7$; ${\sim}$SSL parity at extremes at $2\times$; at $5\times$ SimCLR leads \\
Conv, extreme scarcity (${\leq}1$--$2\%$) & prior (or fwd stem) & $1.02\times$ & SSL margin ${\leq}1$ pt at $2\times$; at $5\times$ SimCLR leads by $+3.1$; forward stem best ${\leq}2\%$ \\
Satellite-like statistics, low data & prior & $1.02\times$ & beats SimCLR at 7 of 11 EuroSAT fractions \\
ViT backbones evaluated here under DeiT-strength augmentation & prior \emph{and} augmentation & $1.02\times$ & flip: beats SSL at every fraction at $2\times$; at $5\times$ only in the higher-data band (both populations); $+14..{+}25$ over recipe alone \\
From-scratch attention backbones evaluated in this study & prior & $1.02\times$ & $+3.2$ at 1.28M imgs; $+13$ and $+26$ at 224\,px at 100 ep, $+4.5$ and $+6.7$ at 200; stabilizes Swin \\
ImageNet-transferable task & transfer; the prior only at reduced $\lambda_0$ &--& interference $-15$ to $-17$ at $\lambda_0{=}1.0$, feature-side; neutral at $\lambda_0 {\leq} 0.3$ \\
Domain-shifted from ImageNet (e.g.\ stains) & prior from scratch competitive & $1.02\times$ & transfer advantage shrinks to $+2..{+}3$ \\
Conv, photo-like, ${\geq}50\%$ & nothing (all converge) &--& conv arms within $\pm 0.8$; \emph{not} ViTs, which still gain at full data \\
\bottomrule
\end{tabularx}
\end{table}

\section{Discussion}\label{sec:discussion}

\paragraph{What the decomposition buys, and what it does not.} \looseness=-1
The grid's size is not its point; its point is that one low-dimensional structure organizes much of it. We are careful with ``explains'': as Sec.~\ref{sec:lawform} quantifies, this accounts for a trend and a \emph{sign}, not a predictor of a cell's value. One mechanism nonetheless covers observations that otherwise need separate stories: why gains collapse at extreme scarcity even when features improve (five labels per class cannot express a better representation); why relabeling the same pixels with coarser classes can as much as double the visible gain (the baseline crosses the \readout{} zero, and we measure the features unchanged); and why every intervention converges at sufficiency (both terms go to zero).

\paragraph{What the currency account contributes to fusion theory.}
\looseness=-1 None of our three outcomes is new, and Sec.~\ref{sec:fusiontheory} sets out the prior claims. What that taxonomy, written for sensors whose relationship follows from the sensing geometry, does not provide is a way to \emph{determine} which case holds when one source is a hand-crafted prior and the other a learned representation, neither evident a priori nor predictable from family names (Sec.~\ref{sec:fusion}). Our contribution is operational, not taxonomic: fusion adds value only when the sources' contributions are non-overlapping \emph{in feature space}, with architecture and intervention strength moderating how much is realized; each source's contribution is measurable by an inexpensive linear evaluation of that source alone, even though, as Sec.~\ref{sec:procedure} reports, comparing those measurements does not yet predict which outcome follows. On the destructive outcome, the grid also locates the damage in the feature term, not the \readout{}, making it a fusion outcome on the same currency logic as the other two, not a separate pathology.

\paragraph{What the attention result does and does not say.} The ViT deficit, persistent at 1.28M images and growing with model scale at full data, is a statement about \emph{training small-data vision transformers from scratch}, where a fixed spectral target supplies structure that neither 1.28M images nor the DeiT recipe nor $2\times$-compute SSL supplies at comparable cost. It is not a claim about web-scale pre-training, where data eventually buys everything, and our own convolutional cells show that limit. The reading is deployment-oriented: wherever from-scratch ViTs are trained on $10^4$--$10^6$ images, a $2\%$-overhead structural prior is close to free accuracy, and at a shared budget the smaller-model-plus-prior configuration beat the larger model outright.

\paragraph{On being wrong in public.} Four of our major pre-registered predictions missed: SSL did not collapse at 1\% on convolutional backbones (it won everywhere under plain augmentation); augmentation did not substitute for the prior (it amplified it); stronger contrastive views hurt SimCLR rather than helping; and an early ``gain tracks deficit'' law was falsified outright and retracted. A later round of controls narrowed three claims already written down: the target ablation's discriminating power is a property of the dataset it was run on (Sec.~\ref{abl:target}), the transfer tax of the auxiliary strength rather than the fusion (Sec.~\ref{sec:taxsub}), and the \readout{} term's negative branch substantially of the evaluation's label budget (Sec.~\ref{sec:lawaudit}). Each miss redirected the program, and a benchmark that cannot be wrong measures nothing.

\section{Limitations and future directions}\label{sec:limits}

\paragraph{The law's status and domain of validity.} The law is an \emph{empirical} regularity, not a theorem, and it was fitted to nothing: we report the measured \readout{} curve and use it directly. Its demonstrated domain is 13 datasets across six visual domains, backbones from 2.5M to 86M parameters, images from $32$ to $224$\,px and training sets from 150 to $1.28$M images, all under one supervised recipe with an auxiliary prior trained from scratch; cells initialized from self-supervised or ImageNet weights lie outside that scope and are scored separately. Its credibility rests on prediction, not fit, in the three registered episodes of Sec.~\ref{sec:lawpredict}. Nothing here licenses extrapolation to web-scale pre-training or to architectures whose baselines do not train under this recipe.

\paragraph{Known boundaries.} The account is informative where the \readout{} term
is large and close to uninformative where it has decayed: below the crossing
the sign is right $95\%$ of the time, above it $67\%$, and \auditUnresolved{}
of \auditScope{} in-scope cells are unresolvable, with \auditBracket{} more
inside the bracket where no sign is predicted. Read it as an account of
the scarce-data flank. The one non-classification test we run, semantic
segmentation (Sec.~\ref{sec:dense}), confirms the positive branch on a
different task and metric and cannot test the negative one, for the reason
given there. The second boundary has its own subsection below: the one place where the account is not merely weak but wrong in a patterned way.

\paragraph{Two frozen constants we can bound but not remove.} \looseness=-1 The first is the evaluation's label budget: $G$ is fit on the full training split while the cell is trained on a fraction of it. Section~\ref{sec:lawaudit} measures the consequence at each cell's own budget: most of the negative branch does not survive, and $\Delta$ ties to $G$ within $0.17$ points on average. What remains is the constant itself: one evaluation budget is held for every cell because that is what makes cells comparable, so the \readout{} term is a gap relative to a declared budget, not a pure property of training. The second constant is the recipe, which fixes epochs, not steps; Sec.~\ref{sec:dataopt} states what that costs and where it surfaced, and holding the step count fixed leaves the rising left flank intact and removes the right one.

\subsection{Better features, worse accuracy}\label{sec:exceptions} 
\begin{figure}[pos=htbp]
\centering
\includegraphics[width=\linewidth]{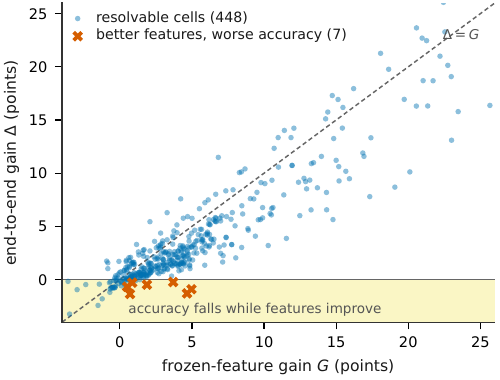}
\caption{Every resolvable law-scope cell, end-to-end gain against frozen-feature gain. On the dashed diagonal, accuracy realizes the feature gain exactly. The shaded quadrant is the patterned failure: features improve measurably while accuracy falls. Only cells whose baseline sits \emph{above} the crossing are marked, since below it a negative \readout{} is what the account predicts, not a violation of it. Produced by the same audit as Table~\ref{tab:audit}, so the figure and the exception list cannot disagree.}
\label{fig:exceptions}
\end{figure}

The decomposition says accuracy tracks features through a \readout{} channel whose sign is set by baseline height. Above the crossing, that channel is predicted to be neutral, so a cell there should realize its feature gain almost exactly. Seven resolvable cells do the opposite: the frozen features measurably improve while end-to-end accuracy falls (Fig.~\ref{fig:exceptions}).

They do not scatter. Five of the seven are Food-101 and PathMNIST at ${\geq}20\%$ of their data and carry the largest magnitudes (Table~\ref{tab:exceptions}); the remaining two, a MobileNetV3 EuroSAT cell and a ResNet-50 CIFAR-100 cell, sit near the threshold at $G \approx +0.5$--$0.9$ and are small enough that we would not build on them. The phenomenon is concentrated on two populations at sufficiency, not confined to them.

It is \emph{not} the deep-left-flank behaviour that superficially resembles it: nine further cells, seven on CUB-200 and two on CIFAR-100, also show positive $G$ with $\Delta < 0$, but their baselines sit far \emph{below} the crossing, where the account predicts a strongly negative \readout{}, so those cells are the account working, not failing, and Fig.~\ref{fig:exceptions} excludes them for that reason. It is also not a measurement artifact on PathMNIST alone, whose linear evaluation we already flag as a compressed measuring stick; Food-101's cells carry five times the magnitude and no such caveat.

What remains is a regime the decomposition has no term for: at sufficiency, on fine-grained and texture-dominated data, the prior can buy representation quality that the network then fails to convert. The overshoot account we use elsewhere, early $\lambda$ shaping costing accuracy where data is already adequate, predicts the sign of $\Delta$ but not the simultaneous \emph{rise} in $G$, which makes these cells informative, not merely inconvenient. Instrumenting the late-training dynamics of these cells is the clearest open problem the grid poses; they stay in every count in Sec.~\ref{sec:lawaudit}.

\paragraph{Scope of the instrument.} \looseness=-1 The frozen recipe is, in effect, a ResNet recipe: architectures that cannot train under it (ConvNeXt, Swin under SGD) enter only through diagnostic AdamW cells, and their bistability, itself a finding, limits headline claims. PathMNIST's shifted test split confounds its high-data cells for every method. Linear evaluations lose interpretability at 100\% cells (the evaluation ceiling), and ImageNet64 evaluations are budget-bound by construction. Both ImageNet stages were subsequently run across $1$--$25\%$ as well as at full data, so the envelope and not only the right flank is measured at scale; a $1\%$ ImageNet64 cell is $12{,}820$ images over $1000$ classes, a label space five times finer than anything else here, and that corner cost us a registered prediction of an interior peak (Sec.~\ref{sec:scaleenvelope}). One prior family (Gabor and moment energy) is tested as the knowledge source; the taxonomy predicts, but does not yet demonstrate, that other structural priors will slot into the same currency logic.

\paragraph{Future directions.} (i) \emph{The exception cluster}: locate where good features fail to become accuracy in the high-data Food-101 and PathMNIST cells. (ii) \emph{Other currencies}: repeat the fusion matrix with priors carrying genuinely different structure (depth, symmetry, frequency-phase) to test whether currency overlap measured by linear evaluation predicts stacking in general. (iii) \emph{Attention at deployment scale}: the model-scale trend ($+13$ to $+26$ from ViT-S to ViT-B at $100$ epochs and $+4.5$ to $+6.7$ at $200$) invites the next point on the curve, and the stabilization-plus-structure account suggests trying the prior as a warmup for large-ViT pre-training proper. (iv) \emph{Protocol}: a benchmark built on this harness should split a validation fold from the start, removing the selection defect Sec.~\ref{sec:selection} discloses, and should estimate the fusion procedure's inputs on that fold so its predictions can be scored prospectively.

\section{Conclusion}\label{sec:conclusion}

\looseness=-1 We opened with one intervention worth $+26$ points, nothing, and $-15$ to $-17$ points in three settings, at one auxiliary strength, one budget, and one schedule, and asked what distinguishes them. Sources trade in currencies: what one is worth is set by which currency is scarce, not by how modern it is, and each source's currency is measurable on frozen features from that source \emph{alone}, before any combination is trained. Across the measured combinations, same-currency pairs substituted, the second source redundant however sophisticated; different-currency pairs could compound, where the architecture and the applied strength permit; and a shaping prior poured at full strength into an initialization already carrying mature features interfered, destroying them in proportion to what they were worth, where a weaker weight does not. Comparing those single-source measurements did not, however, predict unseen combinations when we ran exactly that test, and the taxonomy is stated as retrospective for that reason (Sec.~\ref{sec:procedure}). One decomposition separating feature gain from \readout{} organizes the grid behind this: at a matched label budget the feature gain predicts the end-to-end gain to $0.17$ points, the full-label sign law holds on $\auditRate\%$ of testable cells, and the same discipline cost us two accuracy claims in the budget comparison alone.

The most surprising survivor is the attention result. A bank of oriented filters that predates deep learning is worth $+13$ points to ViT-S/16 and $+26$ to ViT-B/16 at $224$\,px on a shared $100$-epoch budget, $+4.5$ and $+6.7$ when that budget is doubled, \emph{more} to the larger model either way, and still $+3.2$ after 1.28 million images, where our convolutional baseline is neutral
to within $0.04$. Doubling the budget is the one thing that closes part of the gap, and it costs more than the prior does: on that cell, $2\%$ of a training run bought $+26.0$, whereas doubling the run bought $+31.98$. Neither more parameters nor more data closed the gap under any recipe or training budget we evaluated (100 and 200 epochs).

\section*{Acknowledgments}
This work was partially funded by the EU project FoodWISE, 2021-SGR-01094 (AGAUR), Icrea Academia'2022 (Generalitat de Catalunya), and Grants PID2025-173459NB-C21 (BRIDGE-AI) and AIA2025-163919-C51 (EXPLORA), funded by MICIU/AEI, by FEDER (UE), and by European Union NextGenerationEU/PRTR. A.~AlMughrabi acknowledges the support of FPI Becas, MICINN, Spain. The authors thankfully acknowledge the RES resources provided by the Barcelona Supercomputing Center on MareNostrum5 for IM-2025-3-0008.

\section*{CRediT authorship contribution statement}
\textbf{Ahmad AlMughrabi}: Conceptualization, Methodology, Software,
Validation, Formal analysis, Investigation, Data curation, Writing,
original draft, Writing, review \& editing, Visualization, Project
administration. \textbf{Albert Clop}: Methodology, Writing, review \& editing. \textbf{Benjamin Busam}: Writing, review \& editing. \textbf{Ricardo Marques}: Supervision, Writing, review \& editing. \textbf{Petia Radeva}: Supervision, Conceptualization, Resources, Project administration, Funding acquisition, Writing, review \& editing.

\section*{Declaration of competing interest}
The authors declare no known competing financial interests or personal relationships that could have appeared to influence the work reported in this paper.

\section*{Data availability}
Every artifact needed to reproduce the paper is released at \url{https://github.com/GCVCG/MomentAux}: the training and evaluation harness (configs for every cell, the subset indices, the fingerprinted filter banks, the single training command, and the aggregation and audit scripts) and the complete result tables in open formats, together with the pre-registration ledger of Sec.~\ref{sec:prereg} and its commit history, headed by a per-cell table (\computeCells{} rows: dataset, backbone, intervention, fraction, seeds, accuracy, probe accuracy, paired baseline, $\Delta$ with its standard error, $G$ and \readout). The tagged release adds the raw evidence with SHA-256 checksums: every run's final record, every probe record, per-epoch training curves for most runs, and the campaign logs. A small number of runs were archived without their curve, so that one item is incomplete; no reported quantity depends on it, since every number in this paper is computed from the final and probe records. Regenerating the tables and figures runs a small set of exporters, one per task family, covering classification, dense prediction, and detection, and the sign-law audit of Table~\ref{tab:audit} is itself a released script, not a manual query. Public datasets are cited in Sec.~\ref{sec:datasets}, and the released subset indices reproduce the exact image selection without redistributing images. Checkpoints (275\, GB) are not distributed, since every cell retrains from its released configuration.

\section*{Declaration of generative AI and AI-assisted technologies in the
manuscript preparation process} 
During the preparation of this work, the authors used Claude (Anthropic) in order to improve the language and readability of the manuscript. After using this tool, the authors reviewed and edited the content as needed and take full responsibility for the content of the published article.

\bibliographystyle{unsrtnat}
\bibliography{refs}

\appendix
\makeatletter
\renewcommand\section{\@startsection{section}{1}{\z@}%
    {6pt \@plus 2\p@ \@minus 2\p@}%
    {2\p@}%
    {\sectionfont\raggedright\hst[11pt]}}
\makeatother

\section{The self-supervised margin}\label{app:ssl}

\looseness=-1 The margin of a $2\times$-compute SimCLR initialization over the $1.02\times$ prior (CIFAR-100, ResNet-18, plain augmentation) is readable from Table~\ref{tab:budget} as the difference between its SC and Prior columns: $+0.85$, $+2.38$, $+2.31$, $+3.90$, $+4.99$, $+5.01$ and $+4.09$ at $1$--$15\%$, and, on the two fractions Table~\ref{tab:budget} does not carry, $+0.73$ at $50\%$ and $+0.22$ at $100\%$ (three seeds per cell). The shape is the point: unimodal, starving below $1$--$2\%$ where too few images are available to contrast, gone by $50\%$ where every intervention converges. At $5\times$ pre-training compute, the margin instead runs from $+3.1$ to $+9.2$ over $1$--$15\%$, narrowing to $+2.6$ at $25\%$ (Sec.~\ref{sec:budget}).

\section{The law's components}\label{app:law}

\looseness=-1 The component curves behind the sign-law audit. $G$ is non-monotone in data and dataset-specific in shape: CIFAR-10 $4.81$, $5.52$, $3.97$, $0.65$ at $0.5$, $1$, $2.5$, $5$k images; CIFAR-100 $4.16$, $5.14$, $6.26$, $3.55$; Tiny-ImageNet falls monotonically $4.19$, $2.66$, $1.70$ ($1$, $5$, $10$k); STL-10 $4.70$, $4.97$, $3.20$. Where the \readout{} term is flat, the envelope peak coincides with the $G$ peak: the envelope's shape \emph{is} the feature-gain curve. The \readout{} branch, mapped on ceiling-free CIFAR-10, is $+1.56$ at baseline $39.2$, $+1.14$ at $51.7$ and $+0.44$ at both $69.1$ and $80.7$, positive throughout and decaying toward sufficiency; below the crossing it reaches $-2.7$ at bases $8.9$ and $5.3$. The bracket $[31.8, 40.3]$ is pinned by a Swin cell still negative at $31.8$ and a convolutional cell already positive at $40.3$. On byte-identical CIFAR-100 pixels, a full $2{\times}2$ of training against evaluation label space moves $G$ by at most $1\sigma$; on Tiny-ImageNet the same crossing finds two opposing ${\sim}5\sigma$ effects that cancel on the diagonal, which is why $G$ comparisons always fix the evaluation space.

\section{Per-domain and dataset notes}\label{app:domains}

\looseness=-1 EuroSAT: the prior beats SimCLR at 7 of 11 fractions ($+0.6$--$+1.9$ in the 1--10\% band), consistent with SimCLR's photographic view-invariances transferring poorly to satellite statistics while the spectral target is domain-agnostic. DTD is the reverse, SimCLR dominating from a few hundred images, and Food-101 is SSL territory on convolutional backbones. PathMNIST declines with added data for \emph{every} method above ${\sim}10$--$15\%$ ($92.0$ to $86.7$), the signature of its center-shifted test split, so only its low-data cells are interpretable; each domain's statistics decide which currency is scarce, not a photo/non-photo split. On the identity accounting of Table~\ref{tab:partitions}: variable-resolution sources are squash-resized once offline; five relabeled controls share their parents' pixels; four identities are band subsets of two parent instruments, closing the gap between twenty identities and eleven independent image sources.

\section{Selected controls}\label{app:controls}

\looseness=-1 \emph{Contamination.} Re-evaluating the low-data CIFAR-100 cells on ciFAIR-100, which replaces the test set's 927 near-duplicates of training images, moves the prior's gain by at most $0.20$ points at any fraction; both arms eat identical contamination, so it cancels in $\Delta$. \emph{Granularity.} Relabeling CIFAR-100's fixed subsets with their 20 coarse labels (byte-identical pixels, identical steps) reproduces the full gain at matched data ($+5.84$ against $+5.30$ at 5\%): gain follows data and task performance, not class count. The mirrored Tiny-ImageNet control (positional, visually incoherent coarse groups) is the cell that falsified our own earlier account: class count dropped tenfold, baseline accuracy did not rise, and no \readout{} boost appeared, exactly as Eq.~\ref{eq:law} requires and as ``label count drives gain'' does not. \emph{Robustness.} On CIFAR-100-C, corruption robustness tracks $55$--$60\%$ of the clean gain at low data and is neutral at 100\%; the prior neither buys nor costs robustness beyond its accuracy effect.

\section{Operational reproducibility}\label{app:repro}

\looseness=-1 Behind the Data Availability statement: one entry point trains any cell from its configuration name and seed; the result tables regenerate from the run records through one exporter per task family; a released script re-runs the sign-law audit of Table~\ref{tab:audit}. Subset indices, filter-bank fingerprints, and per-run environment records ship with the runs; training refuses to overwrite a completed cell, locks each run, and writes checkpoints atomically; mirrors are verified by evaluating checkpoints against recorded accuracies, not by file listing. An archival DOI will be minted on acceptance.

\end{document}